\documentclass{article}

\usepackage{microtype}
\usepackage{graphicx}
\usepackage{subfigure}
\usepackage{booktabs} 
\usepackage{multirow}

\usepackage{hyperref}
\usepackage[table]{xcolor} 
\usepackage{colortbl}  

\definecolor{myred}{RGB}{255, 0, 0}    %
\definecolor{myblue}{RGB}{0, 0, 249}   %
\definecolor{mygreen}{RGB}{0, 150, 0}  %

\newcommand{\hig}[2]{%
  \ifnum#1=1
    \textcolor{myred}{\textbf{#2}}%
  \else\ifnum#1=2
    \textcolor{myblue}{\textbf{#2}}%
  \else\ifnum#1=3
    \textcolor{mygreen}{\textbf{#2}}%
  \else
    #2%
  \fi\fi\fi
}

\usepackage[accepted]{icml2025}

\usepackage{amsmath}
\usepackage{amssymb}
\usepackage{mathtools}
\usepackage{amsthm}
\usepackage{enumitem}
\usepackage{makecell}
\usepackage{wrapfig}
\usepackage{hyperref}
\usepackage{siunitx}

\usepackage{amsmath,amsfonts,bm}

\def\eqref#1{equation~\ref{#1}}

\def\1{\bm{1}}

\def\vone{{\mathbf{1}}}

\def\vh{{\mathbf{h}}}

\def\vs{{\mathbf{s}}}
\def\vt{{\mathbf{t}}}

\def\vw{{\mathbf{w}}}

\def\vz{{\mathbf{z}}}

\def\mA{{\mathbf{A}}}

\def\mD{{\mathbf{D}}}

\def\mH{{\mathbf{H}}}
\def\mI{{\mathbf{I}}}

\def\mL{{\mathbf{L}}}

\def\mP{{\mathbf{P}}}

\def\mS{{\mathbf{S}}}
\def\mT{{\mathbf{T}}}
\def\mU{{\mathbf{U}}}

\def\mW{{\mathbf{W}}}
\def\mX{{\mathbf{X}}}

\def\mZ{{\mathbf{Z}}}

\def\mLambda{{\mathbf{\Lambda}}}

\DeclareMathAlphabet{\mathsfit}{\encodingdefault}{\sfdefault}{m}{sl}
\SetMathAlphabet{\mathsfit}{bold}{\encodingdefault}{\sfdefault}{bx}{n}

\def\gG{{\mathcal{G}}}

\usepackage{titletoc}
\usepackage{tcolorbox}

\usepackage[capitalize,noabbrev]{cleveref}

\theoremstyle{plain}
\newtheorem{theorem}{Theorem}[section]
\newtheorem{proposition}[theorem]{Proposition}
\newtheorem{lemma}[theorem]{Lemma}

\theoremstyle{definition}
\newtheorem{definition}[theorem]{Definition}

\theoremstyle{remark}

\usepackage[textsize=tiny]{todonotes}

\icmltitlerunning{Rethinking Link Prediction for Directed Graphs}

\begin{document}
\setlength{\textfloatsep}{10pt}

\twocolumn[
\icmltitle{Rethinking Link Prediction for Directed Graphs}




\icmlsetsymbol{equal}{*}

\begin{icmlauthorlist}
\icmlauthor{Mingguo He}{ruc}
\icmlauthor{Yuhe Guo}{ruc}
\icmlauthor{Yanping Zheng}{ruc}
\icmlauthor{Zhewei Wei}{ruc}
\icmlauthor{Stephan Günnemann}{tum}
\icmlauthor{Xiaokui Xiao}{nus}
\end{icmlauthorlist}

\icmlaffiliation{ruc}{Gaoling School of Artifical Intelligence, Renmin University of China}
\icmlaffiliation{tum}{Technical University of Munich}
\icmlaffiliation{nus}{National University of Singapore}

\icmlcorrespondingauthor{Mingguo He}{mingguo@ruc.edu.cn}

\icmlkeywords{Machine Learning, ICML}

\vskip 0.3in
]



\printAffiliationsAndNotice{}  

\begin{abstract}
Link prediction for directed graphs is a crucial task with diverse real-world applications. Recent advances in embedding methods and Graph Neural Networks (GNNs) have shown promising improvements. However, these methods often lack a thorough analysis of their expressiveness and suffer from effective benchmarks for a fair evaluation. In this paper, we propose a unified framework to assess the expressiveness of existing methods, highlighting the impact of dual embeddings and decoder design on directed link prediction performance. To address limitations in current benchmark setups, we introduce \textit{DirLinkBench}, a robust new benchmark with comprehensive coverage, standardized evaluation, and modular extensibility. 
The results on DirLinkBench show that current methods struggle to achieve strong performance, while DiGAE outperforms other baselines overall. We further revisit DiGAE theoretically, showing its graph convolution aligns with GCN on an undirected bipartite graph.
Inspired by these insights, we propose a novel \textit{Spectral Directed Graph Auto-Encoder SDGAE} that achieves state-of-the-art average performance on DirLinkBench. Finally, we analyze key factors influencing directed link prediction and highlight open challenges in this field. The code is available at \href{https://github.com/ivam-he/DirLinkBench-SDGAE}{here}.

\end{abstract}
\section{Introduction}\label{intro}
A directed graph (or digraph) is a type of graph in which the edges between nodes have a specific direction. These graphs are commonly used to model real-world asymmetric relationships, such as ``following" and ``followed" in social networks~\cite{leskovec2016snap} or ``link" and ``linked" in web pages~\cite{Page1999ThePC}. Directed graphs capture the inherent directionality of these relationships and provide a more accurate representation of complex systems. Link prediction is a fundamental and widely studied task in directed graphs, with numerous real-world applications. Examples include predicting follower relationships in social networks~\cite{liben2003link}, recommending products in e-commerce~\cite{rendle2009bpr}, and detecting intrusions in network security~\cite{bhuyan2013network}.


\begin{table}[t]

\centering
\caption{Overview of directed graph learning methods.} 
\resizebox{\columnwidth}{!}{
\begin{tabular}{l|l|l}
\toprule
  Method & Name &Embedding \\ \midrule
\multirow{8}{*}{\shortstack{Embedding \\ Methods}}        
& HOPE~\cite{hope}    &$\vs_{u}, \vt_{u}$ \\
& APP~\cite{app} &$\vs_{u}, \vt_{u}$    \\
& AROPE~\cite{arope} &$\vs_{u}, \vt_{u}$ \\
& STRAP~\cite{strap} &$\vs_{u}, \vt_{u}$  \\
& NERD~\cite{nerd} &$\vs_{u}, \vt_{u}$   \\
& DGGAN~\cite{dggan} &$\vs_{u}, \vt_{u}$ \\
& ELTRA~\cite{eltra} &$\vs_{u}, \vt_{u}$  \\ 
& ODIN~\cite{odin} &$\vs_{u}, \vt_{u}$ \\
                                       \hline
\multirow{13}{*}{\shortstack{Graph Neural\\ Networks}}
&DiGAE~\cite{digae} &$\vs_{u}, \vt_{u}$      \\
&CoBA~\cite{coba}  &$\vs_{u}, \vt_{u}$ \\
&BLADE~\cite{blade} &$\vs_{u}, \vt_{u}$ \\

&Gravity GAE~\cite{gragae} &$\vh_{u}, m_{u}$ \\
&DHYPR~\cite{dhypr} &$\vh_{u}, m_{u}$ \\

&DGCN~\cite{dgcn-tong} &$\vh_{u}$ \\
&DiGCN \& DiGCNIB~\cite{digcn} &$\vh_{u}$       \\
&DirGNN~\cite{dirgnn} &$\vh_{u}$ \\
&HoloNets~\cite{holonets} &$\vh_{u}$\\
&NDDGNN~\cite{nddgnn} &$\vh_{u}$\\ 

&MagNet~\cite{magnet} &$\vz_{u}$       \\
&LightDiC~\cite{lightdic} &$\vz_{u}$   \\
&DUPLEX~\cite{duplex} &$\vz_{u}$   \\
                                      
                                      \bottomrule
\end{tabular}}
\vspace{-2mm}
\label{digraph_methods}
\end{table}

Machine learning techniques have been extensively developed to enhance link prediction performance on directed graphs. Existing methods can be broadly categorized into \textbf{embedding methods}
and \textbf{graph neural networks (GNNs)}. Embedding methods aim to preserve the asymmetry of directed graphs by generating two separate embeddings for each node $u$: a source embedding $\vs_u$ and a target embedding $\vt_u$~\cite{eltra,odin}, which are also known as content/context representations~\cite{line,hope,strap}. GNNs, on the other hand, can be further divided into four classes based on the types of embeddings they generate. (1) Source-target methods, similar to embedding methods, employ specialized propagation mechanisms to learn distinct source and target embeddings for each node~\cite{digae,coba}. (2) Gravity-inspired methods, inspired by Newton's law of universal gravitation, learn a real-valued embedding
$\vh_u \in \mathbb{R}^d$ and a mass parameter $m_u \in \mathbb{R}^{+}$ for each node $u$~\cite{gragae,dhypr}. 
(3) Single real-valued methods follow a conventional approach by learning a single real-valued embedding $\vh_u \in \mathbb{R}^d$~\cite{dirgnn,nddgnn}.
(4) Complex-valued methods use Hermitian adjacency matrices, learning complex-valued embeddings $\vz_u \in \mathbb{C}^d$~\cite{magnet,duplex}.
We summarize these methods in Table~\ref{digraph_methods}.

Although the methods described above have achieved promising results in directed link prediction, several challenges remain. 
First, it is unclear what types of embedding are effective in predicting directed links, as there is a lack of comprehensive research assessing these methods' expressiveness.
Second, existing methods have not been fairly compared and evaluated, highlighting the need for a robust benchmark for directed link prediction. \textbf{Current experimental setups face multiple issues}, such as the omission of basic baselines (e.g., MLP shown in Figures~\ref{fig:dp} and~\ref{fig:ep}),  label leakage (illustrated in Tables~\ref{tb:duplex} and~\ref{tb:app_duplex}), class imbalance (shown in Figures~\ref{fig:duplex_cls_cora} and~\ref{fig:duplex_cls_citeseer}), single evaluation metrics, and inconsistent dataset splits. We discuss these issues in detail in Section~\ref{issues}.


In this paper, we first propose a unified learning framework for directed link prediction methods to assess the expressiveness of different embedding types. 
We demonstrate that dual methods (including source-target, gravity-inspired, and complex-valued methods) are more expressive for directed link prediction than single methods (i.e., single real-valued methods).
Meanwhile, we highlight the often-overlooked importance of decoder design in achieving better performance, as most research has primarily focused on encoders. To address the limitations of existing experimental setups, \textbf{we introduce DirLinkBench, a robust new benchmark for directed link prediction} that offers comprehensive coverage (seven real-world datasets, 16 baseline methods, and seven evaluation metrics), standardized evaluation (unified splits, feature inputs, and task setups), and modular extensibility (support for adding new datasets, decoders, and sampling strategies).

The results in DirLinkBench reveal that current methods struggle to achieve strong and firm performance across diverse datasets. Interestingly, a simple directed graph auto-encoder, DiGAE~\cite{digae}, outperforms other baselines in general. We revisit DiGAE from a theoretical perspective and observe that its graph convolution is equivalent to the GCN~\cite{gcn} convolution on an undirected bipartite graph. Building on this insight, \textbf{we propose SDGAE, a novel spectral directed graph auto-encoder} that learns arbitrary spectral filters via polynomial approximation. SDGAE achieves state-of-the-art (SOTA) results on four of the seven datasets and ranks highest on average. Finally, we investigate key factors influencing directed link prediction (e.g., input features, decoder design, and degree distribution) and conclude with open challenges to advance the field.
We summarize our contributions as follows.
\begin{itemize}
    \item We propose a unified framework to assess the expressiveness of directed link prediction methods, showing that dual methods are more expressive and highlighting the importance of decoder design.
    
    
    
    \item We introduce DirLinkBench, a robust new benchmark with comprehensive coverage, standardized evaluation, and modular extensibility for directed link prediction.
    
    \item We propose a novel directed graph auto-encoder SDGAE, inspired by the theoretical insights of DiGAE and achieving SOTA average performance on DirLinkBench.
    
    \item We empirically analyze the factors affecting the performance of directed link prediction and highlight open challenges for future research.
\end{itemize}



\section{Background and related work}\label{app:related_work}
In this section, we first introduce the background and related work 
of directed graph learning methods, which can be broadly categorized into embedding methods and graph neural networks (GNNs). Additionally, we will introduce some background on spectral-based GNNs for undirected graphs.  

\textbf{Notation}. We consider a directed, unweighted graph $\gG = (V, E)$, with node set $V$ and edge set $E$. Let $n = |V|$ and $m = |E|$ represent the number of nodes and edges in $\gG$, respectively. We use $\mA$ to denote the adjacency matrix of $\gG$, where $\mA_{uv} = 1$ if there exists a directed edge from node $u$ to node $v$, and $\mA_{uv} = 0$ otherwise.
The Hermitian adjacency matrix of $\gG$ is denoted by $\mH$ and defined as $\mH = \mA_{\rm s} \odot \exp\left(i\frac{\pi}{2}\mathbf{\Theta}\right)$. Here, $\mA_{\rm s} = \mA \cup \mA^{\top}$ is the adjacency matrix of the undirected graph derived from $\gG$, and $\mathbf{\Theta} = \mA - \mA^{\top}$ is a skew-symmetric matrix. $i$ is the imaginary unit. We denote the out-degree and in-degree matrices of $\mA$ by $\mD_{\rm out} = {\rm diag}(\mA\vone)$ and $\mD_{\rm int} = {\rm diag}(\mA^{\top}\vone)$, respectively, where $\vone$ is the all-one vector. Let $\mX \in \mathbb{R}^{n \times d^{\prime}}$ denote the node feature matrix, where each node has a $d^{\prime}$-dimensional feature vector. 

\subsection{Embedding Methods} 
Embedding methods for directed graphs aim to capture asymmetric relationships. Most approaches assign each node $u$ two embedding vectors: a source embedding $\vs_u$ and a target embedding $\vt_u$. These embeddings are typically learned using either factorization-based or random walk-based techniques.
Factorization-based methods include HOPE~\cite{hope}, which computes Katz similarity~\cite{katz1953new} followed by singular value decomposition (SVD)\cite{golub2013matrix}, and AROPE~\cite{arope}, which generalizes this approach to preserve arbitrary-order proximities. STRAP~\cite{strap} extends this idea by combining Personalized PageRank (PPR)~\cite{Page1999ThePC} scores from both the original and transposed graphs before applying SVD. Random walk-based methods include APP~\cite{app}, which trains embeddings using PPR-guided random walks, and NERD~\cite{nerd}, which samples nodes based on degree distributions. Other recent notable methods include DGGAN~\cite{dggan} (which employs adversarial training), ELTRA~\cite{eltra} (based on ranking-oriented learning ), and ODIN~\cite{odin} (which incorporates degree bias separation). All these methods focus on generating source–target embeddings from graph structures to support link prediction tasks.

\subsection{Graph Neural Networks for Directed Graphs}
Graph Neural Networks for directed graphs can be broadly classified into four categories based on their embedding strategies: (1) Source–target methods learn separate source and target embeddings for each node. DiGAE~\cite{digae} applies GCN's convolutions~\cite{gcn} to both the adjacency matrix and its transpose. CoBA~\cite{coba} jointly aggregates source and target neighbors, while BLADE~\cite{blade} introduces an asymmetric loss to learn dual embeddings from local neighborhoods. (2) Gravity-inspired methods are motivated by Newton’s law of universal gravitation, learning real-valued node embeddings along with a scalar mass parameter. Gravity GAE~\cite{gragae} combines a gravity-based decoder with a GCN-like encoder, and DHYPR~\cite{dhypr} extends this approach through hyperbolic collaborative learning. (3) Single real-valued methods generate a single real-valued embedding per node. DGCN~\cite{dgcn-tong} constructs a directed Laplacian using first- and second-order proximities. MotifNet~\cite{motifnet} employs motif-based Laplacians. DiGCN and DiGCNIB~\cite{digcn} generalize directed Laplacians using Personalized PageRank (PPR). DirGNN~\cite{dirgnn} introduces a flexible convolution framework for directed message-passing neural networks (MPNNs), and HoloNets~\cite{holonets} leverage holomorphic functional calculus with an architecture similar to DirGNN. (4) Complex-valued methods utilize Hermitian adjacency matrices to learn complex-valued embeddings. MagNet~\cite{magnet} defines a magnetic Laplacian to construct directed graph convolutions. LightDiC~\cite{lightdic} scales this approach to large graphs via a decoupled design, while DUPLEX~\cite{duplex} employs dual GAT encoders with Hermitian adjacency matrices. Additional methods include adapting Transformers to directed graphs~\cite{tf_digraph} and extending over-smoothing analyses to the directed setting~\cite{fractional}. While these methods propose various convolutional and propagation mechanisms for directed graphs, fair evaluation for directed link prediction tasks is often lacking. Many existing experimental setups omit key baselines or suffer from significant issues, such as label leakage. These challenges motivate the development of this work.

\begin{table*}[ht]
\centering
\caption{A unified framework for directed link prediction methods.}
\label{table_fram}
\begin{tabular}{@{}lllllllll@{}}
\toprule
Encoder $\mathrm{Enc}(\cdot)$ 
 &\multicolumn{2}{l}{Embeddings $({\bm \theta_u}, {\bm \phi_u})$}
 &\multicolumn{6}{l}{Possible Decoder $\mathrm{Dec}(\cdot)$}\\
\midrule

Single real-valued  & $\vh_u = \bm{\theta}_u$,  & $\varnothing = \bm{\phi}_u$          & \multicolumn{2}{l}{$\sigma\bigl(\vh_u^{\top}\vh_v\bigr)$;} & \multicolumn{2}{l}{$\mathrm{MLP}\bigl(\vh_u \odot \vh_v\bigr)$;} & \multicolumn{2}{l}{\quad$\mathrm{MLP}\bigl(\vh_u \|\vh_v\bigr)$} \\[3.0pt]

Source-target  & $\vs_u = \bm{\theta}_u$,           & $\vt_u = \bm{\phi}_u$          & \multicolumn{2}{l}{$\sigma\bigl(\vs_u^{\top}\vt_v\bigr)$;} & \multicolumn{2}{l}{$\mathrm{MLP}\bigl(\vs_u \odot \vt_v\bigr)$;} & \multicolumn{2}{l}{\quad$\mathrm{LR}\bigl(\vs_u \|\vt_v\bigr)$} \\[3.0pt]

Complex-valued  & \multicolumn{2}{l}{$\vz_u = \bm{\theta}_u\odot\exp\bigl(i\,\bm{\phi}_u\bigr)$}  & \multicolumn{2}{l}{$\mathrm{Direc}\bigl(\vz_u,\vz_v\bigr)$;} & \multicolumn{4}{l}{$\mathrm{MLP}\bigl(\bm{\theta}_u \|\bm{\theta}_v \|\bm{\phi}_u \|\bm{\phi}_v\bigr)$} \\[3.0pt]

Gravity-inspired  & $\vh_u = \bm{\theta}_u$,  &$m_u = g(\bm{\phi}_u)$         & \multicolumn{3}{l}{$\sigma\bigl(m_v - \lambda\log\|\vh_u-\vh_v\|_2^2\bigr)$;}              & \multicolumn{3}{l}{$\sigma\bigl(m_v - \lambda\log \bigl(\mathrm{dist}(\vh_u,\vh_v)\bigr)\bigr)$}  \\
\bottomrule
\end{tabular}
\vspace{-2mm}
\end{table*}

\subsection{Spectral-based Graph Neural Networks}
In recent years, spectral-based Graph Neural Networks (GNNs) have garnered significant attention and demonstrated strong performance across various tasks~\cite{bo2023survey}. Many popular methods in this category approximate spectral graph filters using polynomials of the adjacency or Laplacian matrix~\cite{chebnet,gprgnn,bernnet,jacobiconv,chebnetii}.
Specifically, let $\mA_{\rm s}$ denote the adjacency matrix of the derived undirected graph for $\gG$, and let $\mD_{\rm s}$ be its diagonal degree matrix, where $\mD_{\rm s}[i,i] = \sum\nolimits_{j}\mA_{\rm s}[i,j]$. We define the symmetric normalized adjacency matrix as $\mP = \mD_{\rm s}^{-1/2}\mA_{\rm s}\mD_{\rm s}^{-1/2}$. The propagation process in spectral-based GNNs is then given by:
\begin{equation}\label{spec_gnn}
\mZ=h(\mP)\mX \approx \sum\limits_{k=0}^K w_k \mP^k\mX,
\end{equation}
where $h(\mP)$ represents the spectral graph filter and $w_k$ are the polynomial coefficients. From a spectral perspective, we denote the eigendecomposition of the symmetric normalized adjacency matrix as $\mP = \mU\mLambda\mU^{\top}$, where $\mU$ is the matrix of eigenvectors and $\mLambda$ is the diagonal matrix of eigenvalues. Accordingly, the spectral graph filter can be expressed as $h(\mP) = \mU h(\mLambda)\mU^{\top}$, meaning it operates as a function of the eigenvalues $\mLambda$. Thus, spectral-based GNNs approximate the graph filter function $h(\mLambda)$ in the spectral domain using a polynomial expansion. Notably, spectral filters can also be equivalently defined using the Laplacian matrix. In this case, the propagation process becomes $\mZ = \sum\nolimits_{k=0}^{K} w_k \mL^k \mX$, where $\mL = \mI - \mP$ is the normalized Laplacian matrix. Within this framework, many powerful spectral-based GNNs have been proposed. For example, ChebNet~\cite{chebnet} uses Chebyshev polynomials to approximate spectral filters, while GCN~\cite{gcn} simplifies ChebNet to improve efficiency. More recently, GPR-GNN~\cite{gprgnn} learns the coefficients $w_k$ directly, and both BernNet~\cite{bernnet} and JacobiConv~\cite{jacobiconv} use Bernstein and Jacobi polynomial bases, respectively. Despite their effectiveness, these methods are primarily designed for undirected graphs. The development of analogous approaches that leverage polynomial approximations of spectral filters for directed graphs remains a relatively unexplored area.

\section{Rethinking Directed Link Prediction}
In this section, we will revisit the link prediction task for directed graphs and introduce a unified framework to assess the expressiveness of existing methods. Meanwhile, we examine the current experimental setups for directed link prediction tasks and highlight four significant issues.


\subsection{Unified Framework for Directed Link Prediction}\label{sec_unif}
The link prediction task on directed graphs is to predict potential directed links (edges) in an observed graph \( \gG^{\prime} \), with the given structure of \( \gG^{\prime} \) and node feature \( \mX \). Formally,
\begin{definition}\label{de_dlp}
\textbf{Directed link prediction problem}. Given an observed graph $\gG^{\prime}=(V, E^{\prime})$ and node feature $\mX$, the goal of directed link prediction is to predict the likelihood of a directed edge $(u,v) \in E^{\star}$ existing, where $E^{\star} \subseteq (V \times V) \setminus E^{\prime}$. The probability of edge $(u,v)$ existing is given by 
\begin{equation}
    p(u,v) = f(u \to v \mid \gG^{\prime}, \mX).
\end{equation}
\end{definition}
The $f(\cdot)$ denotes a prediction model, such as embedding methods or graph neural networks. 
Unlike link prediction on undirected graphs~\cite{zhang2018link}, for directed graphs, it is necessary to account for directionality. Specifically, $p(u,v)$  and $p(v,u)$ are not equal; they represent the probability of a directed edge existing from node $u$ to node $v$, and from node $v$ to node $u$, respectively. To evaluate the expressiveness of existing methods for directed link prediction, we propose a unified framework: 
\begin{align}
    ({\bm \theta}_u, {\bm \phi}_u) &= \mathrm{Enc}(\gG^{\prime}, \mX, u), \quad \forall u \in V, \label{eq_encoder} \\
    p(u, v) &= \mathrm{Dec}({\bm \theta}_u, {\bm \phi}_u, {\bm \theta}_v, {\bm \phi}_v), \quad \forall (u, v) \in E^{\star}. \label{eq_decoder}
\end{align}
Here, $\mathrm{Enc}(\cdot)$ represents an encoder function, which includes various methods described in Section~\ref{intro}. 
And ${\bm \theta}_u \in \mathbb{R}^{d_{\theta}}$, ${\bm \phi}_u \in \mathbb{R}^{d_{\phi}}$ are real-valued dual embeddings of dimensions ${d_{\theta}}$ and ${d_{\phi}}$, respectively. $\mathrm{Dec}(\cdot)$ is a decoder function tailored to the specific encoder method. This framework unifies existing methods for directed link prediction, as summarized in Table~\ref{table_fram}. Specifically, here $\sigma$ represents the activation function (e.g., $\mathrm{Sigmoid}$), while $\mathrm{MLP}$ and $\mathrm{LR}$ denote the multilayer perceptron and the logistic regression predictor, respectively. The symbols $\odot$ and $\|$ represent the Hadamard product and the vector concatenation process, respectively. Then, we provide details on the embeddings and possible decoders used by the different methods listed in Table~\ref{table_fram}.
\begin{itemize}
\item For single real-valued encoder, we define the real-valued embedding as $\vh_u = \bm{\theta}_u$ and set $\bm{\phi}_u=\varnothing$, where $\varnothing$ denotes nonexistence. Possible decoders include:
$\sigma\bigl(\vh_u^{\top}\vh_v\bigr)$,  
$\mathrm{MLP}\bigl(\vh_u \odot \vh_v\bigr)$,  
$\mathrm{MLP}\bigl(\vh_u \|\vh_v\bigr)$. 
\item  For source-target encoder, we define the source embedding as $\vs_u = \bm{\theta}_u$ and the target embedding as $\vt_u=\bm{\phi}_u$ for each node $u \in V$. Possible decoders include:
$\sigma\bigl(\vs_u^{\top}\vt_v\bigr)$,  
$\mathrm{MLP}\bigl(\vs_u \odot \vt_v\bigr)$,  
$\mathrm{LR}\bigl(\vs_u \|\vt_v\bigr)$, etc.  
Here, $\vs_u^{\top}\vt_v$ denotes the inner product of the source and target embeddings.
\item For complex-valued encoder, we define the complex-valued embedding as $\vz_u=\bm{\theta}_u\odot\exp({i\bm{\phi}_u})$, where $i$ is the imaginary unit. Possible decoders include:
$\mathrm{Direc}\bigl(\vz_u,\vz_v\bigr)$,  
$\mathrm{MLP}\bigl(\bm{\theta}_u \|\bm{\theta}_v \|\bm{\phi}_u \|\bm{\phi}_v\bigr)$,  
where $\mathrm{Direc}(\cdot)$ is a direction-aware function defined in DUPLEX~\cite{duplex}.
\item For gravity-inspired encoder, we define the real-valued embedding as $\vh_u = \bm{\theta}_u$ and set the mass parameter $m_u = g(\bm{\phi}_u)$, where $g(\cdot)$ is a function or neural network that converts $\bm{\phi}_u$ into a scalar~\cite{dhypr}. Possible decoders include:
$\sigma\bigl(m_v - \lambda\log\|\vh_u-\vh_v\|_2^2\bigr)$ ~\cite{gragae},  
$\sigma\bigl(m_v - \lambda\log \bigl(\mathrm{dist}(\vh_u,\vh_v)\bigr)\bigr)$~\cite{dhypr}.  
Here, $\lambda$ is a hyperparameter and $\mathrm{dist}(\cdot)$ represents the hyperbolic distance.
\end{itemize}
Based on this framework, if \(0 < d_{\theta}, d_{\phi} \ll n\), these encoders involve two real embeddings \(\bm{\theta}_u\) and \(\bm{\phi}_u\), which we refer to as \textbf{dual methods}. Conversely, if \(d_{\phi} = 0\), these encoders have only a single real embedding \(\bm{\theta}_u\), and are referred to as \textbf{single methods}. Next, we analyse the expressiveness of dual and single methods in terms of asymmetry preservation and graph reconstruction.

\begin{table*}[t]
\centering
\caption{The results of MLP and baselines under PyGSD~\cite{dpyg} setup on \textbf{Direction Prediction (DP)} task.}
\begin{tabular}{lcccccc}
\toprule
Method & Cora-ML & CiteSeer & Telegram & Cornell & Texas & Wisconsin \\ \midrule
MLP       &86.13$\pm$0.45  & 85.51$\pm$0.63  & 95.61$\pm$0.15  & \textbf{86.40$\pm$2.60}  &\textbf{83.08$\pm$4.33}  & \textbf{87.95$\pm$2.65}  \\ \midrule
DGCN      & 85.49$\pm$0.75  & 84.85$\pm$0.56  & 96.03$\pm$0.35  & 84.55$\pm$3.71  & 79.59$\pm$5.09  & 84.57$\pm$3.59  \\ 
DiGCN     & 85.37$\pm$0.54  & 83.88$\pm$0.82  & 94.95$\pm$0.54  & 85.41$\pm$2.78  & 76.57$\pm$3.98  & 82.41$\pm$2.56  \\ 
DiGCNIB   & 86.12$\pm$0.42  & 85.58$\pm$0.56  & 95.99$\pm$0.44  & 86.28$\pm$3.37  &82.27$\pm$3.51  &87.07$\pm$2.16  \\ 
MagNet    & \textbf{86.33$\pm$0.54}  & \textbf{85.80$\pm$0.63}  & \textbf{96.97$\pm$0.21}  & 83.29$\pm$4.28  & 80.25$\pm$4.78  & 86.60$\pm$2.72  \\ \bottomrule
\end{tabular}
\label{tb_dpyg_dp}
\end{table*}

\begin{table*}[t]
\centering
\caption{The results of MLP and baselines under PyGSD~\cite{dpyg} setup on \textbf{Existence Prediction (EP)} task.}
\begin{tabular}{lcccccc}
\toprule
Method & Cora-ML & CiteSeer & Telegram & Cornell & Texas & Wisconsin \\ \midrule
MLP       & \textbf{78.85$\pm$0.83}  & 71.03$\pm$0.89  & 82.72$\pm$0.57  & 68.41$\pm$2.05  & \textbf{69.58$\pm$2.29}  & \textbf{72.53$\pm$2.27}  \\ \midrule 
DGCN      & 76.45$\pm$0.49  & 70.72$\pm$0.79  & 83.18$\pm$1.55  & 67.95$\pm$2.27  & 63.65$\pm$2.40  & 68.12$\pm$2.86    \\ 
DiGCN     & 76.17$\pm$0.49  & 72.00$\pm$0.88  & 83.12$\pm$0.43  & 67.16$\pm$1.82  & 63.54$\pm$3.85  & 67.01$\pm$2.47    \\ 
DiGCNIB   & 78.80$\pm$0.51  & \textbf{74.55$\pm$0.91}  & 84.49$\pm$0.51  & \textbf{69.77$\pm$2.05}  & 67.60$\pm$3.44  & 70.78$\pm$2.79   \\ 
MagNet    & 77.37$\pm$0.45  & 71.47$\pm$0.71  & \textbf{85.82$\pm$0.39}  & 68.98$\pm$2.27  & 65.94$\pm$1.88  & 71.23$\pm$2.53    \\  \bottomrule
\end{tabular}
\label{tb_dpyg_ep}
\vspace{-2mm}
\end{table*}

\textbf{Asymmetry preservation}. Previous source-target encoders have demonstrated that using the source and target embeddings effectively preserves asymmetric information in directed graphs, a capability that single methods lack~\cite{app,strap,hope,eltra}. \textbf{We extend this claim by arguing that all dual methods can preserve asymmetric information in directed graphs}, including complex-valued and gravity-inspired encoders.
Specifically, complex-valued methods~\cite{magnet,duplex} encode directionality as a geometric difference in complex space, meaning that edges from node \(u\) to node \(v\) and edges from node \(v\) to node \(u\) can be distinguished by the difference between \(\vz^{\top}_u\overline{\vz}_v\) and \(\vz^{\top}_v\overline{\vz}_u\). The gravity-inspired methods~\cite{gragae,dhypr}, on the other hand, use direction-sensitive gravity to preserve asymmetry, i.e., \(Gm_u/||\vh_u-\vh_v||^2\) and \(Gm_v/||\vh_u-\vh_v||^2\) distinguish between edges from node \(u\) to node \(v\) and edges from node \(v\) to node \(u\), where \(G\) is the gravitational constant. Thus, in terms of preserving asymmetry in directed graphs, dual methods have a significant advantage over single methods, as the former can naturally preserve the asymmetry of directed edges, facilitating link prediction task.

\textbf{Graph reconstruction}. 
Graph reconstruction involves using node embeddings $\{\bm{\theta}_u, \bm{\phi}_u\}$ and the decoder function $\mathrm{Dec}(\cdot)$ to compute the probability $p(u,v)$ of each directed edge. The top-\(m\) edges are then selected to reconstruct the original graph. 
For accurate reconstruction, the encoder $\mathrm{Enc}(\cdot)$ and decoder $\mathrm{Dec}(\cdot)$ need ensure $p(u,v) \to 1$ if an edge exists from $u$ to $v$, and $p(u,v) \to 0$ otherwise. This task requires the model to capture structural properties of the graph, including edge existence and directionality, reflecting the expressiveness for the directed link prediction~\cite{strap,eltra}.  
Dual methods benefit from the theoretical advantage of asymmetry preservation. When equipped with a suitable $\mathrm{Dec}(\cdot)$ function, these methods can achieve effective graph reconstruction. The underlying intuition is that: source-target methods can represent the neighbor matrix as \(\mA_{uv} = \vs_u^{\top} \vt_v\)~\cite{strap,hope,digae}, complex-valued methods can represent the Hermitian adjacency matrix as \(\mH_{uv} = \vz_u^{\top} \overline{\vz}_v\)~\cite{magnet,duplex}, and gravity-inspired methods can represent the neighbor matrix as \(\mA_{uv} = Gm_u/||\vh_u - \vh_v||^2\)~\cite{gragae,dhypr}. 
In contrast, single methods lack the intrinsic preservation of asymmetry, but they can partially capture the graph structure by using asymmetric decoders, such as $\mathrm{MLP}(\vh_u \| \vh_v)$. While this allows for limited structural reconstruction, single methods are theoretically insufficient to represent arbitrary directed graphs. As formalized in Proposition~\ref{prop_single} (see the Appendix for the proof), they fail to reconstruct certain structures, such as directed ring graphs.  Therefore, although asymmetric decoders enhance the performance of single methods, their overall expressiveness remains limited compared to dual methods.

\begin{figure}[t]
    \centering
   \vspace{-3mm}
   \hspace{-4.8mm}
   \subfigure[Direction Prediction]{
   \includegraphics[width=40mm]{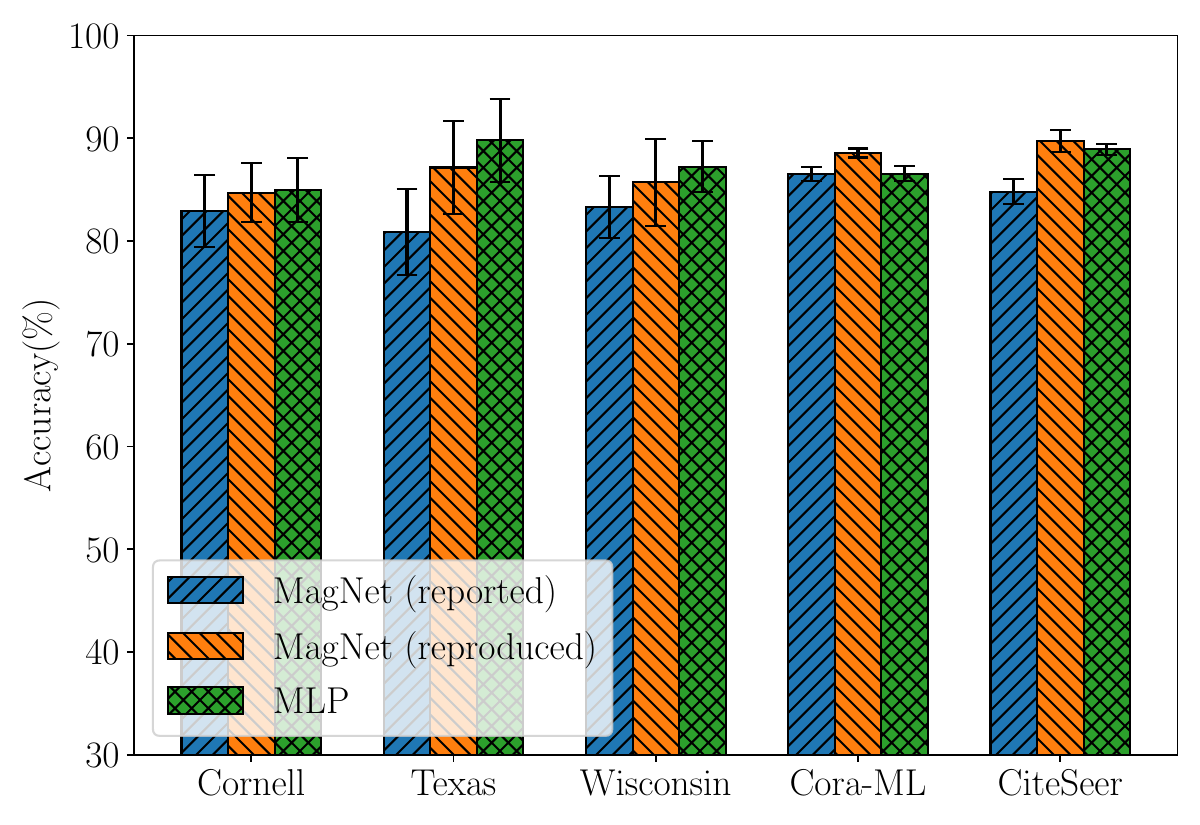}
   \label{fig:dp}
   }
   \hspace{-3.5mm}
   \subfigure[Existence Prediction]{
   \includegraphics[width=40mm]{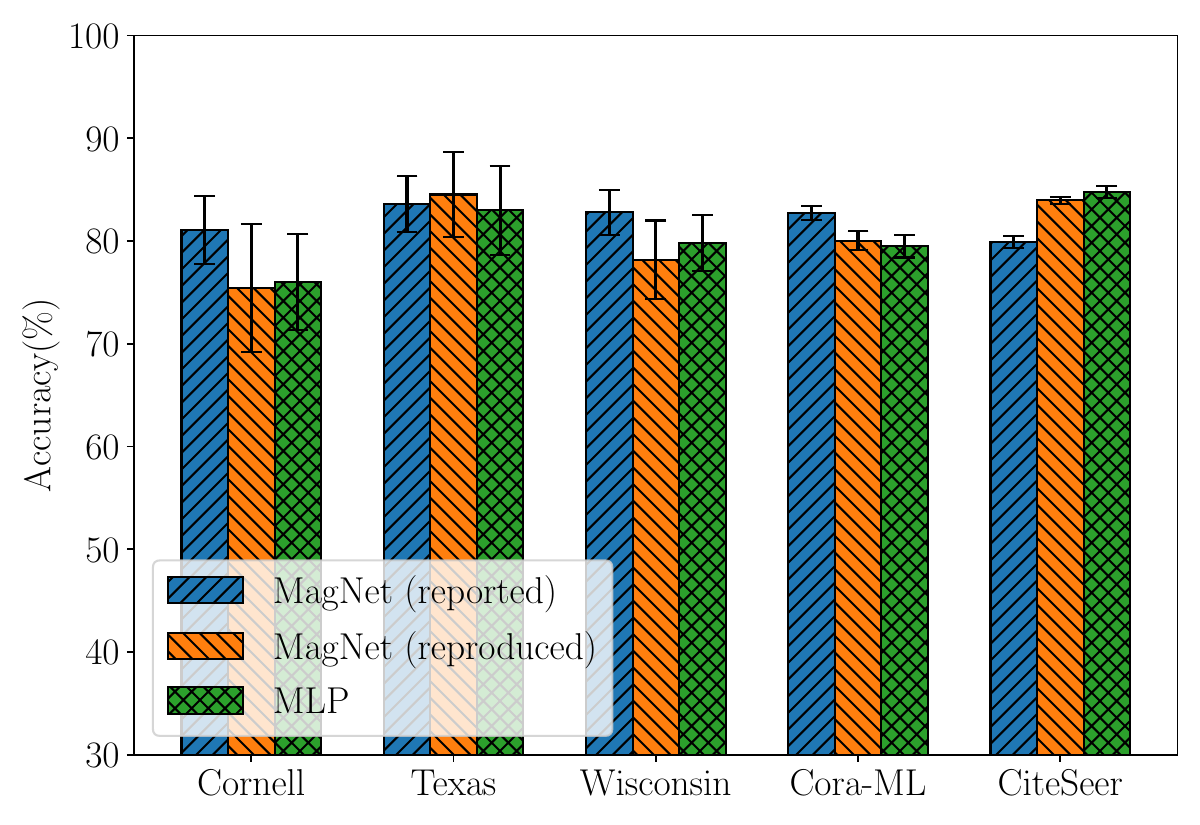}
   \label{fig:ep}
   }
   \hspace{-1.8mm}
   \caption{The results of MagNet~\cite{magnet} as reported in the original paper, alongside the reproduced MagNet and MLP results.}
   \vspace{-2mm}
   \label{fig:magnet_mlp}
\end{figure}

\begin{proposition}\label{prop_single}
    Single methods (single real-valued embedding \(\vh_u\)) with an asymmetric decoder function $\mathrm{MLP}(\vh_u \| \vh_v)$ can capture graph structure and enable reconstruction for some specific directed graphs, but not arbitrary directed graphs, such as directed ring graphs.
\end{proposition}

Overall, \textbf{dual methods have a clear theoretical advantage in both asymmetry preservation and graph reconstruction}, making them more expressive for link prediction compared to single methods. However, we also want to highlight \textbf{the critical role that decoder function design plays in link prediction tasks}. While single embeddings are inherently limited, they can still benefit from asymmetric decoders in practical applications. These theoretical insights motivate a deeper empirical analysis of how different encoder and decoder functions impact performance in link prediction tasks.

\begin{table*}[th]
\centering
\caption{Link prediction results on \textbf{Cora} dataset under the DUPLEX~\cite{duplex} setup: results without superscripts are from the DUPLEX paper, $^\dagger$ indicates reproduction with test set edges in training, and $^\ddagger$ indicates reproduction without test set edges in training.}
\begin{tabular}{@{}lllllll@{}}
\toprule

Method &EP(ACC) &EP(AUC) &DP(ACC) &DP(AUC) &3C(ACC) &4C(ACC) \\ \midrule
 MagNet &81.4$\pm$0.3 &89.4$\pm$0.1 &88.9$\pm$0.4 &95.4$\pm$0.2 &66.8$\pm$0.3 &63.0$\pm$0.3 \\
 DUPLEX &93.2$\pm$0.1 &95.9$\pm$0.1 &95.9$\pm$0.1 &97.9$\pm$0.2 &92.2$\pm$0.1 &88.4$\pm$0.4  \\ \midrule

 DUPLEX$^\dagger$ &93.49$\pm$0.21 &95.61$\pm$0.20 &95.25$\pm$0.16 &96.34$\pm$0.23  &92.41$\pm$0.21 &89.76$\pm$0.25 \\

 MLP$^\dagger$ &88.53$\pm$0.22	&\textbf{95.46$\pm$0.18}	&\textbf{95.76$\pm$0.21}	&\textbf{99.25$\pm$0.06}	&79.97$\pm$0.48	&78.49$\pm$0.26  \\ \midrule

 DUPLEX$^\ddagger$ &87.43$\pm$0.20 &91.16$\pm$0.24 &88.43$\pm$0.16 &91.74$\pm$0.38 &84.53$\pm$0.34 &81.36$\pm$0.46 \\
 MLP$^\ddagger$ &84.00$\pm$0.29	&\textbf{91.52$\pm$0.25}	&\textbf{90.83$\pm$0.16}	&\textbf{96.48$\pm$0.28}	&72.93$\pm$0.21	&71.51$\pm$0.20  \\
  \bottomrule

\end{tabular}
\label{tb:duplex}
\end{table*}

\begin{table*}[t]
\centering
\caption{Link prediction results for \textbf{CiteSeer} dataset under the DUPLEX~\cite{duplex} setup: results without superscripts are from the DUPLEX paper, $^\dagger$ indicates reproduction with test set edges in training, and $^\ddagger$ indicates reproduction without test set edges in training.}
\begin{tabular}{@{}lllllll@{}}
\toprule

Method &EP(ACC) &EP(AUC) &DP(ACC) &DP(AUC) &3C(ACC) &4C(ACC) \\ \midrule

MagNet &80.7$\pm$0.8 &88.3$\pm$0.4 &91.7$\pm$0.9 &96.4$\pm$0.6 &72.0$\pm$0.9 &69.3$\pm$0.4 \\ 
DUPLEX &95.7$\pm$0.5 &98.6$\pm$0.4 &98.7$\pm$0.4 &99.7$\pm$0.2 &94.8$\pm$0.2 &91.1$\pm$1.0  \\ \midrule

DUPLEX$^\dagger$ &92.11$\pm$0.78 &95.85$\pm$0.87 &97.54$\pm$0.54 &98.93$\pm$0.59  &88.22$\pm$1.06 &84.77$\pm$1.01 \\
MLP$^\dagger$ &85.74$\pm$1.80	&93.33$\pm$1.27	&\textbf{97.55$\pm$0.97}	&\textbf{99.58$\pm$0.52}	&81.20$\pm$0.82	&76.30$\pm$0.62  \\ \midrule 

DUPLEX$^\ddagger$  &83.59$\pm$1.47 &89.34$\pm$1.03 &85.56$\pm$1.36 &91.82$\pm$0.98 &76.37$\pm$2.07 &73.80$\pm$2.01 \\
MLP$^\ddagger$ &77.34$\pm$1.86	&87.36$\pm$1.26	&\textbf{89.19$\pm$0.95}	&\textbf{95.97$\pm$0.61}	&67.82$\pm$1.21	&64.25$\pm$1.35 \\ \bottomrule

\end{tabular}
\vspace{-3mm}
\label{tb:app_duplex}
\end{table*}

\subsection{Issues with Existing Experimental Setups}\label{issues}
The existing experimental setups for directed link prediction are generally divided into two categories. The first is the \textbf{multiple subtask setup}, which includes tasks such as existence prediction (EP), direction prediction (DP), three-class prediction (3C), and four-class prediction (4C). This approach treats directed link prediction as a multi-class classification problem, where the model must predict edges as positive (original direction), inverse (reverse direction), bidirectional (both directions), or nonexistent (no connection). Specifically:
\begin{itemize}
    \item EP: The model predicts whether a directed edge $(u, v)$ exists in the graph, treating both reverse and nonexistent edges as nonexistent.
    \item DP: The model predicts the direction of edges for node pairs $(u, v)$, where either $(u, v) \in E$ or $(v, u) \in E$.
    \item 3C: The model classifies an edge as positive, reverse, or nonexistent.
    \item 4C: The model classifies edges into four classes: positive, reverse, bidirectional, or nonexistent. 
\end{itemize}
The multiple subtask setup is widely adopted by existing GNN methods~\cite{magnet,dpyg,fiorini2023sigmanet,lin2023magnetic,duplex,lightdic}.
The second category is the \textbf{non-standardized setup} defined in various papers~\cite{strap,odin,dhypr,digae,coba}. These setups involve different datasets, inconsistent splitting strategies, and varying evaluation metrics. Below, we discuss the four significant issues with existing setups.

\textbf{Issue 1: The Multi-layer Perceptron (MLP) is a neglected but powerful baseline.} 
Most existing setups fail to report the performance of MLP. We evaluate MLP across three popular multiple subtask setups: MagNet~\cite{magnet}, PyGSD~\cite{dpyg}, and DUPLEX~\cite{duplex}, which cover a variety of datasets and baselines. Figures~\ref{fig:dp} and~\ref{fig:ep} present the results of our reproduced MagNet experiments alongside the MLP performance, showing that the MLP performs comparably to MagNet on the DP and EP tasks. Tables~\ref{tb_dpyg_dp} and~\ref{tb_dpyg_ep} display MLP results alongside several other baselines on the DP and EF tasks within the PyGSD setup, with the highest values highlighted in bold. Across six datasets, MLP demonstrates competitive performance, achieving state-of-the-art (SOTA) results for both tasks on the Texas and Wisconsin datasets. Tables~\ref{tb:duplex} and~\ref{tb:app_duplex} present the replicated DUPLEX experiments and MLP results on the Cora and CiteSeer datasets, showing that MLP achieves competitive performance. Interestingly, despite DUPLEX being a recent advancement in directed graph learning, MLP outperforms it on certain tasks. These findings highlight the lack of fundamental baselines in previous studies and suggest that current benchmarks do not provide a sufficient challenge. More importantly, the results indicate that MLP has achieved SOTA performance across various setups and datasets, challenging the conclusions of prior studies and contradicting the theoretical assumption that dual methods are more expressive.

\begin{table*}[t]
\centering
\vspace{-1mm}
\caption{Statistics of DirLinkBench datasets.}
\resizebox{\textwidth}{!}{
\begin{tabular}{lrrrrcr}
\toprule
Datasets & \#Nodes & \#Edges &Avg. Degree & \#Features & \%Directed Edges &Description \\ \midrule
Cora-ML & 2,810 & 8,229 & 5.9 &2,879 & 93.97 & citation network \\

CiteSeer & 2,110 & 3,705 & 3.5 & 3,703 & 98.00 & citation network \\
Photo & 7,487 & 143,590 & 38.4 & 745 & 65.81 & co-purchasing network \\
Computers & 13,381 & 287,076 & 42.9 & 767 & 71.23 & co-purchasing network \\
WikiCS & 11,311 & 290,447 & 51.3 & 300 & 48.43 & weblink network \\
Slashdot & 74,444 & 424,557 & 11.4 &- & 80.17 & social network \\
Epinions & 100,751 & 708,715 & 14.1 &- & 65.04 & social network \\
\bottomrule
\end{tabular}}
\vspace{-2mm}
\label{dataset_info}
\end{table*}

\begin{figure}[t]
    \centering
   \vspace{-3mm}   
   \hspace{-4.8mm}
   \subfigure[Cora]{
   \includegraphics[width=40mm]{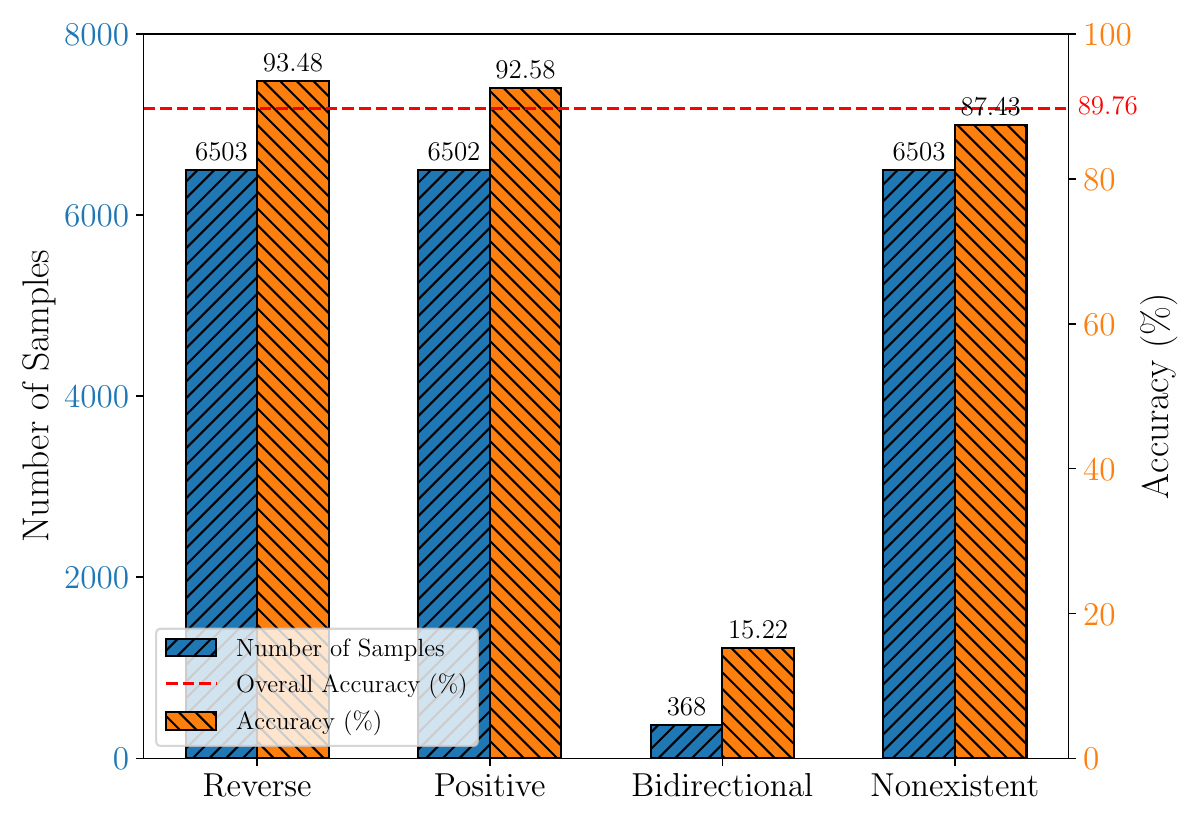}
   \label{fig:duplex_cls_cora}
   }
   \hspace{-3.5mm}
   \subfigure[CiteSeer]{
   \includegraphics[width=40mm]{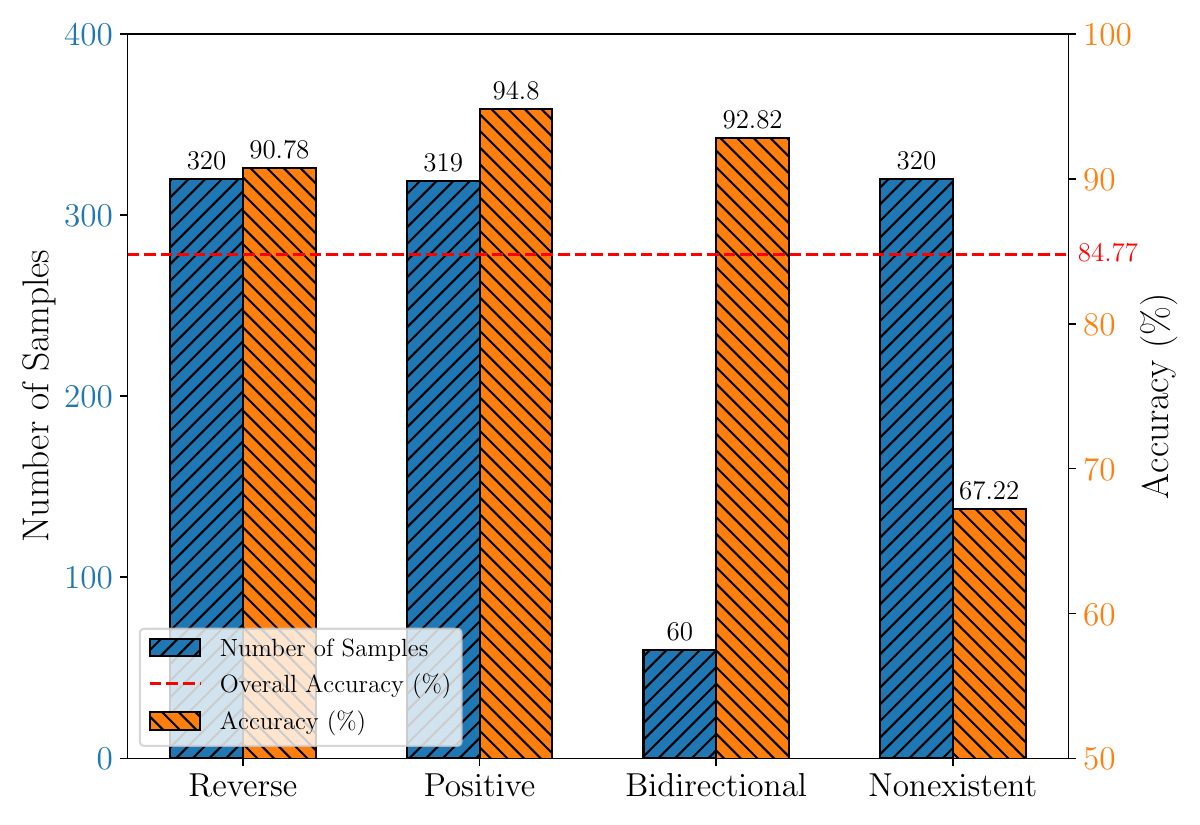}
   \label{fig:duplex_cls_citeseer}
   }
   \hspace{-1.8mm}
   \caption{The number of samples and the accuracy for each class of DUPLEX~\cite{duplex} on the Cora and CiteSeer dataset in the 4C task.}
   \vspace{-1mm}
   \label{fig:duplex_balance}
\end{figure}

\textbf{Issue 2: Many benchmarks suffer from label leakage.}
As defined in Definition~\ref{de_dlp}, directed link prediction aims to predict potential edges from observed graphs, with the key principle that test edges must remain hidden during training to avoid label leakage. However, some current setups violate this principle. For example, 
(1) MagNet~\cite{magnet}, PyGSD~\cite{dpyg}, and DUPLEX~\cite{duplex} expose test edges during negative edge sampling in the training process, indirectly revealing the test edges' presence to the model.
(2) LighDiC~\cite{lightdic} uses eigenvectors of the Laplacian matrix of the entire graph as input features, embedding test edge information in the training input. 
(3) DUPLEX propagates information across the entire graph during training, making the test edges directly visible to the model. 
To investigate, we experiment with DUPLEX using its original code. As shown in Tables~\ref{tb:duplex} and~\ref{tb:app_duplex}, DUPLEX$^\dagger$ (original settings with label leakage) clearly outperforms DUPLEX$^\ddagger$ (propagation restricted to training edges) due to label leakage. The similar results are observed with MLP: MLP$^\dagger$ (using in/out degrees from test edges) significantly outperforms MLP$^\ddagger$ (using only training-edge in/out degrees). These findings underscore that even the leakage of degree information can significantly impact performance.

\textbf{Issue 3: Multiple subtask setups result in class imbalances and limited evaluation metrics.}
The multiple subtask setups treat the directed link prediction task as a multi-class classification problem, causing significant class imbalances that hinder model training. For example, the 4C task in DUPLEX classifies edges into reverse, positive, bidirectional, and nonexistent. However, bidirectional edges are rare in real-world directed graphs, and reverse edges are often assigned arbitrarily, lacking meaningful semantic interpretation.
Figures~\ref{fig:duplex_cls_cora} and~\ref{fig:duplex_cls_citeseer} illustrate this imbalance, highlighting the difficulty of accurately predicting bidirectional and nonexistent edges on the Cora and CiteSeer datasets.
Additionally, these setups rely heavily on accuracy as the evaluation metric, which provides a limited and potentially misleading
assessment of model performance. Given the nature of link prediction, ranking-based metrics such as Hits@K and Mean Reciprocal Rank (MRR) are more appropriate, a perspective well-established in evaluating undirected link prediction methods~\cite{li2023evaluating}.

\textbf{Issue 4: Lack of standardization in dataset splits and feature inputs.}
Current settings face inconsistent dataset splits. In multiple subtask setups, edges are typically split into 80\% for training, 5\% for validation, and 15\% for testing. However, class proportions are further manually adjusted for balance, which leads to varying training and testing ratios across different datasets. Non-standardized setups are more confusing, for example, ELTRA~\cite{eltra} uses 90\% of edges for training, STRAP~\cite{strap} uses 50\%, and DiGAE~\cite{digae} uses 85\%, making cross-study results difficult to evaluate.
Feature input standards are also lacking. Embedding methods often omit node features, while GNNs require them. MagNet and PyGSD use in/out degrees, DUPLEX uses random normal distributions, LightDiC uses original features or Laplacian eigenvectors, and DHYPR~\cite{dhypr} uses identity matrices. This inconsistency undermines reproducibility. As shown in Tables~\ref{tb_dpyg_dp}, \ref{tb_dpyg_ep}, \ref{tb:duplex}, \ref{tb:app_duplex}, and Figure~\ref{fig:magnet_mlp}, reproduced results often deviate significantly, with some better and others worse.


In the above experiments, we strictly follow the configurations of each setup and reproduce the results using the provided codes and datasets. For the MLP model, we implement a simple two-layer network with 64 hidden units, tuning the learning rate and weight decay to match each setup for a fair comparison. These issues highlight the need for a new benchmark in directed link prediction, enabling fair evaluation and supporting future research.

\section{New Benchmark: DirLinkBench}
In this section, we introduce a new robust benchmark for the directed link prediction tasks, \textbf{DirLinkBench}, which offers three key advantages:
\begin{itemize}
\item \textbf{Comprehensive coverage}. DirLinkBench includes seven real-world datasets, 16 baselines, and seven evaluation metrics. These datasets span diverse domains, scales, and structural properties, and are uniformly preprocessed to support consistent evaluation. The baselines cover both embedding methods and GNNs under fair settings. Notably, to the best of our knowledge, DirLinkBench is the first to introduce ranking-based metrics for evaluating directed link prediction.
\item \textbf{Standardized evaluation}. DirLinkBench addresses the issues of existing benchmarks discussed in Section~\ref{issues} by redesigning the task setup. It establishes a unified framework for dataset splitting, feature initialization, and evaluation metrics, ensuring fairness, consistency, and reproducibility across models.
\item \textbf{Modular extensibility}. Built on PyTorch Geometric (PyG)~\cite{pyg}, DirLinkBench is highly modular and extensible, facilitating the integration of new datasets, model architectures, and configurable components such as feature initialization strategies and negative sampling schemes.
\end{itemize}
Next, we provide a detailed overview of DirLinkBench, covering its datasets, task setup, baseline methods, and results.

\subsection{Dataset}
DirLinkBench comprises seven real-world directed graphs from diverse domains. Specifically, the datasets include: (1) Two citation networks, Cora-ML~\cite{mccallum2000cora_ml,bojchevski2018cora_ml} and CiteSeer~\cite{sen2008citeseer}, where nodes represent academic papers and directed edges denote citation relationships. (2) Two Amazon co-purchasing networks, Photo and Computers~\cite{shchur2018computerandphoto}, in which nodes denote products and directed edges represent sequential purchase behavior. (3) WikiCS~\cite{mernyei2020wiki}, a weblink network where nodes correspond to computer science articles on Wikipedia and directed edges indicate hyperlinks between articles. (4) Two social networks, Slashdot~\cite{ordozgoiti2020slash} and Epinions~\cite{massa2005epinion}. 
In Slashdot, nodes represent users and directed edges capture explicit social interactions such as friendships or replies, while in Epinions, directed edges encode trust relationships, offering a view into user-to-user reliability assessments.



These datasets are publicly available and have been widely used in tasks such as node classification~\cite{appnp} and link prediction~\cite{dpyg}. However, many of them contain noise, such as duplicate edges, self-loops, and isolated nodes, that can negatively impact the evaluation of link prediction methods. To address this, we preprocess the datasets by removing duplicate edges and self-loops. Following the protocols in~\cite{appnp,shchur2018computerandphoto}, we also eliminate isolated nodes and retain only the largest connected component to ensure standardized evaluation conditions for link prediction. We summarize the statistical characteristics of the datasets in Table~\ref{dataset_info}, where Avg. Degree indicates average node connectivity, and \%Directed Edges reflects inherent directionality.

The datasets not only originate from different domains but also vary significantly in size. For example, Epinions contains over 700,000 edges and more than 100,000 nodes. They also differ in average node degree; CiteSeer has the lowest average degree (3.5), while WikiCS has the highest (51.3). Additionally, all datasets include original node features except Slashdot and Epinions. \textbf{Compared to previous benchmarks, our collection spans multiple domains, dataset sizes, structural characteristics, and standardized pre-processing}, enabling a more comprehensive evaluation of link prediction methods across diverse real-world scenarios.



\begin{table*}[htbp]
    \centering
    \caption{Performance under the \textbf{Hits@100} metric (mean ± standard error, \%). Results ranked \hig{1}{first}, \hig{2}{second}, and \hig{3}{third} are highlighted. ``TO" indicates methods that did not complete within 24 hours, and ``OOM" indicates methods that exceeded memory limits.}
    \label{tab:performance_comparison}
    \vspace{0mm}
    \resizebox{\textwidth}{!}{
    \begin{tabular}{
        lcccccccr 
    }
        \toprule
        {Method} & {Cora-ML} & {CiteSeer} & {Photo} & {Computers} & {WikiCS} & {Slashdot} & {Epinions} & {Avg.\ Rank $\downarrow$} \\
        \midrule
        STRAP & 79.09$\pm$1.57 & 69.32$\pm$1.29 & \hig{1}{69.16$\pm$1.44} & \hig{2}{51.87$\pm$2.07} &\hig{1}{76.27$\pm$0.92} & 31.43$\pm$1.21 & \hig{1}{58.99$\pm$0.82} &\hig{3}{5.57}\\
        
        ODIN & 54.85$\pm$2.53 & 63.95$\pm$2.98 & 14.13$\pm$1.92 & 12.98$\pm$1.47 & 9.83$\pm$0.47 & 34.17$\pm$1.19 & 36.91$\pm$0.47 &13.57 \\
        
        ELTRA &\hig{2}{87.45$\pm$1.48} & 84.97$\pm$1.90 & 20.63$\pm$1.93 & 14.74$\pm$1.55  & 9.88$\pm$0.70 & 33.44$\pm$1.00  & 41.63$\pm$2.53 &9.29 \\ \midrule
        
        MLP        & 60.61$\pm$6.64 & 70.27$\pm$3.40 & 20.91$\pm$4.18 & 17.57$\pm$0.85 & 12.99$\pm$0.68 & 32.97$\pm$0.51 & 44.59$\pm$1.62 &11.43  \\ 
        
        GCN        & 70.15$\pm$3.01 & 80.36$\pm$3.07 & \hig{3}{58.77$\pm$2.96} & \hig{3}{43.77$\pm$1.75} & 38.37$\pm$1.51 & 33.16$\pm$1.22 & 46.10$\pm$1.37 &6.14 \\
        
        GAT        & 79.72$\pm$3.07 & \hig{3}{85.88$\pm$4.98} & 58.06$\pm$4.03 & 40.74$\pm$3.22 & 40.47$\pm$4.10  & 30.16$\pm$3.11 & 43.65$\pm$4.88 &6.29 \\
        
        APPNP      & 86.02$\pm$2.88 & 83.57$\pm$4.90 & 47.51$\pm$2.51 & 32.24$\pm$1.40 & 20.23$\pm$1.72 & 33.76$\pm$1.05 & 41.99$\pm$1.23 &8.14\\
        
        GPRGNN & 86.03$\pm$2.73 & 88.70$\pm$2.96 & 47.60$\pm$5.09 & 38.39$\pm$2.64 & 20.87$\pm$3.15 & 32.61$\pm$1.05 & 41.14$\pm$2.10 &7.43\\ \midrule

        DGCN       & 63.32$\pm$2.59 & 68.97$\pm$3.39 & 51.61$\pm$6.33 & 39.92$\pm$1.94 & 25.91$\pm$4.10 & TO               & TO               &10.86  \\
        
        DiGCN      & 63.21$\pm$5.72 & 70.95$\pm$4.67 & 40.17$\pm$2.38 & 27.51$\pm$1.67 & 25.31$\pm$1.84 & TO               & TO               &12.14\\
        
        DiGCNIB    & 80.57$\pm$3.21 & 85.32$\pm$3.70 & 48.26$\pm$3.98 & 32.44$\pm$1.85 & 28.28$\pm$2.44 & TO               & TO               &9.14  \\
        
        DirGNN     & 76.13$\pm$2.85 & 76.83$\pm$4.24 & 49.15$\pm$3.62 & 35.65$\pm$1.30 & \hig{3}{50.48$\pm$0.85} &\hig{3}{41.74$\pm$1.15} &50.10$\pm$2.06 &6.43 \\ \midrule
        
        MagNet     & 56.54$\pm$2.95 & 65.32$\pm$3.26 & 13.89$\pm$0.32 & 12.85$\pm$0.59 & 10.81$\pm$0.46  & 31.98$\pm$1.06 & 28.01$\pm$1.72 &14.29 \\
        
        DUPLEX &69.00$\pm$2.52	&73.39$\pm$3.42	&17.94$\pm$0.66	&17.90$\pm$0.71	&8.52$\pm$0.60	&18.42$\pm$2.59	&16.50$\pm$4.34  &13.14\\ \midrule
        
        DHYPR & \hig{3}{86.81$\pm$1.60} & \hig{2}{92.32$\pm$3.72} & 20.93$\pm$2.41 &TO &TO &OOM/TO &OOM/TO &11.29\\
        
        DiGAE & 82.06$\pm$2.51 & 83.64$\pm$3.21 & 55.05$\pm$2.36 & 41.55$\pm$1.62 & 29.21$\pm$1.36 &\hig{2}{41.95$\pm$0.93} & \hig{3}{55.14$\pm$1.96} & \hig{2}{4.71}\\ \midrule
        
        SDGAE & \hig{1}{90.37$\pm$1.33} &\hig{1}{93.69$\pm$3.68} &\hig{2}{68.84$\pm$2.35} &\hig{1}{53.79$\pm$1.56} &\hig{2}{54.67$\pm$2.50} &\hig{1}{42.42$\pm$1.15} &\hig{2}{55.91$\pm$1.77} &\hig{1}{1.43}\\
        \bottomrule
        
    \end{tabular}}
    \vspace{-2mm}
\end{table*}

\begin{table*}[htbp]
    \centering
    \caption{Performance across seven different metrics (mean ± standard error, \%) on cora-ml, photo, and slashdot datasets. The results ranked \hig{1}{first} and \hig{2}{second} are highlighted.}
    \label{tab:performance_metric}
    \resizebox{\textwidth}{!}{
    \begin{tabular}{
        llcccccccr 
    }
        \toprule
        Dataset &Method & Hits@20 & Hits@50 & Hits@100 & MRR & AUC & AP & ACC \\
        \toprule

        \multirow{13}{*}{Cora-ML} 

        &STRAP & 67.10$\pm$2.55 & 75.05$\pm$1.66 & 79.09$\pm$1.57 & \hig{1}{30.91$\pm$7.48} & 87.38$\pm$0.83 & 90.46$\pm$0.70 & 78.47$\pm$0.77 \\
        
        &ELTRA &\hig{2}{70.74$\pm$5.47} &\hig{2}{81.93$\pm$2.38} &\hig{2}{87.45$\pm$1.48} & 19.77$\pm$5.22 & 94.83$\pm$0.58 & 95.37$\pm$0.52 & 85.40$\pm$0.45  \\ \cmidrule(l){2-9} 
        
        &MLP & 29.84$\pm$4.83 & 44.28$\pm$5.72 & 60.61$\pm$6.64 & 11.84$\pm$3.40 & 89.93$\pm$2.09 & 88.55$\pm$2.51 & 81.32$\pm$2.41 \\
        
        
        &GAT & 33.42$\pm$8.93 & 58.40$\pm$6.55 & 79.72$\pm$3.07 & 11.07$\pm$3.68 & 94.57$\pm$0.45 & 92.77$\pm$0.42 & 88.95$\pm$0.71 \\

        &APPNP & 55.47$\pm$4.63 & 75.16$\pm$4.09 & 86.02$\pm$2.88 & 22.49$\pm$6.33 &\hig{2}{95.94$\pm$0.57} & \hig{2}{95.82$\pm$0.59} &\hig{2}{89.90$\pm$0.73} \\  \cmidrule(l){2-9} 
        
        &DiGCNIB  & 45.90$\pm$4.97 & 66.62$\pm$3.67 & 80.57$\pm$3.21 & 17.08$\pm$3.90 & 94.67$\pm$0.56 & 94.21$\pm$0.65 & 87.25$\pm$0.58 \\
        
        &DirGNN & 42.48$\pm$7.34 & 59.41$\pm$4.00 & 76.13$\pm$2.85 &  13.34$\pm$5.56  & 93.05$\pm$0.87 & 92.52$\pm$0.79 & 85.83$\pm$0.93  \\ \cmidrule(l){2-9} 
        
        &MagNet     &  29.38$\pm$3.39 & 43.28$\pm$3.76 & 56.54$\pm$2.95 & 11.19$\pm$3.04 & 85.23$\pm$0.84 & 86.06$\pm$0.94 & 77.51$\pm$0.92  \\
        
         &DUPLEX & 21.73$\pm$3.19 & 34.98$\pm$2.37 & 69.00$\pm$2.52 & 7.74$\pm$1.95 & 88.02$\pm$0.95 & 86.62$\pm$1.43 & 82.28$\pm$0.93   \\ \cmidrule(l){2-9} 
         
         &DiGAE & 56.13$\pm$3.80 & 72.23$\pm$2.51 & 82.06$\pm$2.51 & 20.53$\pm$4.21 & 92.56$\pm$0.66 & 93.70$\pm$0.54 & 86.25$\pm$0.80 \\ 
        
         &SDGAE &\hig{1}{70.89$\pm$3.35} & \hig{1}{83.63$\pm$2.15} &\hig{1}{90.37$\pm$1.33} & \hig{2}{28.45$\pm$5.82} & \hig{1}{97.24$\pm$0.34} & \hig{1}{97.21$\pm$0.17} &\hig{1}{91.36$\pm$0.70}  \\ \toprule

         \multirow{12}{*}{Photo} 
        
        &STRAP & \hig{2}{38.54$\pm$5.20} & \hig{2}{55.21$\pm$2.19} & \hig{1}{69.16$\pm$1.44} &\hig{2}{12.08$\pm$3.15} & 98.54$\pm$0.04 & 98.65$\pm$0.05 & 94.61$\pm$0.10 \\
        
        &ELTRA & 7.09$\pm$1.23 & 12.86$\pm$1.58 & 20.63$\pm$1.93 & 2.22$\pm$0.59 & 96.89$\pm$0.06 & 95.84$\pm$0.13 & 91.84$\pm$0.12 \\ \cmidrule(l){2-9} 
        
        &MLP & 8.83$\pm$2.06 & 14.07$\pm$2.87 & 20.91$\pm$4.18 & 3.18$\pm$1.06 & 95.29$\pm$0.37 & 93.60$\pm$1.06 & 88.31$\pm$0.50 \\ 
        
        
        &GAT  & 25.97$\pm$4.22 & 42.85$\pm$4.90 & 58.06$\pm$4.03 & 8.62$\pm$3.07 & \hig{2}{99.13$\pm$0.09} & \hig{2}{98.93$\pm$0.11} & \hig{1}{96.17$\pm$0.20} \\  

        &APPNP  & 22.13$\pm$2.60 & 35.30$\pm$2.04 & 47.51$\pm$2.51 & 6.56$\pm$1.14 & 98.54$\pm$0.09 & 98.26$\pm$0.12 & 94.71$\pm$0.29 \\ \cmidrule(l){2-9} 
        
        &DiGCNIB   & 21.42$\pm$2.77 & 34.97$\pm$2.67 & 48.26$\pm$3.98 & 6.82$\pm$1.53 & 98.67$\pm$0.14 & 98.39$\pm$0.16 & 95.05$\pm$0.27   \\
        
         &DirGNN   & 22.59$\pm$2.77 & 34.65$\pm$3.31 & 49.15$\pm$3.62 &  8.72$\pm$2.08 & 98.76$\pm$0.09 & 98.47$\pm$0.13 & 95.34$\pm$0.17  \\ \cmidrule(l){2-9} 
        
         &MagNet &  5.35$\pm$0.50 & 9.04$\pm$0.52 & 13.89$\pm$0.32 &  1.62$\pm$0.39 & 88.11$\pm$0.21 & 87.48$\pm$0.21 & 80.31$\pm$0.14  \\
         
         &DUPLEX & 7.84$\pm$1.17 & 12.64$\pm$1.22 & 17.94$\pm$0.66 & 2.53$\pm$0.66 & 94.22$\pm$0.76 & 93.37$\pm$0.91 & 87.68$\pm$0.52  \\ \cmidrule(l){2-9} 
         
         &DiGAE& 27.79$\pm$3.85 & 43.32$\pm$3.36 & 55.05$\pm$2.36 & 9.38$\pm$2.47 & 97.98$\pm$0.08 & 97.99$\pm$0.10 & 91.77$\pm$0.18 \\ 
        
         &SDGAE & \hig{1}{40.89$\pm$3.86} & \hig{1}{55.76$\pm$4.08} & \hig{2}{68.84$\pm$2.35} & \hig{1}{14.82$\pm$4.22} &\hig{1}{99.25$\pm$0.05} & \hig{1}{99.16$\pm$0.06} & \hig{2}{96.16$\pm$0.14} \\ \toprule

         \multirow{12}{*}{Slashdot} 
        
        &STRAP & 19.10$\pm$1.06 & 25.39$\pm$1.43 & 31.43$\pm$1.21 & \hig{1}{9.82$\pm$1.82} & 94.74$\pm$0.05 & 95.13$\pm$0.05 & 88.19$\pm$0.08 \\
        
        &ELTRA & 18.02$\pm$2.11 & 26.31$\pm$0.95 & 33.44$\pm$1.00 & 5.53$\pm$1.77 & 94.65$\pm$0.03 & 95.23$\pm$0.04 & 88.11$\pm$0.11  \\ \cmidrule(l){2-9} 

        &MLP   & 14.16$\pm$5.22 & 24.01$\pm$0.79 & 32.97$\pm$0.51 & 4.14$\pm$1.71 & 95.84$\pm$0.07 & 96.21$\pm$0.05 & 89.62$\pm$0.11 \\ 
        
        
        &GAT & 14.82$\pm$2.70 & 22.19$\pm$2.78 & 30.16$\pm$3.11 & 5.11$\pm$1.53 & 96.26$\pm$0.18 & 96.45$\pm$0.20 & 88.01$\pm$0.49 \\ 
        
        &APPNP & 15.00$\pm$5.47 & 24.34$\pm$1.47 & 33.76$\pm$1.05 & 5.86$\pm$2.78 & 96.21$\pm$0.06 & 96.43$\pm$0.05 & 90.07$\pm$0.10 \\ \cmidrule(l){2-9} 
        
        
         &DirGNN & 20.55$\pm$2.85 & 31.20$\pm$1.18 & 41.74$\pm$1.15 & 7.52$\pm$3.24 &\hig{1}{96.95$\pm$0.05} & \hig{1}{97.14$\pm$0.06} &\hig{1}{ 90.65$\pm$0.13}  \\ \cmidrule(l){2-9} 
        
         &MagNet & 12.55$\pm$0.75 & 22.34$\pm$0.42 & 31.98$\pm$1.06 & 2.83$\pm$0.51 & 96.57$\pm$0.09 & 96.69$\pm$0.10 & 90.45$\pm$0.09\\
         
         &DUPLEX &5.67$\pm$1.85	&11.49$\pm$3.36	&18.42$\pm$2.59	&1.81$\pm$0.83	&94.36$\pm$3.25	&94.48$\pm$2.76	&85.42$\pm$3.42  \\ \cmidrule(l){2-9} 
         
         &DiGAE &\hig{1}{23.68$\pm$0.94} &\hig{1}{33.97$\pm$1.06} &\hig{2}{41.95$\pm$0.93} & 5.54$\pm$1.51 & 95.26$\pm$0.29 & 96.13$\pm$0.19 & 85.67$\pm$0.28 \\ 
        
         &SDGAE &\hig{2}{23.57$\pm$2.11} &\hig{2}{33.75$\pm$1.48} & \hig{1}{42.42$\pm$1.15} &\hig{2}{8.41$\pm$3.80} & \hig{2}{96.70$\pm$0.10} & \hig{2}{97.06$\pm$0.08} & \hig{1}{91.05$\pm$0.20} \\ 
    
        \bottomrule
        
    \end{tabular}}
    \vspace{-2mm}
\end{table*}

\subsection{Task setup}
In Section~\ref{issues}, we discussed the issues of existing benchmarks, which mainly stem from their task setup. DirLinkBench addresses these issues through the redesigned task setup. 
First, we argue that \textbf{the commonly used multiple subtask setup is flawed and inadequate for evaluating directed link prediction methods}. In Issue 1, we demonstrate that under this setup, a simple MLP achieves unexpectedly high accuracy (80–90\%) across several datasets, surpassing SOTA methods in some cases. This counterintuitive result highlights a fundamental weakness of the multiple subtask setup—it fails to distinguish between simplistic baselines and specialized approaches. Furthermore, in Issue 3, we show that this setup inherently introduces class imbalance and limits the applicability of ranking-based metrics. DirLinkBench adopts a binary classification task for directed link prediction to address these limitations: given two nodes $u$ and $v$, the task is to predict whether a directed edge exists from $u$ to $v$. Notably, this setup aligns with the directed link prediction problem definition in Definition~\ref{de_dlp}. Moreover, this setting has been widely adopted in prior studies on embedding methods~\cite{strap,eltra}, and its extension to GNNs is straightforward and intuitive.

Second, to address label leakage (Issue 2) and unstandardized data splits (Issue 4), DirLinkBench adopts a standardized task setup. Specifically, given a directed graph $\gG$, we randomly split 15\% of the edges for testing, 5\% for validation, and use the remaining 80\% for training, while ensuring that the training graph $\gG^{\prime}$ remains weakly connected~\cite{dpyg}.  
For testing and validation, we sample an equal number of negative edges under the full graph $\gG$ visible, while for training, only the training graph $\gG^{\prime}$ is accessible. Feature inputs are provided in three forms: (1) original node features, (2) in/out degrees computed from the training graph $\gG^{\prime}$~\cite{magnet}, or (3) random feature vectors sampled from a standard normal distribution~\cite{duplex}. For fair comparisons, we generate 10 fixed splits using random seeds, and all models are evaluated on shared splits to report the mean performance. Each model learns from the training graph and feature inputs to compute the probability $p(u, v)$ of test edges. 


\subsection{Baseline}
We carefully select 15 state-of-the-art baselines, including three embedding methods: STRAP~\cite{strap}, ODIN~\cite{odin}, ELTRA~\cite{eltra}; a basic method MLP; four classic undirected GNNs: GCN~\cite{gcn}, GAT~\cite{gat}, APPNP~\cite{appnp}, GPRGNN~\cite{gprgnn}; four single real-valued methods: DGCN~\cite{dgcn-tong}, DiGCN~\cite{digcn}, DiGCNIB~\cite{digcn}, DirGNN~\cite{dirgnn}; two complex-valued methods: MagNet~\cite{magnet}, DUPLEX~\cite{duplex}; a gravity-inspired method: DHYPR~\cite{dhypr}; and a source-target GNN: DiGAE~\cite{digae}. 
Note that some recent methods, such as CoBA~\cite{coba}, BLADE~\cite{blade}, and NDDGNN~\cite{nddgnn}, are not included due to unavailable code. 

For baseline implementation, we use the original code released by the authors or widely adopted libraries such as PyTorch Geometric (PyG)~\cite{pyg} and PyTorch Geometric Signed Directed (PyGSD)~\cite{dpyg}. For methods without publicly available link prediction code (e.g., GCN, DGCN), we implement a variety of decoders and loss functions. For methods with official link prediction code (e.g., MagNet, DiGAE), we strictly follow the authors' reported settings. Hyperparameters are tuned via grid search, adhering to the configurations specified in each paper. Specifically, for STRAP, ODIN, and ELTRA, we consider four decoder options:  \(\sigma\bigl(\vs_u^{\top}\vt_v\bigr)\), \(\mathrm{LR}\bigl(\vs_u \odot \vt_v\bigr)\), \(\mathrm{LR}\bigl(\vs_u \|\vt_v\bigr)\), and \(\mathrm{LR}\bigl(\vs_u \| \vs_v \| \vt_u \| \vt_v\bigr)\)~\cite{odin}. Embedding generation for these methods also follows the parameter settings provided in their respective papers.
For MLP, GCN, GAT, APPNP, GPRGNN, DGCN, DiGCN, DiGCN-IB, and DirGNN, we consider two commonly used loss functions: cross-entropy (CE) loss~\cite{magnet,duplex} and binary cross-entropy (BCE) loss~\cite{digae,dhypr}, and three decoders: $\mathrm{MLP}(\vh_u \| \vh_v)$, $\mathrm{MLP}(\vh_u \odot \vh_v)$, and $\sigma(\vh_u^{\top} \vh_v)$. For MagNet, DUPLEX, DHYPR, and DiGAE, we adopt the loss functions and decoders as reported in their original implementations. For all methods, we train for up to 2000 epochs with an early stopping criterion of 200 epochs. We repeat each experiment 10 times and report the mean and standard deviation. More detailed hyperparameters for each baseline are provided in the Appendix.

\subsection{Result}
For evaluation, we introduce ranking-based metrics to directed link prediction benchmarks for the first time, including Hits@20, Hits@50, Hits@100, and MRR, which are widely used in undirected link prediction benchmarks~\cite{ogb,li2023evaluating}. In addition, we report AUC, AP, and ACC for comparative purposes. A detailed description of all metrics is provided in the Appendix. For each method, we report the best mean result across different combinations of feature inputs, loss functions, and decoders. Table~\ref{tab:performance_comparison} shows the results under the Hits@100 metric, while Table~\ref{tab:performance_metric} shows the results across all metrics for some datasets. Complete results for all metrics are included in the supplemental material due to space constraints.

The results reveal that ACC, AUC, and AP offer limited discriminatory ability across different baselines. For example, some simple undirected GNNs (e.g., GAT, APPNP) achieve competitive performance, with only minor performance gaps across methods. Conversely, methods that perform well under ranking-based metrics (e.g., STRAP, DiGAE) tend to perform relatively poorly. These findings suggest that ACC, AUC, and AP are inadequate for reliably evaluating directed link prediction performance, aligning with recent discussions about their limitations even in undirected settings~\cite{yang2015evaluating,li2023evaluating}. Therefore, we argue that \textbf{ranking-based metrics are better suited for the link prediction task}. Since Hits@100 is widely used in popular benchmarks (e.g., Open Graph Benchmark (OGB)~\cite{ogb}) and reveals significant performance differences among methods, we adopt it as our primary evaluation metric.

From the results in Table~\ref{tab:performance_comparison}, we observe that embedding methods maintain a strong advantage, even without feature inputs. Early single real-valued undirected and directed GNNs also demonstrate competitive performance. In contrast, several newer directed GNNs (e.g., MagNet~\cite{magnet}, DUPLEX~\cite{duplex}, DHYPR~\cite{dhypr}) exhibit weaker performance or face scalability challenges. We provide a deeper analysis of these observations in Section~\ref{analysis}. Interestingly, the simple directed graph autoencoder DiGAE~\cite{digae} is the top-performing method overall, but it underperforms on specific datasets (e.g., Cora-ML, CiteSeer). This finding motivates us to revisit DiGAE’s design and propose improved methods to enhance directed link prediction.

\begin{figure}[t]
    \centering
   \includegraphics[width=75mm]{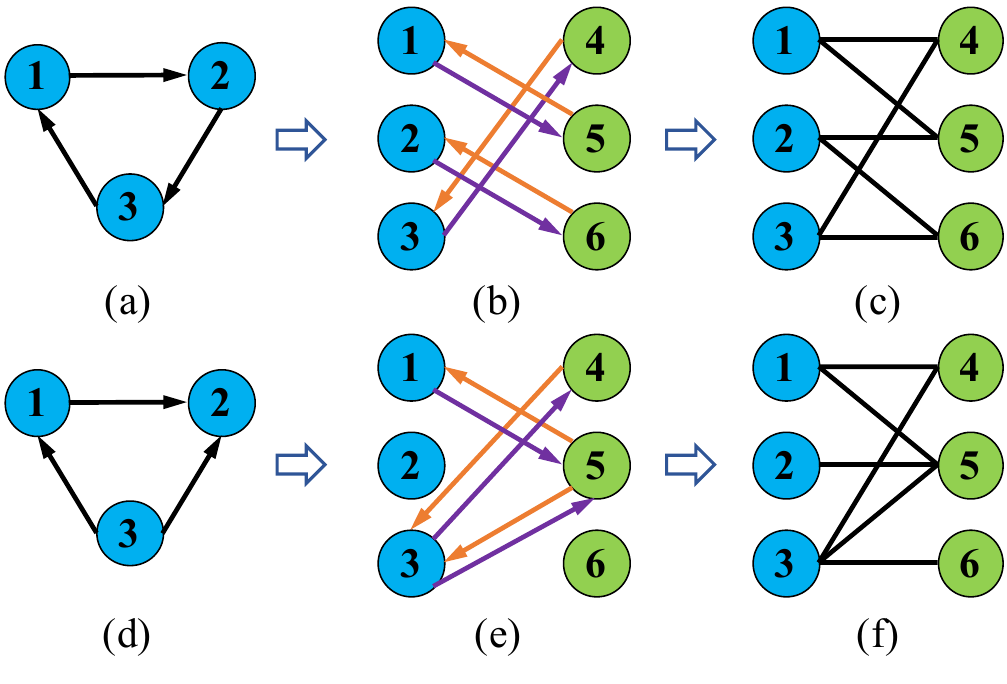}
   \vspace{-2mm}
    \caption{The bipartite graph representation of two toy directed graphs.}
    \label{fig:bg}
    \vspace{-2mm}
\end{figure}

\section{New Method: SDGAE}
In this section, we revisit DiGAE's model architecture to better understand its encoder's graph convolution mechanism. We then introduce a novel method: Spectral Directed Graph Auto-Encoder (SDGAE).

\subsection{Understand the Graph Convolution of DiGAE}
DiGAE~\cite{digae} is a graph auto-encoder designed for directed graphs. Its encoder's graph convolutional layer is denoted as 
\begin{align}
\mS^{(\ell+1)} &= \sigma\left(\hat{\mD}_{\rm out}^{-\beta}\hat{\mA}\hat{\mD}_{\rm in}^{-\alpha}\mT^{(\ell)}\mW_{T}^{(\ell)}\right), \label{digae_s}\\
\mT^{(\ell+1)} &= \sigma\left(\hat{\mD}_{\rm in}^{-\alpha}\hat{\mA}^{\top}\hat{\mD}_{\rm out}^{-\beta}\mS^{(\ell)}\mW_{S}^{(\ell)}\right).\label{digae_t}
\end{align}
Here, $\hat{\mA} = \mA + \mI$ denotes the adjacency matrix with added self-loops, and $\hat{\mD}_{\rm out}$ and $\hat{\mD}_{\rm in}$ represent the corresponding out-degree and in-degree matrices, respectively. 
$\mS^{(\ell)}$ and $\mT^{(\ell)}$ denote the source and target embeddings at the $\ell$-th layer, initialized as $\mS^{(0)} = \mT^{(0)} = \mX$.
The hyperparameters $\alpha$ and $\beta$ are degree-based normalization factors, $\sigma$ is the activation function (e.g., $\mathrm{ReLU}$), and $\mW_{T}^{(\ell)}, \mW_{S}^{(\ell)}$ represents the learnable weight matrices.
The design of this graph convolution is inspired by the connection between GCN~\cite{gcn} and the 1-WL~\cite{wl} algorithm for directed graphs. However, the underlying principles of this convolution remain unexplored,  leaving its meaning unclear. Additionally, the heuristic hyperparameters $\alpha$ and $\beta$ introduce significant challenges for effective parameter optimization.

   

We revisit the graph convolution of DiGAE's encoder and observe that it corresponds to the GCN convolution~\cite{gcn} applied to an undirected bipartite graph. Specifically, given a directed graph $\gG$ with adjacency matrix $\mA$, the adjacency matrix of its bipartite representation~\cite{book_digraph} is defined as:
\begin{equation}
    \mathcal{S}(\hat{\mA}):=
    \left[ 
        \begin{array}{cc}
            \mathbf{0} & \hat{\mA} \\
            \hat{\mA}^{\top} & \mathbf{0}
        \end{array}
    \right] \in \mathbb{R}^{2n \times 2n}.
\end{equation}
Here, $\mathcal{S}(\hat{\mA})$ is a block matrix consisting of $\hat{\mA}$ and $\hat{\mA}^{\top}$. It is evident that $\mathcal{S}(\hat{\mA})$ represents the adjacency matrix of an undirected bipartite graph with $2n$ nodes. Figure~\ref{fig:bg} provides two toy examples of directed graphs and their corresponding undirected bipartite graph representations. Notably, the self-loops in $\hat{\mA}$ serve a fundamentally different purpose than in GCN~\cite{gcn}. In this context, the self-loops ensure the connectivity of the undirected bipartite graph. Without the self-loops, as illustrated in graphs (b) and (e) of Figure~\ref{fig:bg}, the graph structure suffers from significant connectivity issues. Based on these insights, we present the following Lemma~\ref{lemma_digae_conv}.


\begin{lemma}\label{lemma_digae_conv}
If omitting degree-based normalization in Eqs. (\ref{digae_s}) and (\ref{digae_t}), the graph convolution of DiGAE's encoder is
\begin{align}
&\resizebox{0.88\hsize}{!}{$\left[\mS^{(\ell+1)},\mT^{(\ell+1)}\right]^{\top} =\sigma\left(\mathcal{S}(\hat{\mA})\left[\mS^{(\ell)}\mW_{S}^{(\ell)},\mT^{(\ell)}\mW_{T}^{(\ell)}\right]^{\top} \right).$}\label{lemma3_eq}
\end{align}
\end{lemma}
The proof and its extension with degree-based normalization are provided in the Appendix. Lemma~\ref{lemma_digae_conv} and the extension in its proof reveal that \textbf{the graph convolution of DiGAE's encoder essentially corresponds to the GCN convolution applied to an undirected bipartite graph}.


\subsection{Spectral Directed Graph Auto-Encoder (SDGAE)}


Building on the understanding of DiGAE, we identify its three key limitations: (1) DiGAE struggles to use deep networks for capturing rich structural information due to excessive learnable weight matrices, a problem also existing in deep GCN~\cite{pengover}. We provide experimental evidence for this in Section~\ref{compare_sdage}. (2) From the spectral perspective, DiGAE uses a fixed low-pass filter and cannot learn arbitrary spectral graph filters~\cite{hf-lf}, similar to GCN, since both rely on the same convolutional formulation. (3) DiGAE relies on heuristic hyperparameters $\alpha$ and $\beta$, which are difficult to tune effectively.

To overcome these limitations, we propose SDGAE, which uses a polynomial approximation to learn arbitrary graph filters for directed graphs. SDGAE is inspired by spectral-based undirected GNNs~\cite{gprgnn,bernnet} and incorporates symmetric normalization of the directed adjacency matrix. The propagation process of SDGAE's encoder is defined as follows:
\begin{tcolorbox}[height=15mm,boxrule=0.5pt,valign=center,left=0.2mm, right=0.2mm,top=0mm, bottom=1mm]
\begin{equation}\label{eq_sdgae}
    \left[ 
        \begin{array}{c}
            \mS^{(K)} \\
            \mT^{(K)}
        \end{array}
    \right] = \sum\limits_{k=0}^{K}
    \mW^{(k)} 
    \left[ 
        \begin{array}{cc}
            \mathbf{0} & \Tilde{\mA} \\
            \Tilde{\mA}^{\top} & \mathbf{0}
        \end{array}
    \right]^k
    \left[ 
        \begin{array}{c}
            \mS^{(0)} \\
            \mT^{(0)}
        \end{array}
    \right].
\end{equation}
\end{tcolorbox}
Here, $\mW^{(k)} = \mathrm{diag}([w_S^{(k)}\mI_n, w_T^{(k)}\mI_n])\in \mathbb{R}^{2n \times 2n}$ represents the diagonal coefficients matrix and $w^{(k)}_S, w^{(k)}_T \in \mathbb{R}$ are the corresponding polynomial coefficients. The parameter $K \in \mathbb{N}^{+}$ is the polynomial order and $\Tilde{\mA}=\hat{\mD}_{\rm out}^{-1/2}\hat{\mA}\hat{\mD}_{\rm in}^{-1/2}$ denotes the normalization of the directed adjacency matrix $\hat{\mA}$. Below Lemma~\ref{le:norm} (see the Appendix for the proof) shows that this normalization is equivalent to the symmetric normalization of $\mathcal{S}(\hat{\mA})$.
In this Equation, $\mS^{(0)}$ and $\mT^{(0)}$ represent the initial source and target embeddings, while $\mS^{(K)}$ and $\mT^{(K)}$ denote the corresponding embeddings after $K$ propagation steps.

\begin{lemma}\label{le:norm}
    The symmetrically normalized block adjacency matrix $ \mD_{\mathcal{S}}^{-1/2} \mathcal{S}(\hat{\mA}) \mD_{\mathcal{S}}^{-1/2} = \mathcal{S}(\Tilde{\mA})$, where $\mD_{\mathcal{S}} = \mathrm{diag}(\hat{\mD}_{\rm out}, \hat{\mD}_{\rm in})$ is the diagonal degree matrix of \(\mathcal{S}(\hat{\mA})\).
\end{lemma}

\begin{figure}[t]
    \centering
   \vspace{-2mm} 
   \includegraphics[width=90mm]{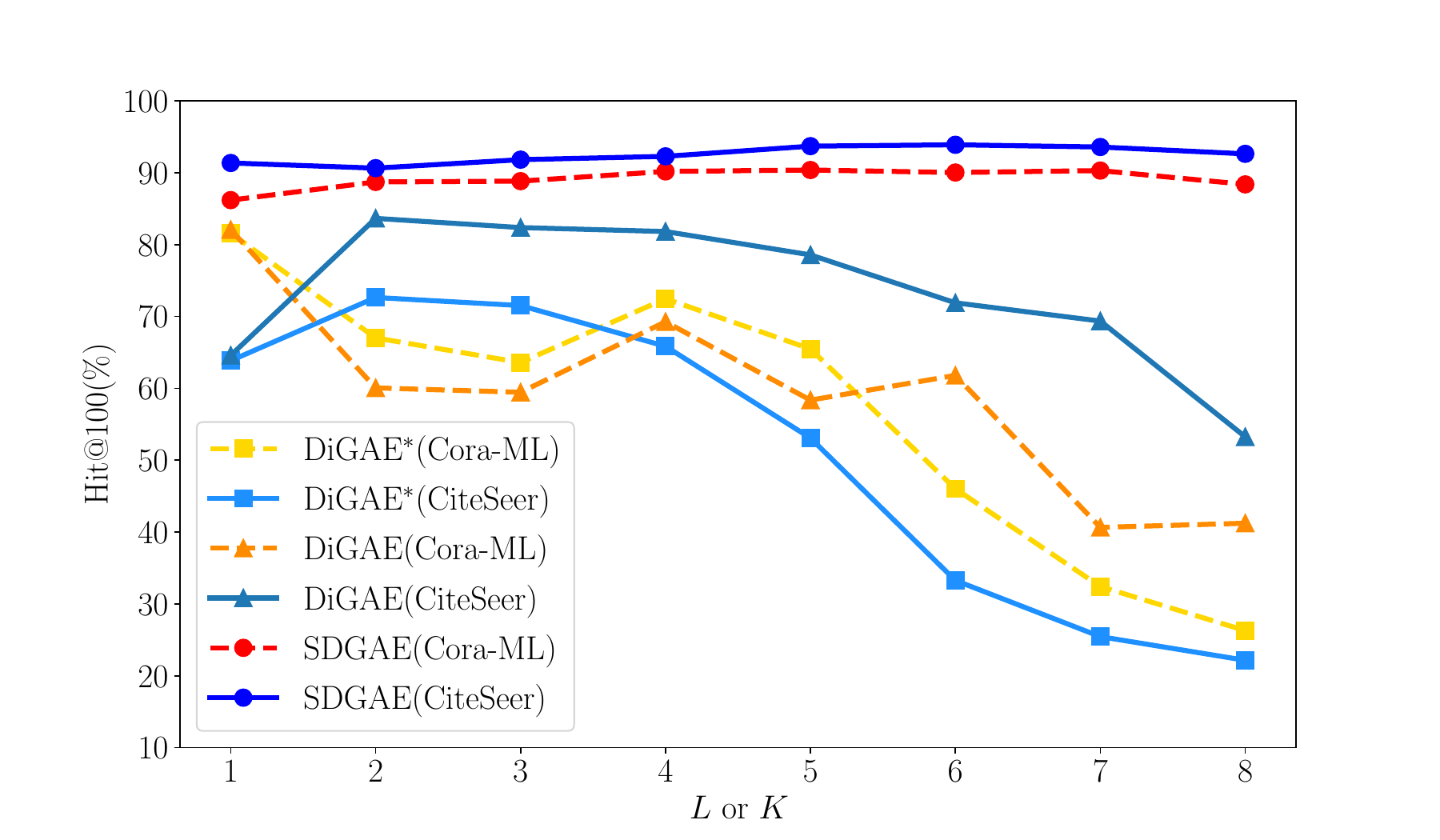}
   \vspace{-2mm}
   \caption{Performance comparison of SDGAE, DiGAE, and DiGAE with residual connections (i.e., DiGAE$^*$) on the Cora-ML and CiteSeer datasets, with varying numbers of convolutional layers or polynomial orders.}
   \vspace{-3mm}
   \label{fig:layer_sdage}
 \end{figure}

\textbf{Spectral Analysis}. 
Intuitively, SDGAE's encoder is a polynomial spectral-based GNN defined on an undirected bipartite graph with $2n$ nodes, where the normalized adjacency matrix is given by $\mathcal{S}(\Tilde{\mA})$. Equation~(\ref{eq_sdgae}) aligns with the propagation process of polynomial spectral-based GNNs presented in Equation~(\ref{spec_gnn}): $\mZ = \sum\nolimits_{k=0}^{K} w_k \mP^k \mX$. Unlike the undirected methods, SDGAE maintains separate source and target embeddings, with corresponding polynomial coefficients $w_S^{(k)}$ and $w_T^{(k)}$.
SDGAE is capable of approximating arbitrary graph filters in the spectral domain. Specifically, we denote the eigendecomposition of $\mathcal{S}(\Tilde{\mA})$ as $\mathcal{S}(\Tilde{\mA}) = \mU_{\mathcal{S}} \mLambda_{\mathcal{S}} \mU_{\mathcal{S}}^{\top}$, where $\mU_{\mathcal{S}}, \mLambda_{\mathcal{S}} \in \mathbb{R}^{2n \times 2n}$ are the eigenvector and diagonal eigenvalue matrices, respectively. The spectral graph filter of SDGAE's encoder is then expressed as:
\begin{equation}
    h(\mathcal{S}(\Tilde{\mA})) = \mU_{\mathcal{S}}\left(\sum\limits_{k=0}^{K}\mW^{(k)} \mLambda_{\mathcal{S}}^{k} \right)\mU_{\mathcal{S}}^{\top}.
\end{equation}
By learning the coefficients $w_S^{(k)}$ and $w_T^{(k)}$, SDGAE can approximate arbitrary graph filters $h(\mathcal{S}(\Tilde{\mA}))$ on directed graphs.

\begin{figure}[t]
    \centering
   \vspace{-2mm}
   \hspace{-8mm}
   \subfigure{
   \includegraphics[width=45mm]{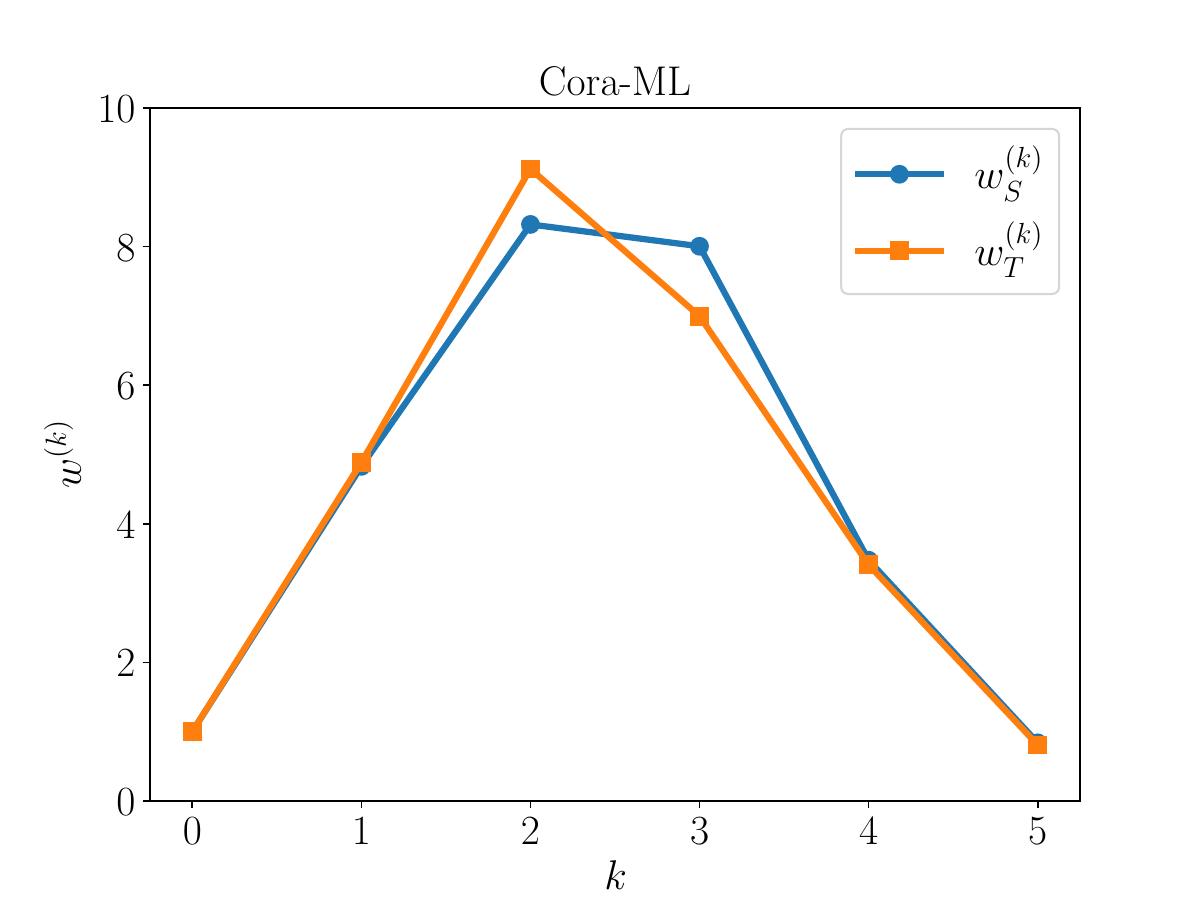}
   }
   \hspace{-8mm}
   \subfigure{
   \includegraphics[width=45mm]{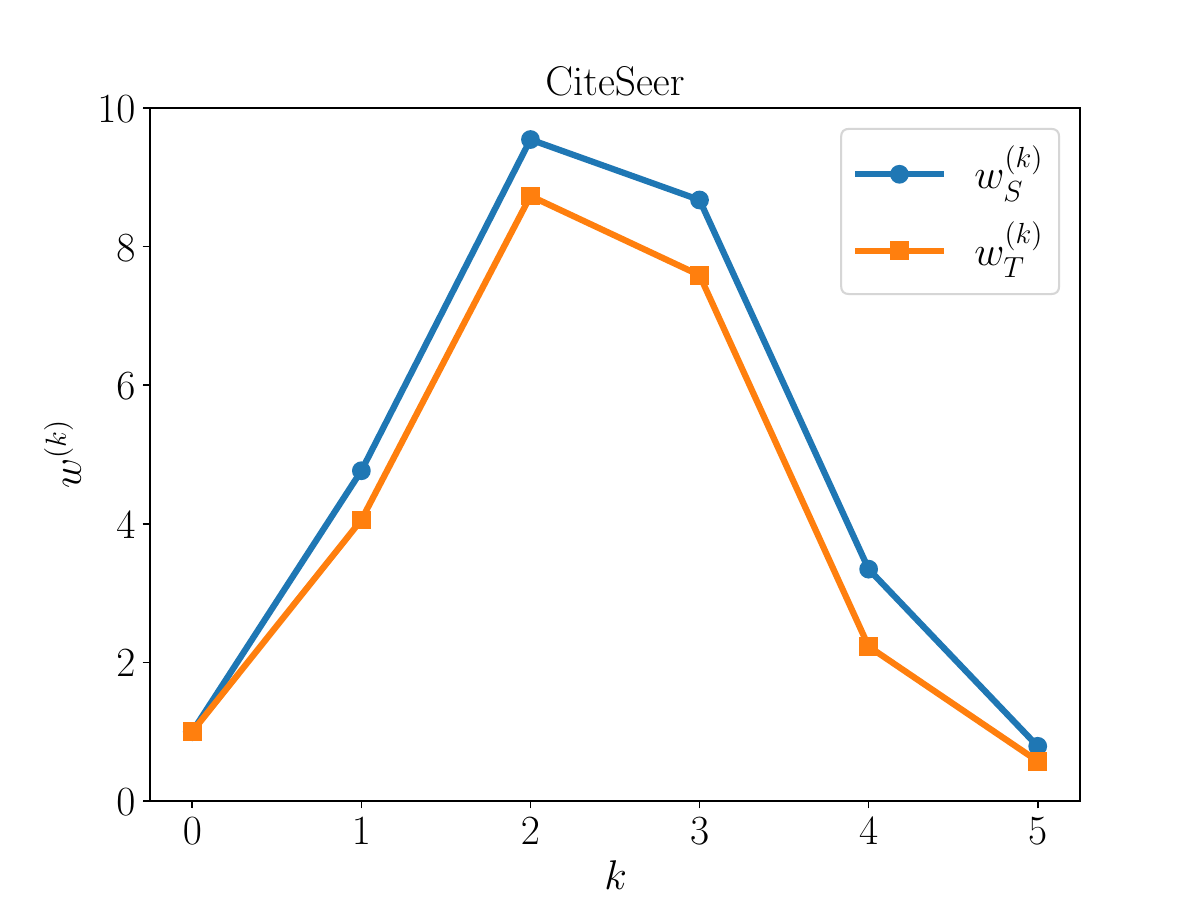}
   }
   \hspace{-8mm}
   \vspace{-1mm}
   \caption{Polynomial coefficients learned by SDGAE on the Cora-ML and CiteSeer datasets with $K = 5$.}
   \label{fig:coe}
   \vspace{-2mm}
\end{figure}

\textbf{Implementation}.  In the implementation of SDGAE, we use two MLPs to initialize the source and target embeddings, i.e., $\mS^{(0)} = \mathrm{MLP}_S(\mX)$ and $\mT^{(0)} = \mathrm{MLP}_T(\mX)$, following the implementation of many spectral-based GNNs~\cite{gprgnn,bernnet}. For polynomial coefficients, we empirically find that directly learning $w_S^{(k)}$ and $w_T^{(k)}$ proves challenging, as different initialization strategies significantly impact the results. One possible explanation is that the coefficients must satisfy convergence constraints in polynomial approximations of filter functions~\cite{chebnetii}. To address this, we instead adopt an iterative formulation that implicitly learns the coefficients through the propagation process:
\begin{align}
   \mS^{(k+1)} &=\gamma_{S}^{(k)}\Tilde{\mA}\mT^{(k)}+\mS^{(k)}, \label{app_eq_sdgae}\\ 
   \mT^{(k+1)} &=\gamma_{T}^{(k)}\Tilde{\mA}^{\top}\mS^{(k)}+\mT^{(k)}. \label{app_eq_sdgae2}
\end{align}
where $\gamma_S^{(k)}$ and $\gamma_T^{(k)}$ are learnable scalar weights, initialized to one. These weights express the polynomial coefficients in Equation (\ref{eq_sdgae}) through the iterative propagation process. For example, when the polynomial order is set to $K = 2$, the output can be expanded as:
\begin{align}
&\resizebox{0.88\hsize}{!}{$
\mS^{(2)} = \mS^{(0)}+ \left(\gamma_{S}^{(1)} + \gamma_{S}^{(0)}\right)\Tilde{\mA}\mT^{(0)} + \gamma_{S}^{(1)}\gamma_{T}^{(0)}\Tilde{\mA}\Tilde{\mA}^{\top}\mS^{(0)},$}\\
&\resizebox{0.88\hsize}{!}{$
\mT^{(2)} = \mT^{(0)} + \left(\gamma_{T}^{(1)} + \gamma_{T}^{(0)}\right)\Tilde{\mA}^{\top}\mS^{(0)} + \gamma_{T}^{(1)}\gamma_{S}^{(0)}\Tilde{\mA}^{\top}\Tilde{\mA}\mT^{(0)}.$}
\end{align}
This expansion means the equivalent polynomial coefficients are: $w_S^{(0)}=1,w_S^{(1)}=\gamma_{S}^{(1)} + \gamma_{S}^{(0)}, w_S^{(2)}=\gamma_{S}^{(1)}\gamma_{T}^{(0)}$ and $w_T^{(0)}=1,w_T^{(1)}=\gamma_{T}^{(1)} + \gamma_{T}^{(0)}, w_T^{(2)}=\gamma_{T}^{(1)}\gamma_{S}^{(0)}$. Notably, the coefficients $w^{(k)}$ tend to decrease at higher orders. This is because the learned values of $\gamma^{(k)}$ generally lie within the range $(0, 1)$, and the product of multiple $\gamma^{(k)}$ terms at higher orders causes $w^{(k)}$ to diminish. The coefficients  $w^{(k)}$ learned by SDGAE on real-world datasets, shown in Figure~\ref{fig:coe}, exhibit this trend. This convergence property contributes to the efficiency of learning polynomial filters.


For the experimental setting of SDGAE, we aligned its configuration with that of other baselines in DirLinkBench to ensure a fair comparison. Regarding the loss function and decoders, SDGAE uses BCE loss with three decoder options: $\sigma(\vs_u \| \vt_v)$, $\mathrm{MLP}(\vs_u \odot \vt_v)$, and $\mathrm{MLP}(\vs_u \| \vt_v)$. For the polynomial order $K$, we search over the set $\{3, 4, 5\}$. The MLP architecture, including the number of layers and hidden units, is consistent with the baseline methods. Detailed hyperparameter settings are provided in the Appendix.



\textbf{Time Complexity}. The computational time complexity of SDGAE primarily stems from its propagation process, as defined in Equation~(\ref{eq_sdgae}). Based on the iterative implementation shown in Equations~(\ref{app_eq_sdgae}) and\~(\ref{app_eq_sdgae2}), the time complexity of SDGAE’s propagation is $O(2Kmd)$, where $m$ is the number of edges, $K$ is the polynomial order, and $d$ is the embedding dimension. This complexity scales linearly with the number of edges $m$. In contrast, DiGAE has a higher time complexity of $O(2Lmd + Lnd^2)$, where $L$ is the number of layers and $n$ is the number of nodes. This is due to the use of multiple learnable weight matrices, i.e., $\mW_S^{(\ell)}$ and $\mW_T^{(\ell)}$ as defined in Equations~(\ref{digae_s}) and~(\ref{digae_t}). So, SDGAE achieves lower time complexity than DiGAE and shows faster training performance.

\textbf{Results}. 
We report the directed link prediction results of SDGAE on DirLinkBench under the Hits@100 metric in Table~\ref {tab:performance_comparison}, and across seven different evaluation metrics in Table~\ref {tab:performance_metric}. Comprehensive results are also provided in the supplemental material.
As shown in Table~\ref{tab:performance_comparison}, SDGAE achieves the best performance on 4 out of the 7 datasets under the Hits@100 metric and delivers competitive results on the remaining three. Moreover, as shown in Table~\ref{tab:performance_metric}, SDGAE ranks among the top two methods across all seven evaluation metrics. Notably, SDGAE significantly outperforms both DiGAE~\cite{digae} and the spectral-based method GPRGNN~\cite{gprgnn}, which is designed for undirected graphs. These results collectively demonstrate the substantial effectiveness of SDGAE for the task of directed link prediction.

 \begin{figure*}[t]
    \centering
   \vspace{-2mm}
   \subfigure[Cora-ML]{
   \label{fig:app_feature_cora}
   \includegraphics[width=65mm]{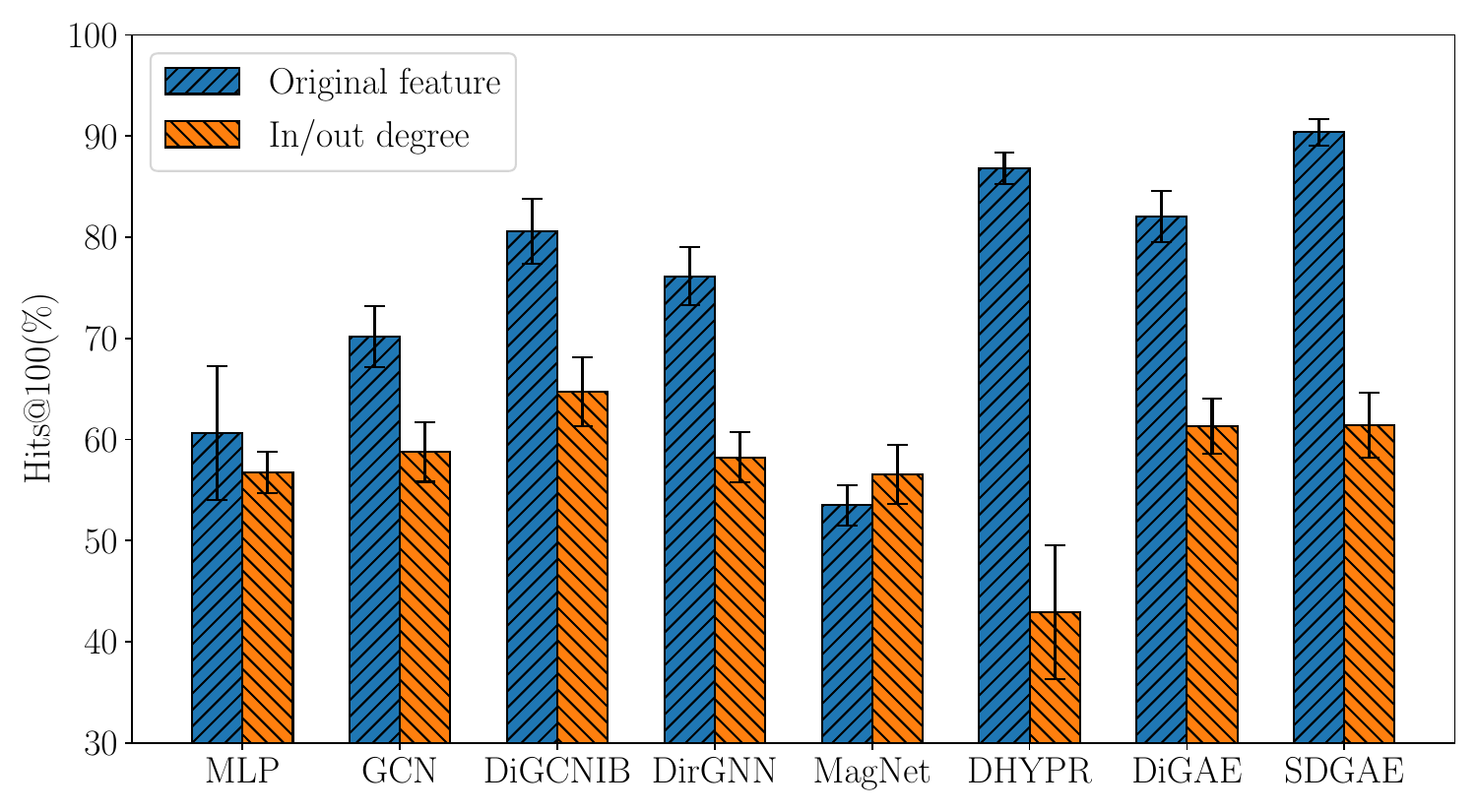}
   }
   \subfigure[Photo]{
   \label{fig:app_feature_photo}
   \includegraphics[width=65mm]{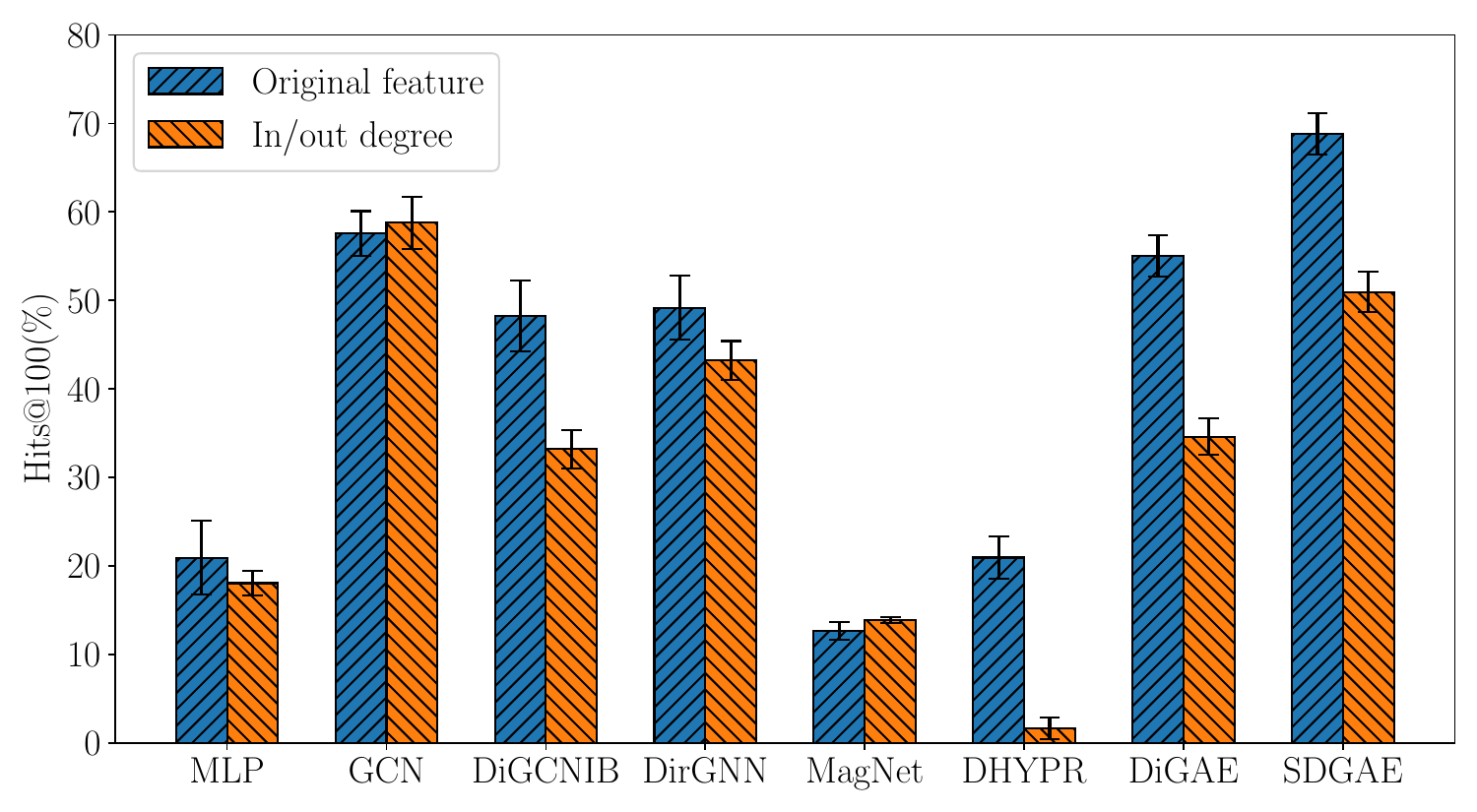}
   }
   
   \subfigure[WikiCS]{
   \label{fig:app_feature_wiki}
   \includegraphics[width=65mm]{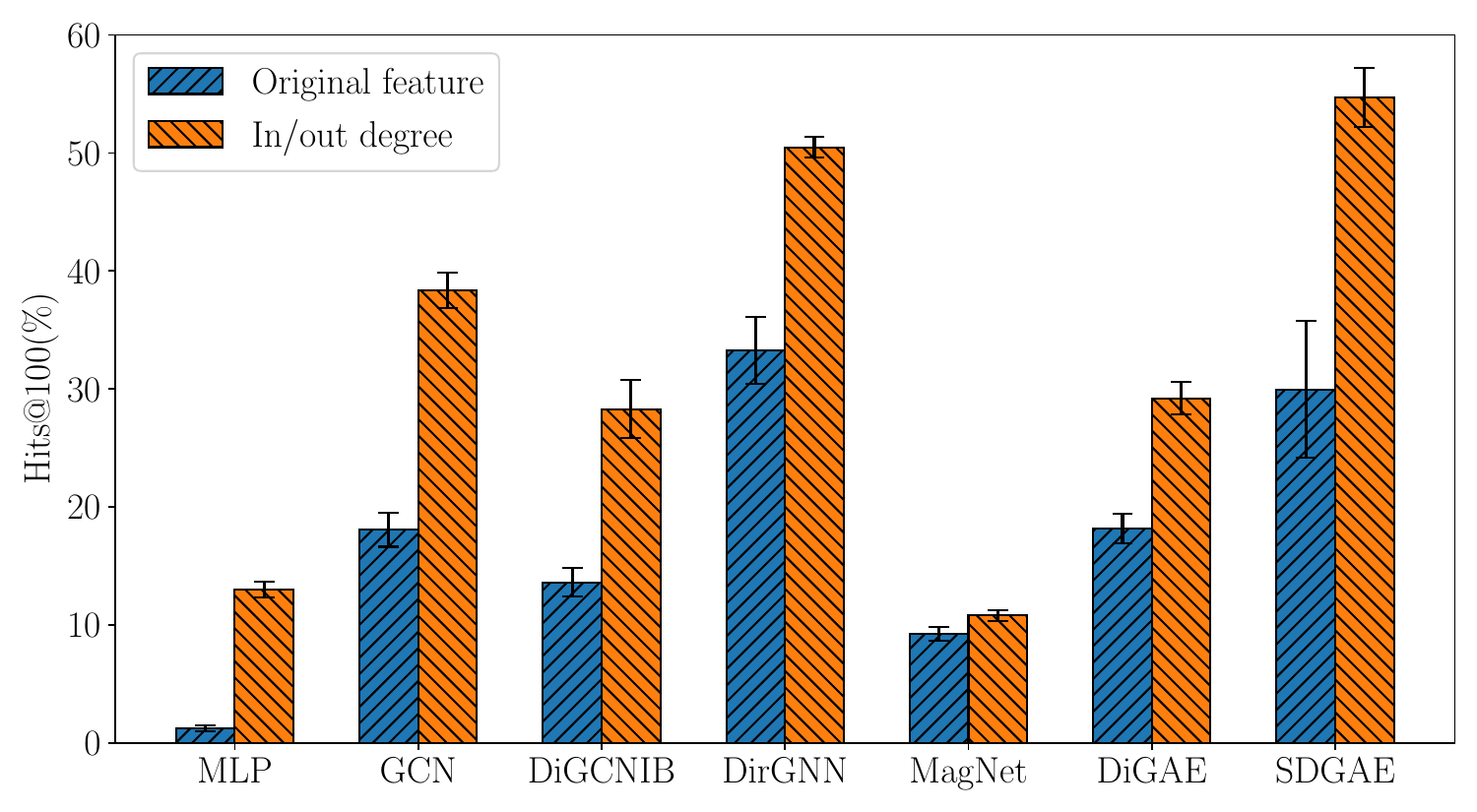}}
   \subfigure[Slashdot]{
   \label{fig:app_feature_slsh}
   \includegraphics[width=65mm]{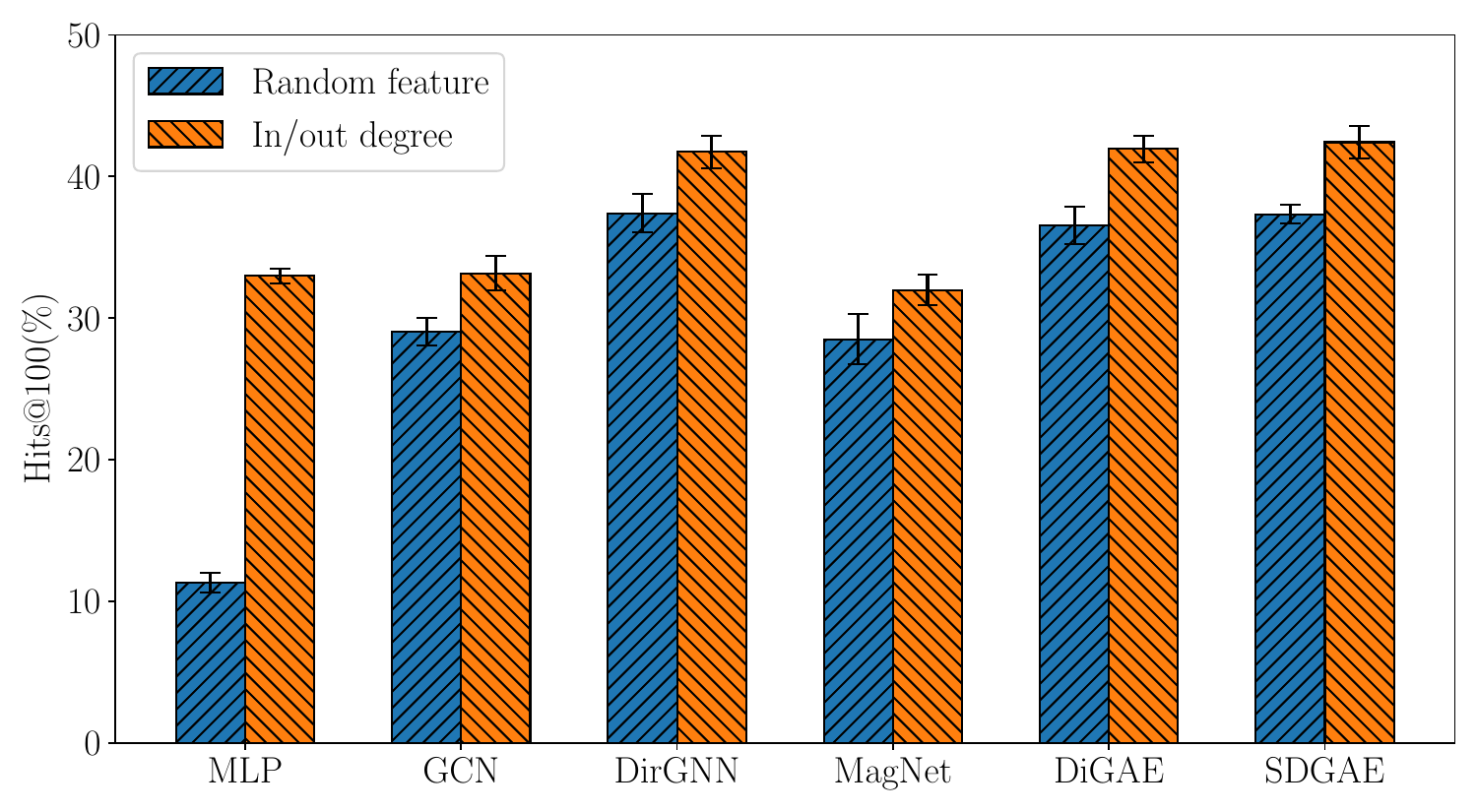}}
   \vspace{-1mm}
   \caption{Performance comparison of various GNN methods using original features, in/out degrees, or random features as input on four datasets.}
    \vspace{-2mm}
   \label{fig:app_feature}
\end{figure*}

\section{Analysis}\label{analysis}
In this section, we first analyze SDGAE with respect to the choice of polynomial order $K$ and the learned polynomial coefficients, aiming to understand the reasons behind its improved performance compared to DiGAE. Next, we conduct a series of experimental investigations on various aspects of directed link prediction methods, including feature inputs, loss functions and decoders, degree distribution, and negative sampling strategies, to offer new insights and highlight open challenges in this field.

\begin{table}[t]
\centering
\caption{Performance of DiGAE and SDGAE under different adjacency matrix normalizations. Best results are highlighted in \textbf{bold}.}
\resizebox{0.48\textwidth}{!}{
\begin{tabular}{lcccc}
\toprule
Methods &Cora-ML	&CiteSeer	&Photo	&Computers  \\ \midrule

DiGAE$_{(\alpha, \beta)}$ & 82.06$_{\pm\text{2.51}}$ & 83.64$_{\pm\text{3.21}}$ & 55.05$_{\pm\text{2.36}}$ & 41.55$_{\pm\text{1.62}}$ \\

DiGAE$_{(-1/2, -1/2)}$ & 70.89$_{\pm\text{3.59}}$ & 71.60$_{\pm\text{6.21}}$ & 38.75$_{\pm\text{5.20}}$ & 32.73$_{\pm\text{5.28}}$ \\

SDGAE$_{(-1/2, -1/2)}$ & \textbf{90.37}$_{\pm\textbf{1.33}}$ & \textbf{93.69}$_{\pm\textbf{3.68}}$ & \textbf{68.84}$_{\pm\textbf{2.35}}$ & \textbf{53.79}$_{\pm\textbf{1.56}}$ \\


\bottomrule
\end{tabular}}
\label{app_norm}
\end{table}

\subsection{Comparative Analysis of SDGAE and DiGAE}\label{compare_sdage}
To investigate the performance differences between SDGAE and DiGAE in utilizing higher neighborhood information, Figure~\ref {fig:layer_sdage} shows how SDGAE’s performance changes with increasing polynomial order $K$, and how the performance of DiGAE and DiGAE$^*$ (DiGAE with residual connections) changes with increasing numbers of convolutional layers $L$. The results on both Cora-ML and CiteSeer datasets show that SDGAE’s performance consistently improves with increasing $K$, achieving optimal results at $K = 5$. This trend aligns with the theoretical advantages of using polynomial filter approximations~\cite{gprgnn}. In contrast, the performance of DiGAE and DiGAE$^*$ declines as the number of layers increases. This is attributed to DiGAE’s essentially GCN-like architecture, which suffers from well-known issues such as over-smoothing and difficulty training deeper networks~\cite{pengover}. Notably, although the iterative formulation of SDGAE in Equations~(\ref{app_eq_sdgae}) and (\ref{app_eq_sdgae2}) may superficially resemble the addition of residual connections, it fundamentally differs by learning a polynomial filter as shown in Equation~(\ref{eq_sdgae}). This fact is also supported by the experiments with DiGAE$^*$, where simply adding residual connections fails to improve DiGAE's performance. In summary, SDGAE effectively uses higher neighborhood information by learning polynomial filter coefficients, enabling it to achieve superior performance as $K$ increases.

We show in Figure~\ref{fig:coe} the polynomial filter coefficients $w_S^{(k)}$ and $w_T^{(k)}$ learned by SDGAE on Cora-ML and CiteSeer. These coefficients are computed from the learned weights $\gamma_S^{(k)}$ and $\gamma_T^{(k)}$. We observe that the coefficients are largest at $k = 2$ and gradually decrease at higher orders (e.g., $k = 3, 4, 5$), which aligns with the expected behavior of polynomial expansions. SDGAE learns distinct sets of coefficients for different datasets, effectively adapting its polynomial filters to the underlying graph structure. This flexibility contrasts with DiGAE, which cannot learn specific filter functions.

\begin{figure}[t]
    \centering
   \vspace{-2mm}
   \hspace{-6mm}
   \subfigure[MagNet]{
   \label{fig:loss_magnet}
   \includegraphics[width=40mm]
   {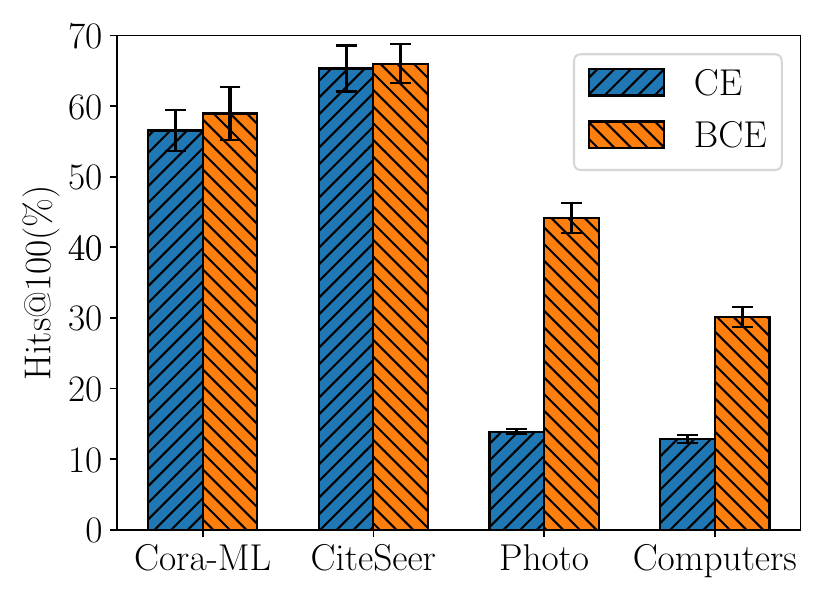}
   }
   \hspace{-3mm}
   \subfigure[DiGAE]{
   \label{fig:loss_digae}
   \includegraphics[width=40mm]{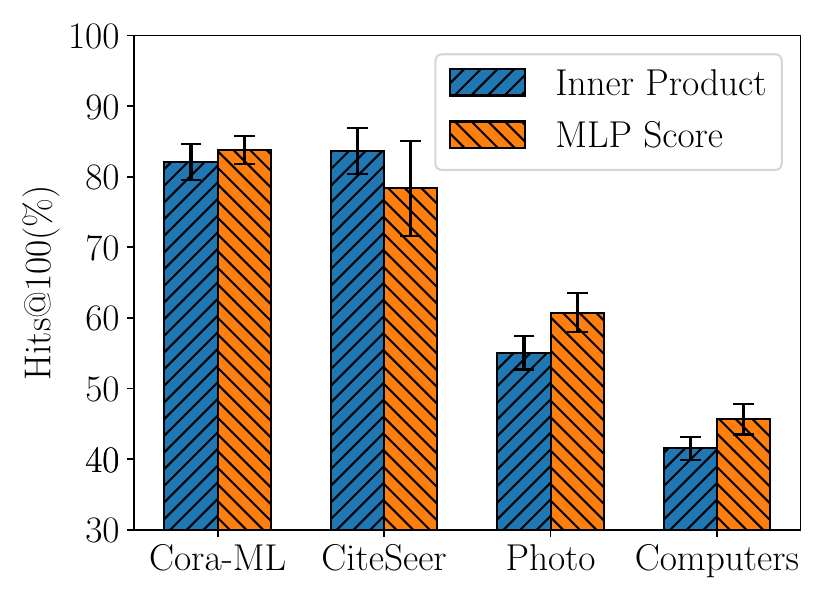}
   }
   \hspace{-6mm}
   \vspace{-0mm}
   \caption{Performance comparison of MagNet and DiGAE with different loss functions and decoders.}
   \label{fig:loss}
   \vspace{-2mm}
\end{figure}

Next, we analyze the impact of adjacency matrix normalization on DiGAE's performance. Table~\ref{app_norm} presents the results of DiGAE and SDGAE under different normalization strategies. Here, DiGAE$_{(\alpha,\beta)}$ refers to the original settings used in the its paper~\cite{digae}, where the adjacency matrix is normalized as $\hat{\mD}_{\rm out}^{-\beta} \hat{\mA} \hat{\mD}_{\rm in}^{-\alpha}$, with $(\alpha, \beta)$ searched over the grid $\{0.0, 0.2, 0.4, 0.6, 0.8\}^2$. Notably, this search space excludes the symmetric normalization $\hat{\mD}_{\rm out}^{-1/2} \hat{\mA} \hat{\mD}_{\rm in}^{-1/2}$, which is used in both DiGAE$_{(-1/2,-1/2)}$ and SDGAE$_{(-1/2,-1/2)}$. The results show that applying symmetric normalization to DiGAE does not improve performance, suggesting that SDGAE's performance gains are not solely attributable to its normalization choice. Furthermore, as shown in Figure~\ref{fig:loss_digae}, modifying DiGAE’s decoder can improve its performance on certain datasets. However, DiGAE fails to achieve results comparable to SDGAE even with these modifications.

These comparative experiments demonstrate that SDGAE significantly outperforms DiGAE, primarily due to its theoretically grounded design based on polynomial graph filters. By learning adaptive polynomial filters, SDGAE can better capture structural patterns across different datasets.


 \begin{figure}[t]
    \centering
   \vspace{-1mm}
   \hspace{-6mm}
    \subfigure[Cora-ML]{
   \includegraphics[width=40mm]{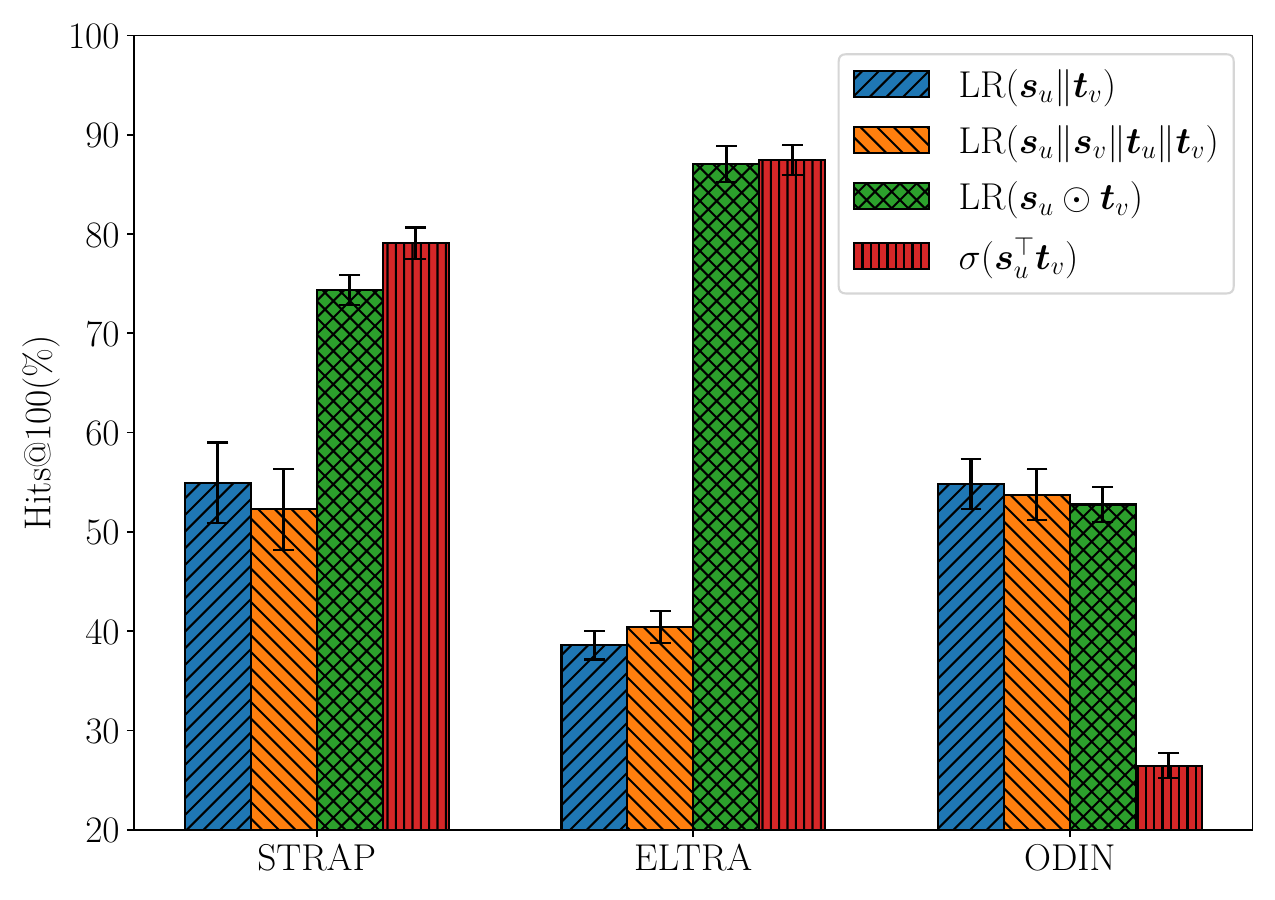}
   }
   \hspace{-2mm}
   \subfigure[Slashdot]{
   \includegraphics[width=40 mm]{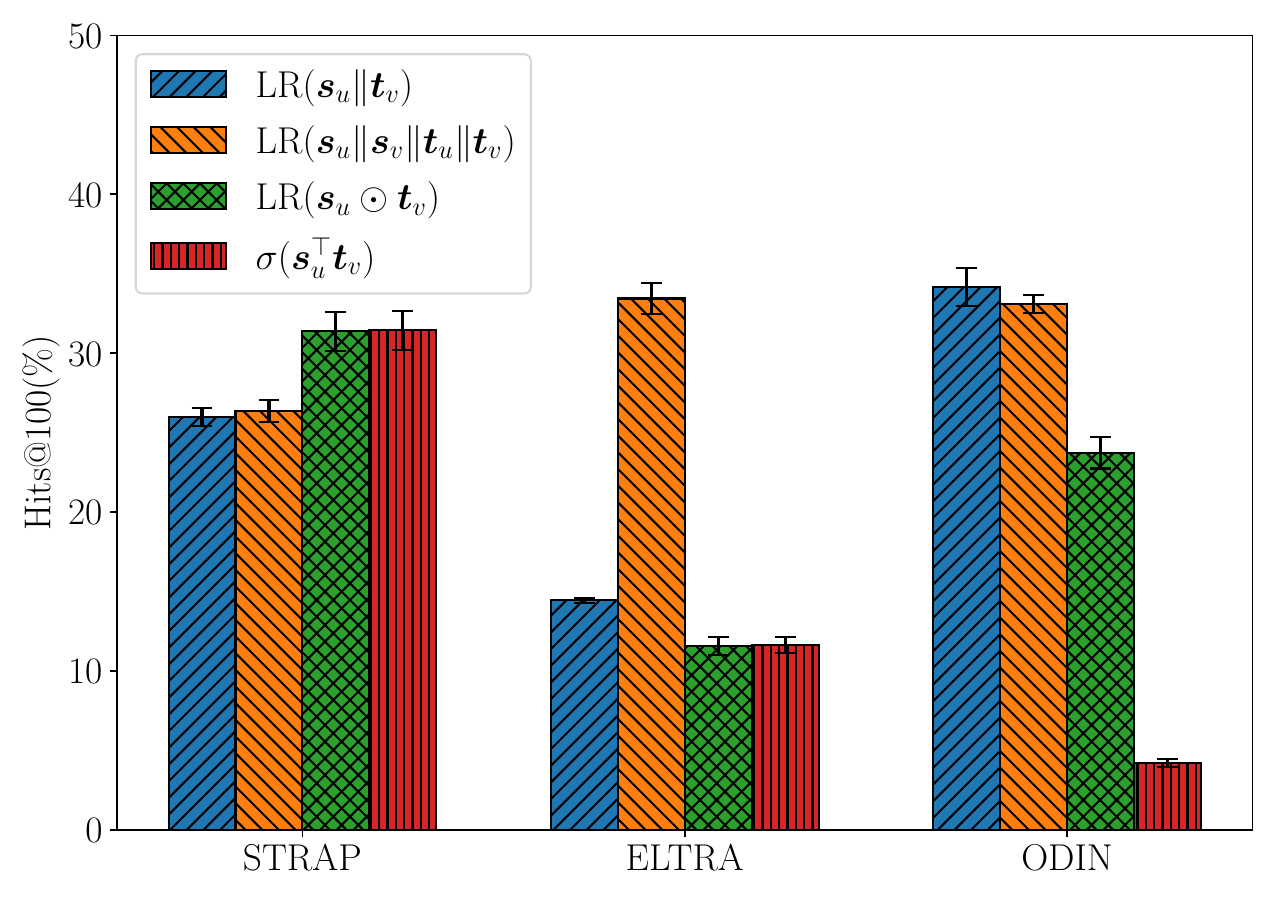}
   }
    \vspace{-0mm}
    \hspace{-6mm}
   \caption{Performance comparison of three embedding methods with four different decoders on the Cora-ML and Slashdot datasets.}
   \vspace{-2mm}
    \label{fig:app_loss_embedding}
 \end{figure}

 \begin{figure}[t]
    \centering
   \vspace{-1mm}
   \hspace{-6mm} 
    \subfigure[Cora-ML]{
   \includegraphics[width=40mm]{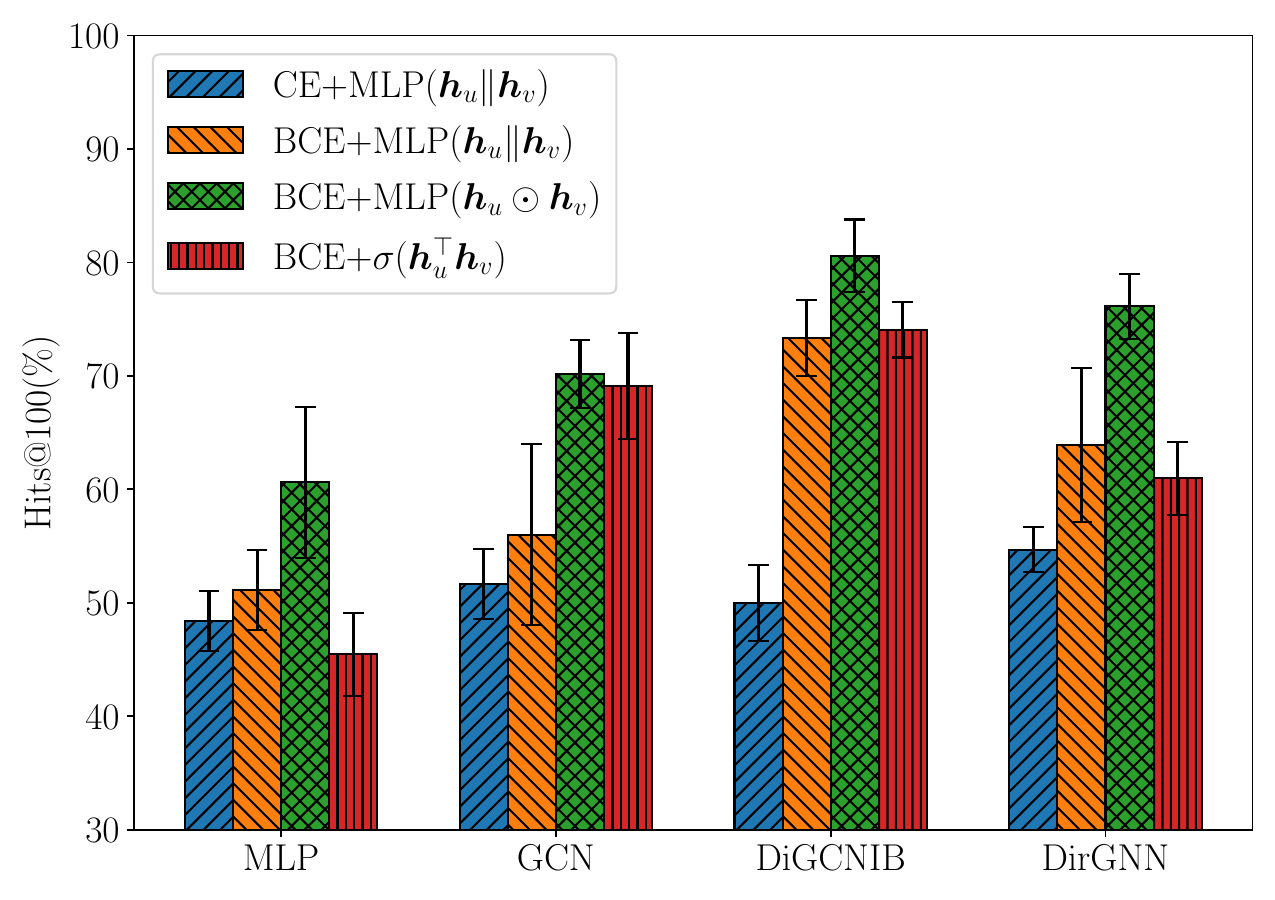}
   }
   \hspace{-3mm}
   \subfigure[Photo]{
   \includegraphics[width=40mm]{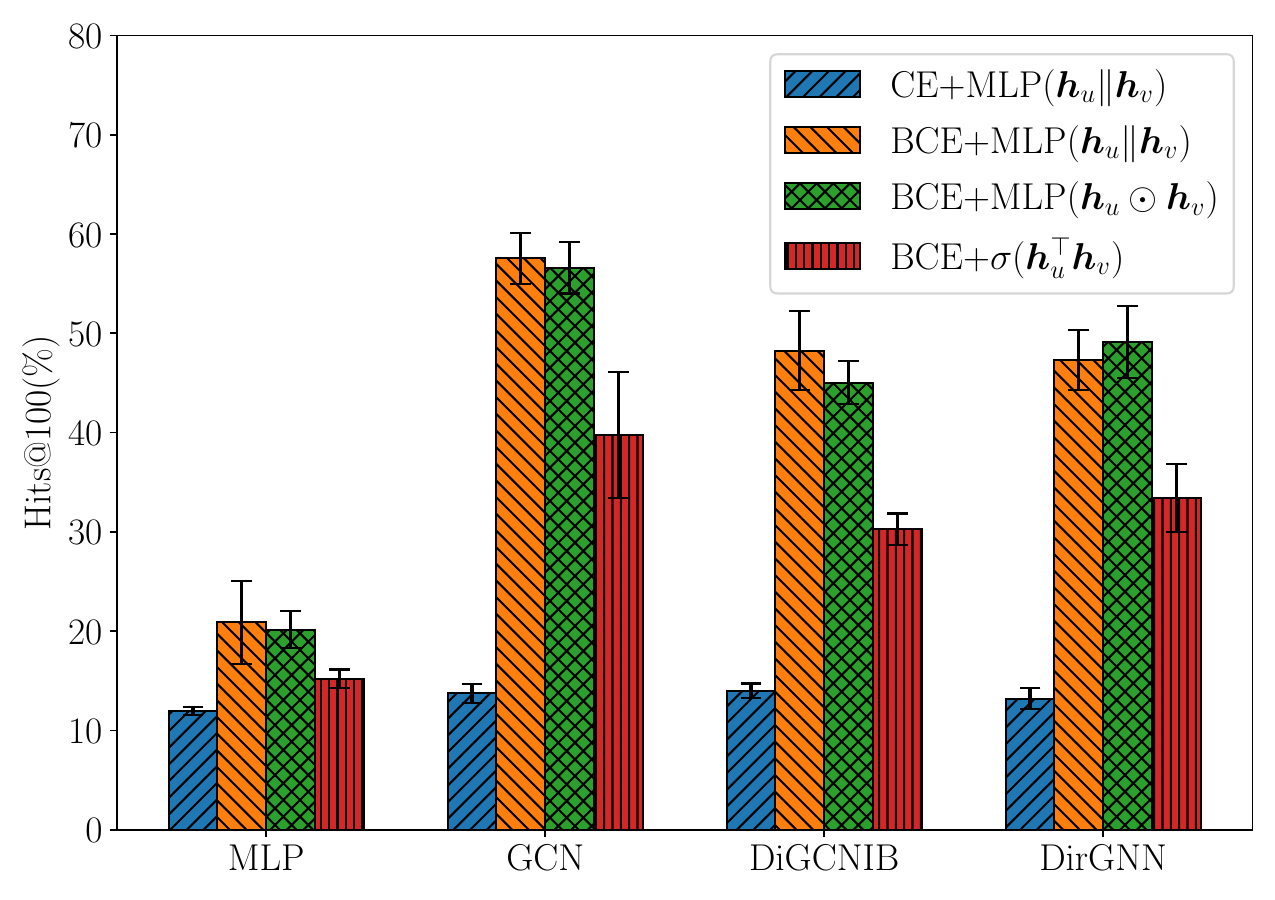}
   }
   \hspace{-6mm}
   \vspace{-0mm}
   \caption{Performance comparison of GNNs with various loss functions and decoders on Cora-ML and Photo datasets.}
   \vspace{-2mm}
    \label{fig:app_loss_gnn}
 \end{figure}


\subsection{Feature Input}
We examine the impact of different feature inputs on GNN performance in directed link prediction. Figure~\ref{fig:app_feature} shows the results of various GNN methods using original features, in/out degrees, or random features as inputs across four datasets. Specifically, Figures~\ref{fig:app_feature_cora}, \ref{fig:app_feature_photo}, and~\ref{fig:app_feature_wiki} compare the performance of GNNs using original features versus in/out degrees on Cora-ML, Photo, and WikiCS, respectively. Figure~\ref{fig:app_feature_slsh} presents a similar comparison using in/out degrees and random features on Slashdot, which lacks original node features.

The results indicate that original features enhance GNN performance on certain datasets (e.g., Cora-ML and Photo). However, for datasets like WikiCS, in/out degrees are more effective. For datasets without original features (e.g., Slashdot and Epinions), in/out degrees significantly outperform random features. \textbf{These findings highlight the critical role of appropriate feature inputs in improving GNN performance on directed link prediction tasks}. Enhancing feature quality remains an important direction for future research, particularly for datasets with weak or missing original features.

\begin{figure*}[htbp]
\centering
   \vspace{-2mm}
   \hspace{-2mm}
   \subfigure{
   \includegraphics[width=37mm]{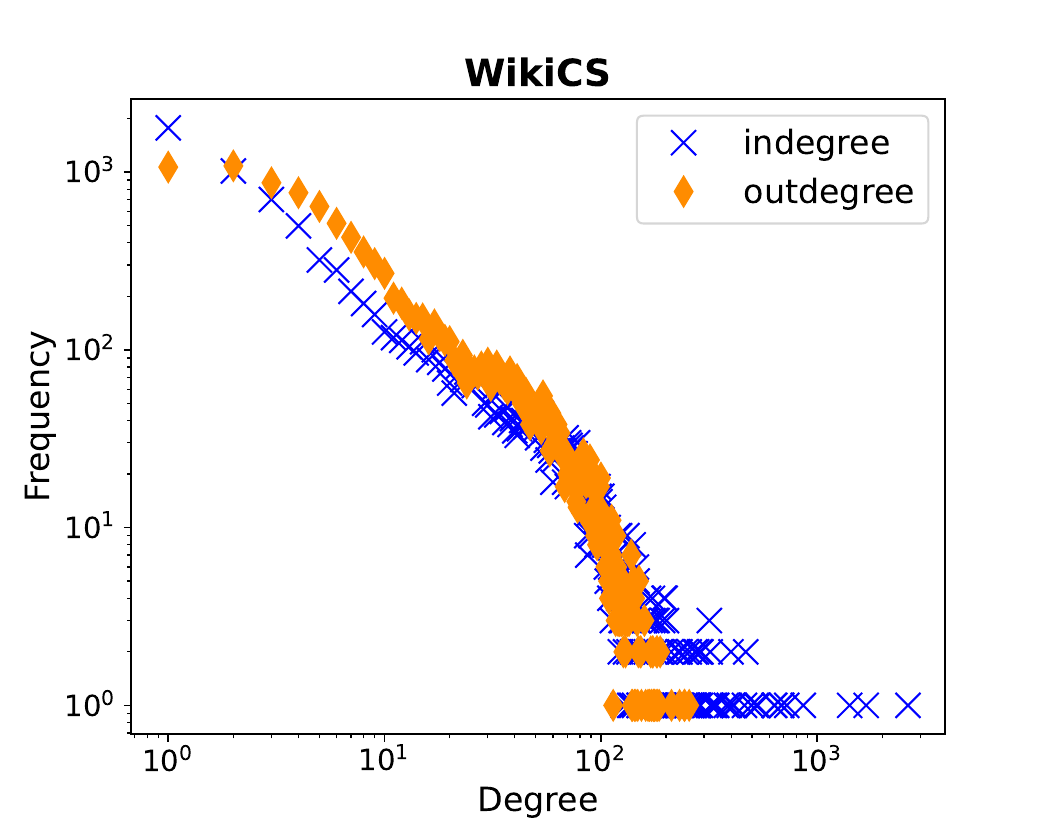}
   }
   \hspace{-7mm}
   \subfigure{
   \includegraphics[width=37mm]{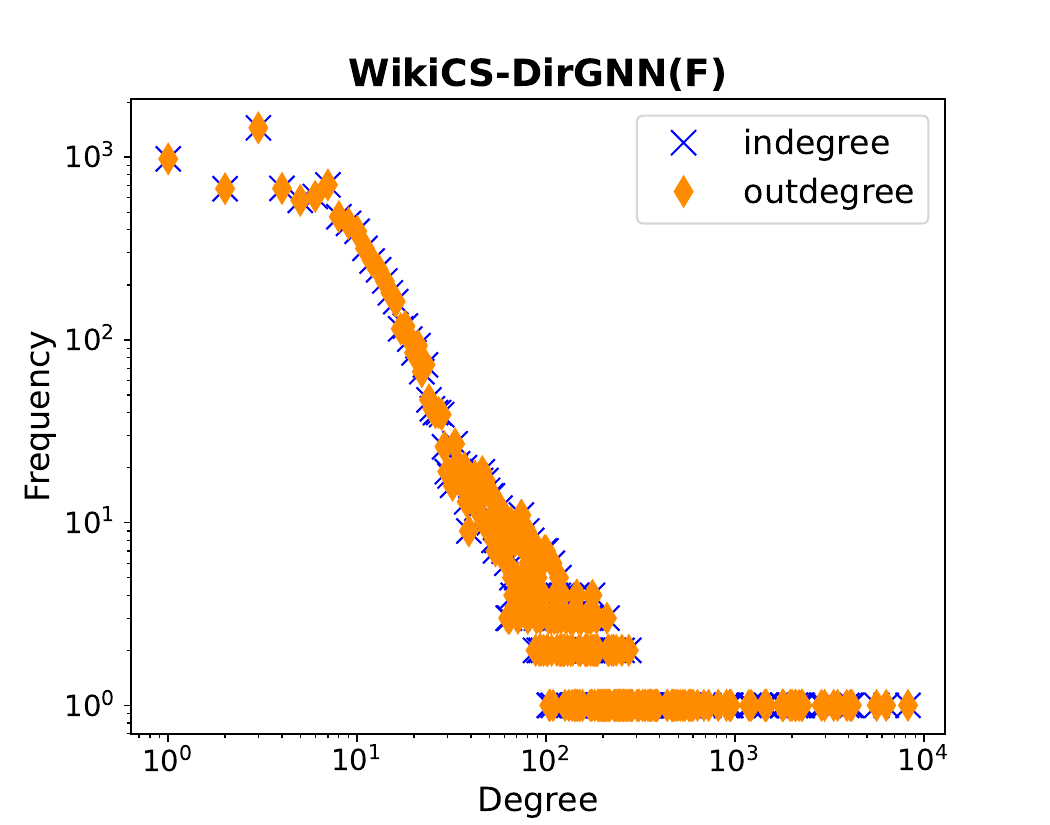}
   }
   \hspace{-7mm}
   \subfigure{
   \includegraphics[width=37mm]{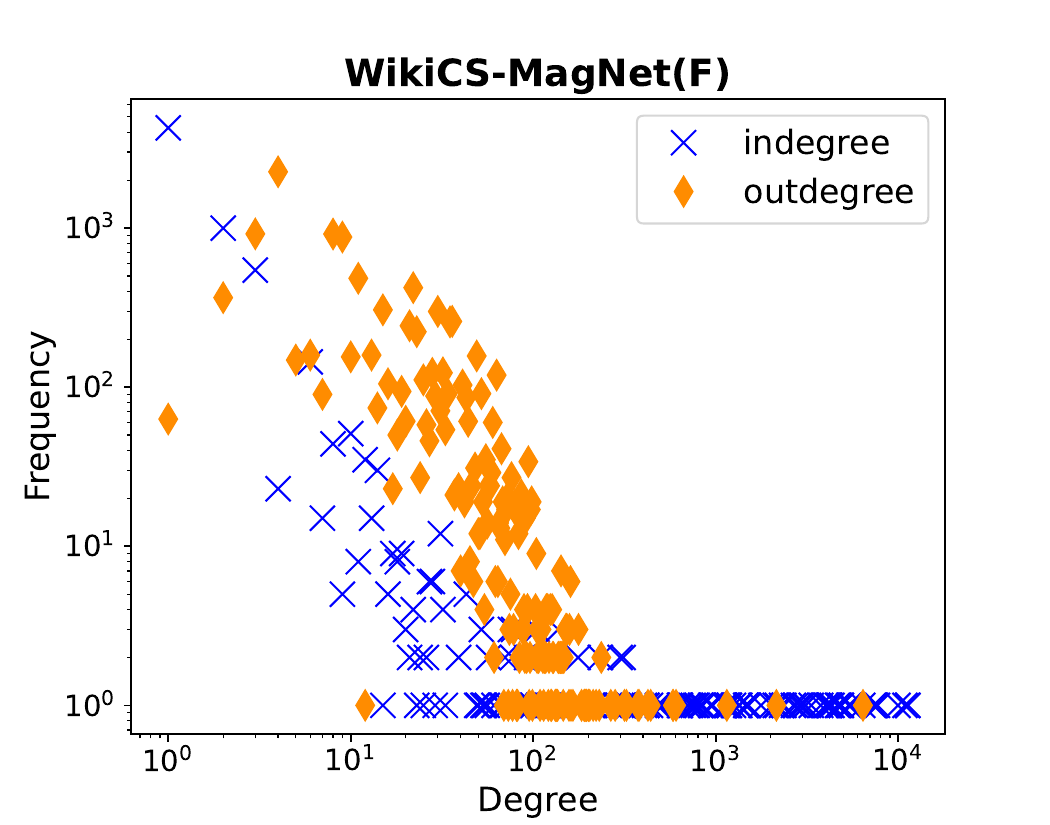}
   }
   \hspace{-7mm}
   \subfigure{
   \includegraphics[width=37mm]{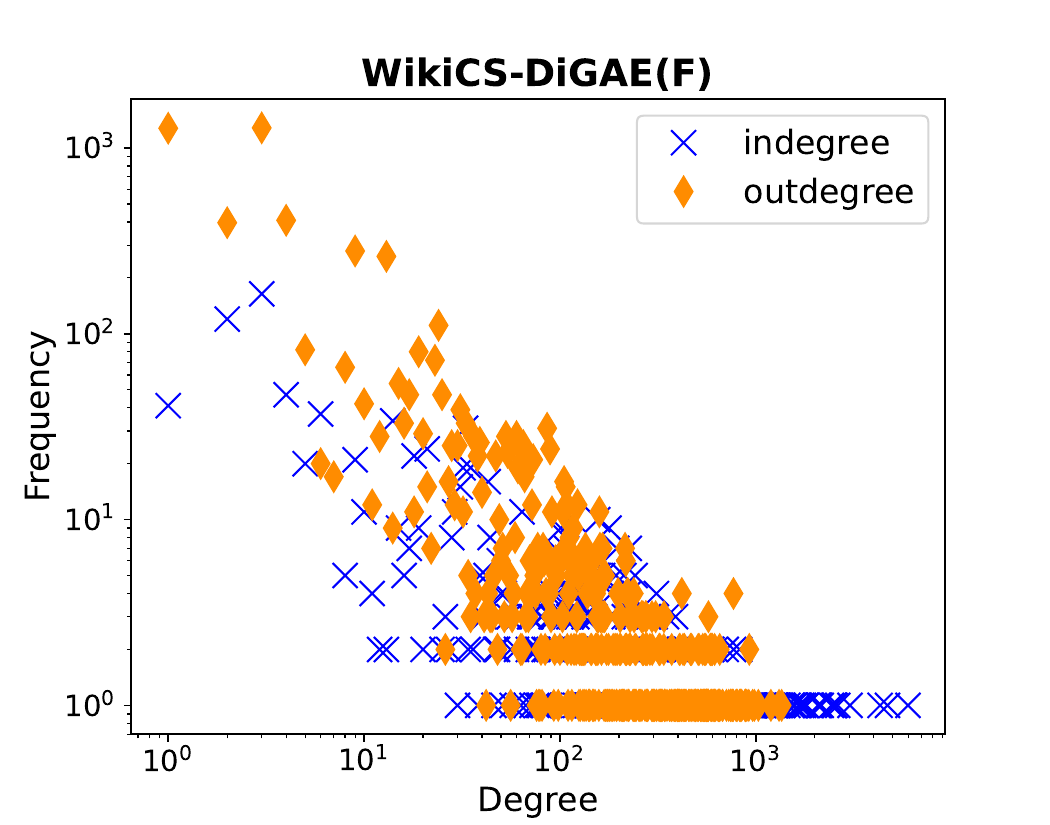}
   }
   \hspace{-7mm}
   \subfigure{
   \includegraphics[width=37mm]{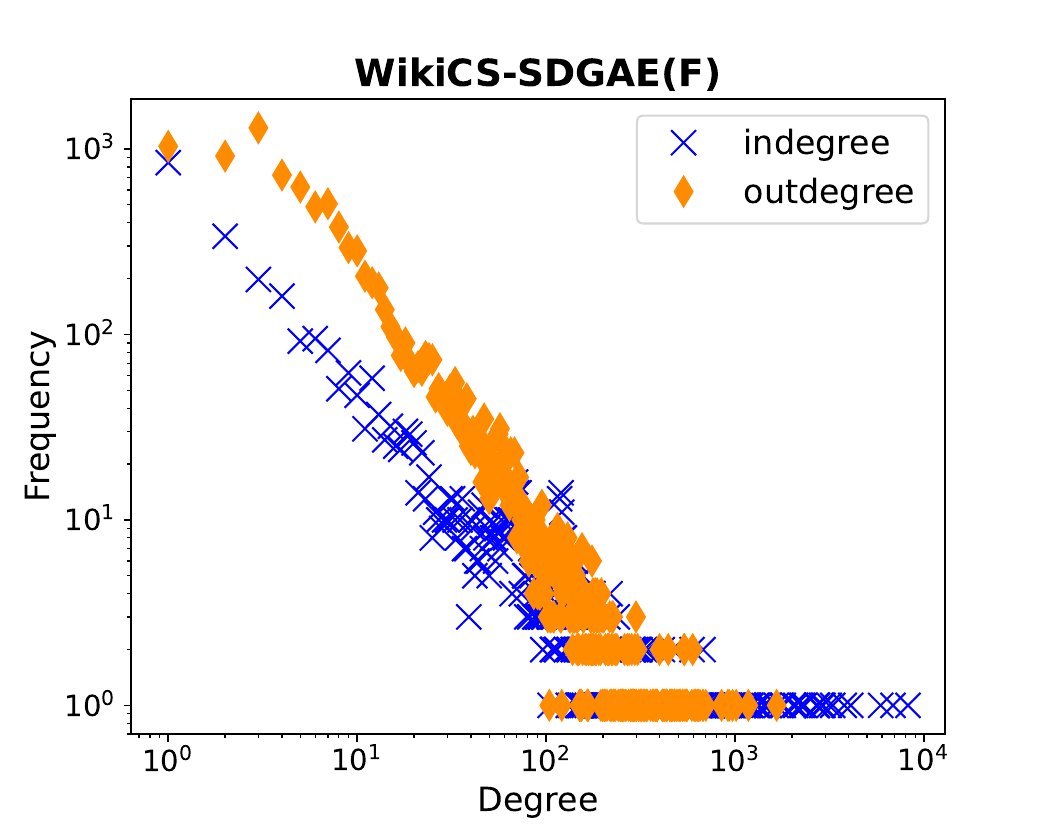}
   }
   \centering
   \caption{Degree distribution of WikiCS graph and its reconstruction graph generated by four GNNs using the original node feature as feature inputs.}
   \vspace{1mm}
   \label{fig:app_degree_feature}
\end{figure*}

\begin{figure*}[t]
    \centering
    \vspace{-2mm}
   \hspace{-3mm}
   \subfigure{
   \includegraphics[width=37mm]{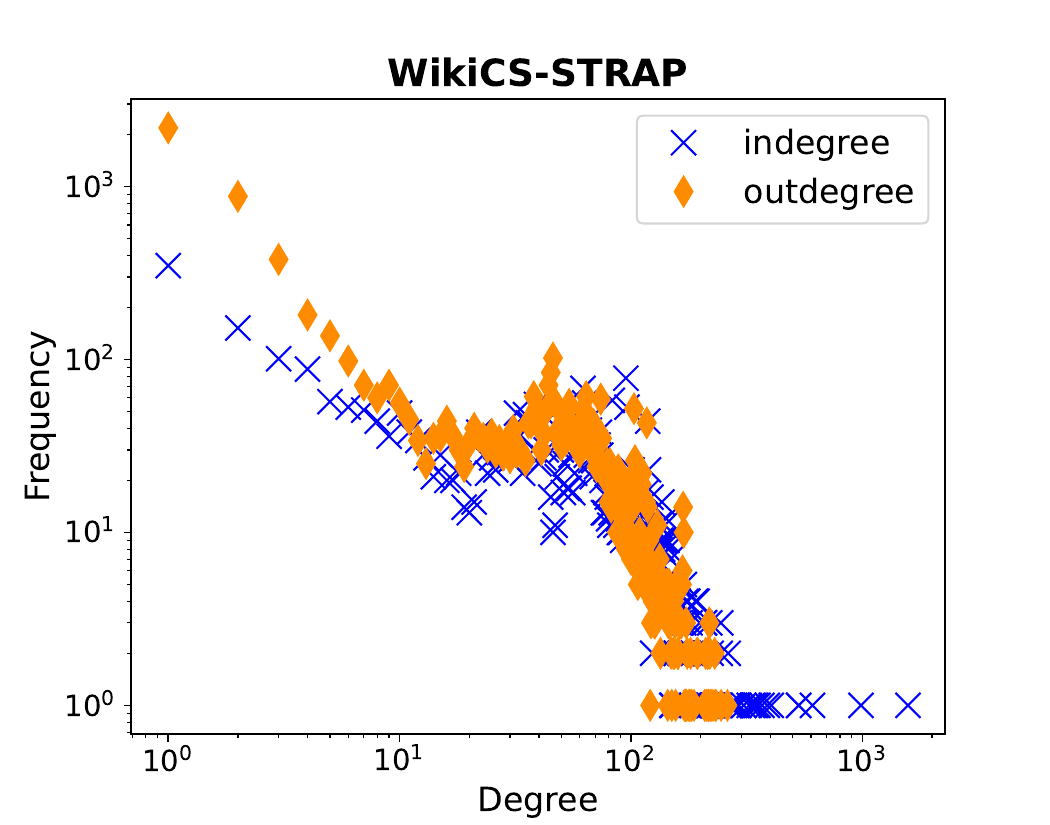}
   }
   \hspace{-7mm}
   \subfigure{
   \includegraphics[width=37mm]{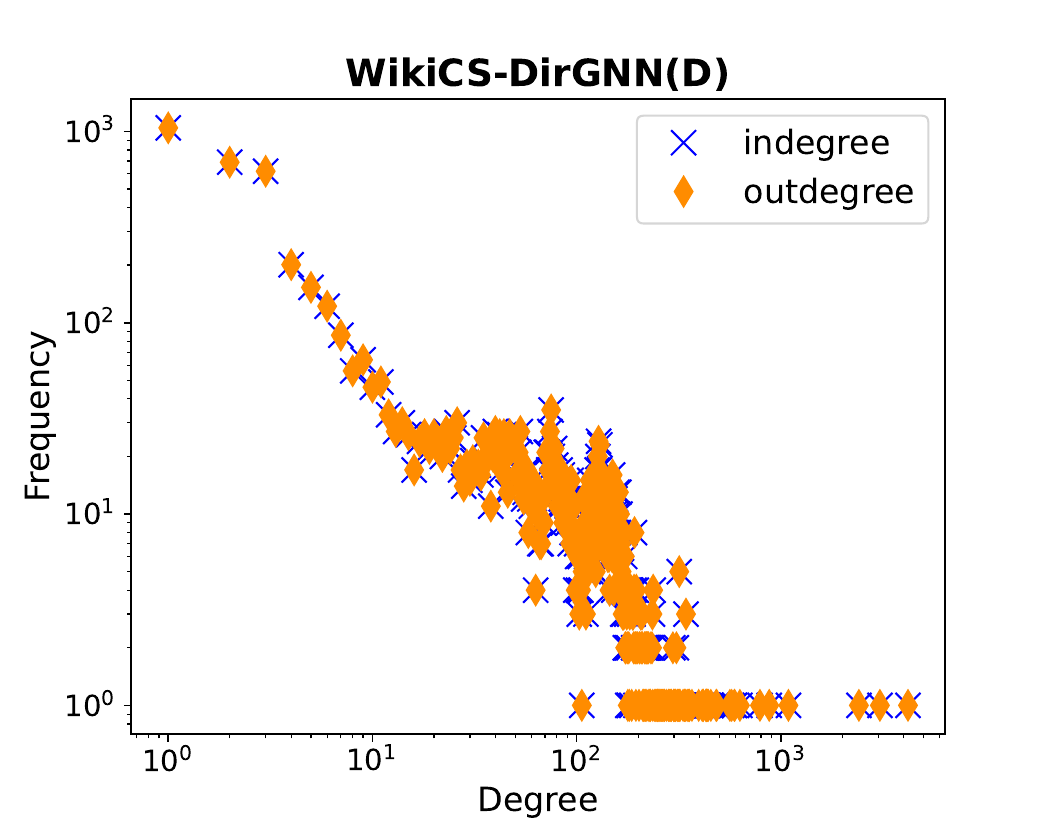}
   }
   \hspace{-7mm}
   \subfigure{
   \includegraphics[width=37mm]{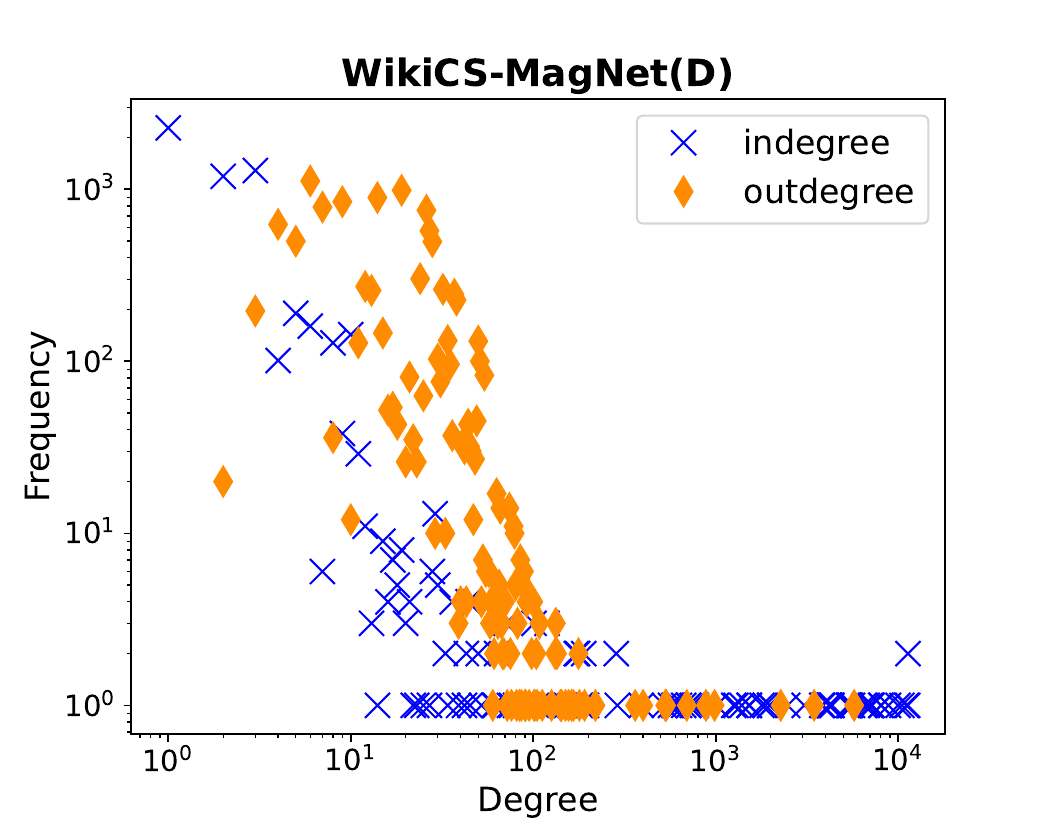}
   }
   \hspace{-7mm}
   \subfigure{
   \includegraphics[width=37mm]{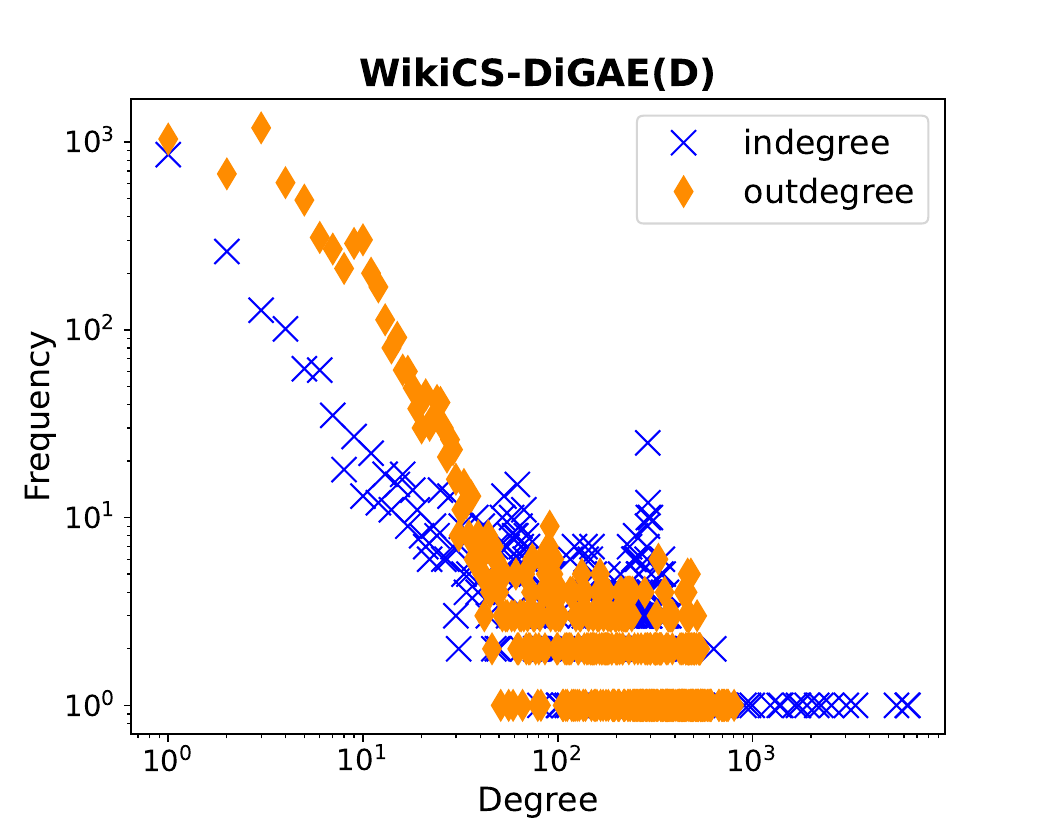}
   }
   \hspace{-7mm}
   \subfigure{
   \includegraphics[width=37mm]{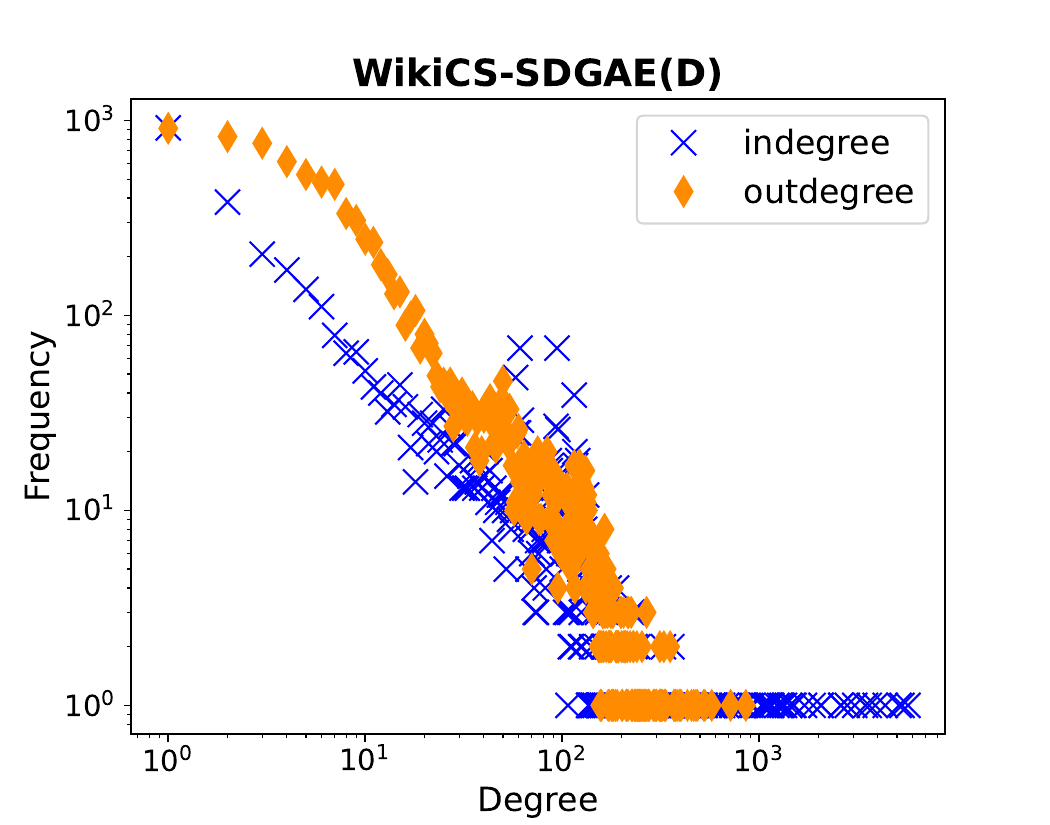}
   }
   \caption{Degree distribution of WikiCS's reconstruction graph generated by STRAP and four GNNs using the in/out degrees as feature inputs.}
    \label{fig:app_degree_degree}
    \vspace{1mm}
\end{figure*}

\begin{table*}[htbp]
    \centering
    \caption{Performance comparison of various GNNs using different negative sampling strategies on the Cora-ML and CiteSeer datasets. Results are reported under the \textbf{Hits@100} metric, with the best results highlighted in \textbf{bold}.}
    \resizebox{\textwidth}{!}{
    \begin{tabular}{llccccccc}
        \toprule
        Dataset & Sample & MLP & GCN & DiGCNIB & DirGNN & MagNet & DiGAE & SDGAE \\
        \midrule
        \multirow{2}{*}{Cora-ML} &each run& \textbf{60.61$\pm$6.64}  & \textbf{70.15$\pm$3.01}  & \textbf{80.57$\pm$3.21}  & \textbf{76.13$\pm$2.85}  & \textbf{56.54$\pm$2.95}  & \textbf{82.06$\pm$2.51}  & \textbf{90.37$\pm$1.33}  \\
        & each epoch & 34.15$\pm$3.54  & 59.86$\pm$9.89  & 56.74$\pm$4.08  & 49.89$\pm$3.59  & 54.79$\pm$2.98  & 79.76$\pm$3.28  & 89.71$\pm$2.36  \\
        \midrule
        \multirow{2}{*}{CiteSeer} & each run  & \textbf{70.27$\pm$3.40}  & \textbf{80.36$\pm$3.07}  & \textbf{85.32$\pm$3.70}  & \textbf{76.83$\pm$4.24}  & 65.32$\pm$3.26  & 83.64$\pm$3.21  & \textbf{93.69$\pm$3.68}  \\
        &  each epoch & 66.92$\pm$6.50  & 69.48$\pm$6.60  & 61.69$\pm$6.87  & 53.8$\pm$12.41  & \textbf{70.56$\pm$2.06}  & \textbf{87.32$\pm$3.79}  & 92.12$\pm$3.96  \\
        \bottomrule
    \end{tabular}}
    
    \label{app_sample}
\end{table*}

\subsection{Loss Function and Decoder}
We analyze the impact of different decoders and loss functions on the performance of directed link prediction methods. For the embedding methods STRAP, ELTRA, and ODIN, we compare their performance with four decoders: logistic regression with concatenation $\mathrm{LR}(\vs_u \| \vt_v)$, extended concatenation $\mathrm{LR}(\vs_u \| \vs_v \| \vt_u \| \vt_v)$, element-wise product $\mathrm{LR}(\vs_u \odot \vt_v)$, and inner product $\sigma(\vs_u^{\top} \vt_v)$~~\cite{eltra, strap, odin}. Results on the Cora-ML and Slashdot datasets are shown in Figure~\ref{fig:app_loss_embedding}. For single real-valued GNN methods, MLP, GCN, DiGCN-IB, and DirGNN, we evaluate different combinations of loss functions and decoders on the Cora-ML and Photo datasets, as shown in Figure~\ref{fig:app_loss_gnn}. The settings include CE loss with $\mathrm{MLP}(\vh_u \| \vh_v)$, and BCE loss with three decoders: $\mathrm{MLP}(\vh_u \| \vh_v)$, $\mathrm{MLP}(\vh_u \odot \vh_v)$, and inner product $\sigma(\vh_u^{\top} \vh_v)$.

The results highlight the significant impact of both the decoder and loss function on model performance. For embedding methods, even with fixed embeddings, different decoders result in substantial performance variations. Notably, ELTRA and ODIN exhibit high sensitivity to decoder choices on the Slashdot dataset. For GNNs, the results show that BCE loss offers a consistent advantage over CE loss.

Additionally, we compare the performance of MagNet and DiGAE under different loss functions and decoders. In Figure~\ref{fig:loss_magnet}, we show that BCE loss consistently outperforms CE loss for MagNet, underscoring the limitations of prior approaches that rely on CE loss~\cite{magnet,dpyg,duplex,lightdic}. Since link prediction is fundamentally a binary classification task, BCE loss is more appropriate, consistent with findings from undirected settings~\cite{li2023evaluating}. In Figure~\ref{fig:loss_digae}, we compare two decoders for DiGAE, inner product and MLP score, and observe that the MLP-based decoder achieves superior performance across three datasets.

These findings lead to two key insights: (1) \textbf{Decoder design significantly affects model performance}, and (2) \textbf{BCE loss is better suited for link prediction tasks than CE loss}. Furthermore, the poor performance of complex-valued methods (e.g., MagNet and DUPLEX) may be partly attributed to their reliance on CE loss and suboptimal decoder choices.

\subsection{Degree Distribution}

We evaluate how well different models preserve the asymmetry of directed graphs by analyzing their degree distributions. Following STRAP~\cite{strap}, we compute the predicted probability for every edge and select the top $m^{\prime}$ edges, where $m^{\prime}$ is the number of edges in the training graph, to reconstruct the graph. Figure~\ref{fig:app_degree_feature} compares the true in-/out-degree distributions of the WikiCS training graph with those of reconstructed graphs generated by four GNNs (DirGNN~\cite{dirgnn}, MagNet~\cite{magnet}, DiGAE~\cite{digae}, and SDGAE), using original node features as input. Figure~\ref{fig:app_degree_degree} presents a similar comparison, but the reconstructed graphs are generated by STRAP and the same four GNNs using in/out degrees as feature inputs.

The results show that STRAP and SDGAE most accurately preserve the degree distributions, with STRAP performing exceptionally well in capturing in-degree components, explaining its strong performance on WikiCS. And DirGNN, using the decoder $\mathrm{MLP}\bigl(\vh_u \odot \vh_v\bigr)$, produces identical in-/out-degree distributions but still captures in-degree components correctly. In contrast, MagNet fails to learn meaningful degree distributions, resulting in poor performance. Moreover, when GNNs are provided with in/out degrees as input features, they better preserve the degree distribution, aligning with their improved WikiCS performance (see supplemental material for detailed results). Finally, a direct comparison between DiGAE and SDGAE reveals that SDGAE more effectively maintains the degree distribution, further demonstrating its advantage.

\textbf{These findings reinforce the importance of preserving asymmetry in directed link prediction}, as discussed in Section~\ref{sec_unif}. They also highlight an underexplored challenge: the need for GNNs to better preserve in-/out-degree distributions, a task that embedding methods like STRAP currently handle more effectively.

\subsection{Negative Sampling Strategy}

We present a performance comparison of various GNNs trained using different negative sampling strategies on the Cora-ML and CiteSeer datasets in Table~\ref{app_sample}. Results are reported using the Hits@100 metric, with the best results highlighted in bold. In this context, ``each run" strategy refers to the default setting in DirLinkBench, where a random negative sample is generated for each run and shared across all models. In contrast, ``each epoch" represents a strategy where different models randomly sample negative edges in each training epoch. In both settings, the positive sample splits and test sets remain consistent for fair comparison.

The results demonstrate that the choice of negative sampling strategy during training can significantly affect model performance, particularly for single real-valued GNNs, where performance declines noticeably under the ``each epoch" strategy. In contrast, the ``each run" strategy tends to improve performance across most GNN models. These findings underscore the importance of further research into negative sampling techniques. For example, heuristic-based approaches have been proposed for undirected graphs as alternatives to random sampling~\cite{li2023evaluating}, and similar methods could be explored for directed graphs in future work.

\section{Conclusion}  

This paper presents a unified framework for evaluating the expressiveness of directed link prediction methods, emphasizing the theoretical importance of dual embeddings and decoder design. To address the lack of standardized benchmarks in this area, we introduce DirLinkBench, a new robust benchmark featuring diverse real-world directed graphs, standardized data splits, varied feature initialization strategies, and comprehensive evaluation metrics, including ranking-based metrics introduced to this task. Using DirLinkBench, we find that current methods often perform inconsistently across datasets, and that simple design choices, such as feature inputs, loss functions, decoders, and negative sampling, can significantly impact performance. Then we revisit the DiGAE model, showing its graph convolution is theoretically equivalent to GCN on an undirected bipartite graph. Building on this insight, we propose SDGAE, a novel Spectral Directed Graph Auto-Encoder that uses polynomial approximation to learn graph filters for directed graphs. SDGAE achieves state-of-the-art average performance and better preserves directed structural properties.

Our findings highlight two key challenges for future research:
(1) \textbf{How can more expressive and efficient decoders be developed, especially for complex-valued methods?}
(2) \textbf{How can GNN architectures better capture and preserve asymmetry, such as in- and out-degree distributions?}
We believe that DirLinkBench, along with our proposed SDGAE and the insights presented in this work, will serve as a foundation for advancing the field of directed link prediction. We hope this benchmark encourages the development of more robust, expressive, and theoretically grounded methods. The complete benchmark and implementation are provided in the Appendix and source code to facilitate future research.

\bibliography{ref}
\bibliographystyle{icml2025}

\appendix
\onecolumn
\clearpage
\section{Proof}
\subsection{The proof of Proposition~\ref{prop_single}}\label{app_proof_prop_single}
\begin{proof}
In the case of single methods, each node \( u \) in a directed graph is represented by a real-valued embedding \( \vh_u \).  We will use the directed graphs (a) and (d) from Figure~\ref{fig:bg} as examples: (a) is a directed ring graph, and (d) is a regular directed graph. We will show that single methods with the decoder function \( \mathrm{MLP}(\vh_u \| \vh_v) \) can enable reconstruction for graph (d) but not for graph (a).

For graph (d), which consists of three nodes and three edges (i.e., \( 1 \to 2, 3 \to 2, 3 \to 1 \)), each node is assigned a real-valued embedding, \( \vh_1, \vh_2, \vh_3 \in \mathbb{R}^{d \times 1} \). Let the decoder be a simple MLP with the sigmoid activation function \( \sigma = \mathrm{Sigmoid}(\cdot) \). For the edge \( 1 \to 2 \), the probability of the directed edge is given by:
\begin{align}
    p(1,2) &= \sigma(\vh_1\vw_1 + \vh_2\vw_2) > 0.5,\\ \quad p(2,1) &= \sigma(\vh_2\vw_1 + \vh_1\vw_2) < 0.5,
\end{align}
where $\vw_1, \vw_2 \in \mathbb{R}^{d \times 1}$ are the learnable weights. From these inequalities, we obtain the following system of constraints:
\begin{equation}\label{1-2}
    \vh_1\vw_1 + \vh_2\vw_2 > 0, \quad \vh_2\vw_1 + \vh_1\vw_2 < 0.
\end{equation}
Similarly, for edges $2 \to 3$ and $3 \to 1$, we obtain the following inequalities:
\begin{align}
    \vh_3\vw_1 + \vh_2\vw_2 > 0&, \quad \vh_2\vw_1 + \vh_3\vw_2 < 0,\label{3-2} \\ 
    \vh_3\vw_1 + \vh_1\vw_2 > 0&, \quad \vh_1\vw_1 + \vh_3\vw_2 < 0. \label{3-1}
\end{align}
By solving the three sets of inequalities~(\ref{1-2}), (\ref{3-2}), and (\ref{3-1}), we find that they have a solution. For instance, if \( d = 2 \), one possible solution is \( \vw_1 = (1, 0) \), \( \vw_2 = (0, 1) \), \( \vh_1 = (1, -1) \), \( \vh_2 = (-1, 1) \), and \( \vh_3 = (2, -2) \). Therefore, single methods with \( \mathrm{MLP}(\vh_u \| \vh_v) \) can successfully capture the graph structure and enable reconstruction for graph (d).

For the directed ring graph (a), which consists of three nodes and three edges (\( 1 \to 2, 2 \to 3, 3 \to 1 \)), we similarly derive a set of inequalities. For the edge \( 1 \to 2 \), we obtain the same inequalities as in~(\ref{1-2}). By subtracting the second inequality from the first, we get:
\begin{equation}\label{eq_1_2}
    (\vh_1 - \vh_2)\vw_1 + (\vh_2 - \vh_1)\vw_2 > 0.
\end{equation}
Similarly, for edges $2 \to 3$ and $3 \to 1$, we derive:
\begin{align}
    &(\vh_2 - \vh_3)\vw_1 + (\vh_3 - \vh_2)\vw_2 > 0,\label{eq_2_3}\\  &(\vh_3 - \vh_1)\vw_1 + (\vh_1 - \vh_3)\vw_2 > 0.\label{eq_3_1} 
\end{align}
Adding these three inequalities~(\ref{eq_1_2}), (\ref{eq_2_3}), and (\ref{eq_3_1}), results in $0 > 0$, which is a contradiction and indicates that no embeddings \( \vh_1, \vh_2, \vh_3 \) and weights \( \vw_1, \vw_2 \) exist that can satisfy these conditions. The same result holds even when nonlinearities are added to the MLP. Therefore, single methods with the decoder \( \mathrm{MLP}(\vh_u \| \vh_v) \) fail to compute the probabilities for the directed ring graph.

This example demonstrates that while single methods with \( \mathrm{MLP}(\vh_u \| \vh_v) \) can capture the graph structure and enable reconstruction for certain directed graphs, they fail for directed ring graphs.

\end{proof}



\subsection{The proof of Lemma~\ref{lemma_digae_conv}}\label{app_proof_lemma_digae_conv}

\begin{proof}
Substituting the block matrix $\mathcal{S}(\hat{\mA})=
    \left[ 
        \begin{array}{cc}
            \mathbf{0} & \hat{\mA} \\
            \hat{\mA}^{\top} & \mathbf{0}
        \end{array}
    \right]$ into the graph convolution of DiGAE's encoder (i.e., Equation (\ref{lemma3_eq}), we obtain: 
\begin{align}
    \left[ 
        \begin{array}{c}
\mS^{(\ell+1)}\\
\mT^{(\ell+1)}
        \end{array}
    \right] &=
    \sigma\left(\left[ 
        \begin{array}{cc}
            \mathbf{0} & \hat{\mA} \\
            \hat{\mA}^{\top} & \mathbf{0}
        \end{array}
    \right] 
    \left[ 
        \begin{array}{c}
\mS^{(\ell)}\mW_{S}^{(\ell)}\\
\mT^{(\ell)}\mW_{T}^{(\ell)}
        \end{array}
    \right]\right) \\
    &= \left[ 
        \begin{array}{c}
\sigma\left(\hat{\mA}\mT^{(\ell)}\mW_{T}^{(\ell)}\right)\\
\sigma\left(\hat{\mA}^{\top}\mS^{(\ell)}\mW_{S}^{(\ell)}\right)
        \end{array}
    \right].
\end{align}
This result corresponds exactly to Equations (\ref{digae_s}) and (\ref{digae_t}) when the degree-based normalization of $\hat{\mA}$ is not considered. If we include the degree-based normalization, it actually applies to $\mathcal{S}(\hat{\mA})$. Notably, the diagonal degree matrix of $\mathcal{S}(\hat{\mA})$ is given by $\mathrm{diag}\left(\hat{\mD}_{\rm out}, \hat{\mD}_{\rm in}\right)$. Therefore, we have
\begin{align}
 \left[ 
        \begin{array}{c}
        \mS^{(\ell+1)}\\
        \mT^{(\ell+1)}
        \end{array}
    \right]
    &= \sigma\left(
    \left[ 
        \begin{array}{cc}
            \hat{\mD}_{\rm out}^{-\beta} & \mathbf{0} \\
            \mathbf{0} & \hat{\mD}_{\rm in}^{-\alpha}
        \end{array}
    \right]
    \left[ 
        \begin{array}{cc}
            \mathbf{0} & \hat{\mA} \\
            \hat{\mA}^{\top} & \mathbf{0}
        \end{array}
    \right] 
    \left[ 
        \begin{array}{cc}
            \hat{\mD}_{\rm out}^{-\beta} & \mathbf{0} \\
            \mathbf{0} & \hat{\mD}_{\rm in}^{-\alpha}
        \end{array}
    \right]
    \left[ 
        \begin{array}{c}
        \mS^{(\ell)}\mW_{S}^{(\ell)}\\
        \mT^{(\ell)}\mW_{T}^{(\ell)}
        \end{array}
    \right]\right) \\
    &= \left[ 
        \begin{array}{c}
\sigma\left(\hat{\mD}_{\rm out}^{-\beta}\hat{\mA}\hat{\mD}_{\rm in}^{-\alpha}\mT^{(\ell)}\mW_{T}^{(\ell)}\right)\\
\sigma\left(\hat{\mD}_{\rm in}^{-\alpha}\hat{\mA}^{\top}\hat{\mD}_{\rm out}^{-\beta}\mS^{(\ell)}\mW_{S}^{(\ell)}\right)
        \end{array}
    \right].
\end{align}
These results exactly equal the graph convolution used in DiGAE’s encoder, as defined in Equations (\ref{digae_s}) and (\ref{digae_t}). Therefore, the graph convolution in DiGAE’s encoder matches the form of Equation (\ref{lemma3_eq}), suggesting that it effectively corresponds to a GCN convolution on an undirected bipartite graph.
\end{proof}


\hypertarget{app_proof_le_norm}{}
\subsection{The proof of Lemma~\ref{le:norm}}\label{app_proof_le_norm}

\begin{proof}
     Given the block adjacency matrix $\mathcal{S}(\hat{\mA})=
    \left[\begin{array}{cc}
        \mathbf{0} & \hat{\mA}\\
        \hat{\mA}^{\top} & \mathbf{0}
    \end{array} \right]$ and its degree matrix $\mD_{\mathcal{S}}= \left[\begin{array}{cc}
        \hat{\mD}_{\rm out} & \mathbf{0}\\
        \mathbf{0} & \hat{\mD}_{\rm in}
    \end{array} \right]$, we have 
    \begin{align} 
    \mD_{\mathcal{S}}^{-1/2}\mathcal{S}(\hat{\mA})\mD_{\mathcal{S}}^{-1/2}&=
    \left[\begin{array}{cc}
        \hat{\mD}_{\rm out} & \mathbf{0}\\
        \mathbf{0} & \hat{\mD}_{\rm in}
    \end{array} \right]^{-1/2}
    \left[\begin{array}{cc}
        \mathbf{0} & \hat{\mA}\\
        \hat{\mA}^{\top} & \mathbf{0}
    \end{array} \right]
    \left[\begin{array}{cc}
        \hat{\mD}_{\rm out} & \mathbf{0}\\
        \mathbf{0} & \hat{\mD}_{\rm in}
    \end{array} \right]^{-1/2} \\
    &=
    \left[\begin{array}{cc}
        \hat{\mD}_{\rm out}^{-1/2} & \mathbf{0}\\
        \mathbf{0} & \hat{\mD}_{\rm in}^{-1/2}
    \end{array} \right]
    \left[\begin{array}{cc}
        \mathbf{0} & \hat{\mA}\\
        \hat{\mA}^{\top} & \mathbf{0}
    \end{array} \right]
    \left[\begin{array}{cc}
        \hat{\mD}_{\rm out}^{-1/2} & \mathbf{0}\\
        \mathbf{0} & \hat{\mD}_{\rm in}^{-1/2}
    \end{array} \right] \\
    &=
    \left[\begin{array}{cc}
        \mathbf{0} & \hat{\mD}_{\rm out}^{-1/2}\hat{\mA}\hat{\mD}_{\rm in}^{-1/2}\\
        \left(\hat{\mD}_{\rm out}^{-1/2}\hat{\mA}\hat{\mD}_{\rm in}^{-1/2} \right)^{\top} & \mathbf{0}
    \end{array} \right]\\
    &=
     \left[\begin{array}{cc}
        \mathbf{0} & \Tilde{\mA}\\
        \Tilde{\mA^{\top}} & \mathbf{0}
    \end{array} \right]\\
    &=
    \mathcal{S}(\Tilde{\mA}).
    \end{align}
Therefore, $\mD_{\mathcal{S}}^{-1/2}\mathcal{S}(\hat{\mA})\mD_{\mathcal{S}}^{-1/2} = \mathcal{S}(\Tilde{\mA})$, showing that the normalization $\Tilde{\mA} = \hat{\mD}_{\rm out}^{-1/2}\hat{\mA}\hat{\mD}_{\rm in}^{-1/2} $ is equivalent to the symmetric normalization of $\mathcal{S}(\hat{\mA})$. These findings suggest that $\Tilde{\mA} = \hat{\mD}_{\rm out}^{-1/2}\hat{\mA}\hat{\mD}_{\rm in}^{-1/2} $ performs symmetric normalization on the adjacency matrix of an undirected bipartite graph, aligning with the common normalization scheme used in graph neural networks~\cite{gcn,chebnet}.
\end{proof}

\section{More Details of Experimental Settings}
\subsection{Metric description}\label{app_bench_metric}
\textbf{Mean Reciprocal Rank (MRR)} evaluates the capability of models to rank the first correct entity in link prediction tasks. It assigns higher weights to top-ranked predictions by computing the average reciprocal rank of the first correct answer across queries: $\text{MRR} = \frac{1}{|Q|} \sum_{i=1}^{|Q|} \frac{1}{\text{rank}_i}$, where \(|Q|\) is the total number of queries and \(\text{rank}_i\) denotes the position of the first correct answer for the \(i\)-th query. MRR emphasizes early-ranking performance, making it sensitive to improvements in top predictions.

\textbf{Hits@K} measures the proportion of relevant items that appear in the top-$\text{K}$ positions of the ranked list of items. For \(N\) queries, Hits@K$=\frac{1}{N}\sum\nolimits_{i=1}^{N}\mathbf{1}(\text{rank}_i\leq \text{K})$,
where rank$_i$ is the rank of the $i$-th sample and the indicator function $\mathbf{1}$ is 1 if rank$_i \leq \text{K}$, and 0 otherwise. Following the OGB benchmark~\cite{ogb}, link prediction implementations compare each positive sample's score against a set of negative sample scores. A ``hit" occurs if the positive sample's score surpasses at least K-1 negative scores, with final results averaged across all queries.

\textbf{Area Under the Curve (AUC)} measures the likelihood that a positive sample is ranked higher than a random negative sample. \(\text{AUC} = \frac{\sum_{i=1}^{M}\sum_{j=1}^{N}\mathbf{1}(s_i^{\text{pos}} > s_j^{\text{neg}})}{M \times N}\), where \(M\) and \(N\) are positive/negative sample counts, \(s_i^{\text{pos}}\) and \(s_j^{\text{neg}}\) their prediction scores. Values approaching 1 indicate perfect separation of positive and negative edges.

\textbf{Average Precision (AP)} is defined as the area under the Precision-Recall (PR) curve. Formally, \(\text{AP} = \sum_{i=1}^{N} (R_i - R_{i-1}) \times P_i,\) where \(P_i\) is the precision at the \(i\)-th threshold, \(R_i\) is the recall at the \(i\)-th threshold, and \(N\) is the number of thresholds considered. 

\textbf{Accuracy (ACC)} measures the proportion of correctly predicted samples among all predictions. Formally, \(\text{ACC} = \frac{TP + TN}{TP + TN + FP + FN},\) where \(TP\), \(TN\), \(FP\), and \(FN\) represent true positives, true negatives, false positives, and false negatives, respectively.

\subsection{DirLinkBench Setting}\label{app_dirlinbench_set}
\textbf{Baseline Implementations}. For MLP, GCN, GAT, and APPNP, we use the PyTorch Geometric (PyG) library~\cite{pyg} implementations. For DCN, DiGCN, and DiGCNIB, we use the PyTorch Geometric Signed Directed (PyGSD) library~\cite{dpyg} implementations. For other baselines, we use the original code released by the authors. Here are the links to each repository.

\begin{itemize}
    \item PyG: \href{https://github.com/pyg-team/pytorch_geometric/tree/master/benchmark}{https://github.com/pyg-team/pytorch\_geometric}
    \item PyGSD: \href{https://github.com/SherylHYX/pytorch_geometric_signed_directed}{https://github.com/SherylHYX/PyGSD}
    \item STRAP: \href{https://github.com/yinyuan1227/STRAP-git}{https://github.com/yinyuan1227/STRAP-git}
    \item ODIN: \href{https://github.com/hsyoo32/odin}{https://github.com/hsyoo32/odin}
    \item ELTRA: \href{https://github.com/mrhhyu/ELTRA}{https://github.com/mrhhyu/ELTRA}
    \item GPRGNN: \href{https://github.com/jianhao2016/GPRGNN}{https://github.com/jianhao2016/GPRGNN}
    \item DiGAE: \href{https://github.com/gidiko/DiGAE}{https://github.com/gidiko/DiGAE}
    \item DHYPR: \href{https://github.com/hongluzhou/dhypr}{https://github.com/hongluzhou/dhypr}
    \item DirGNN: \href{https://github.com/emalgorithm/directed-graph-neural-network}{https://github.com/emalgorithm/DirGNN}
    \item MagNet: \href{https://github.com/matthew-hirn/magnet}{https://github.com/matthew-hirn/magnet}
    \item DUPLEX: \href{https://github.com/alipay/DUPLEX}{https://github.com/alipay/DUPLEX}
\end{itemize}

\textbf{Hyperparameter settings}. 
The hyperparameter settings for the baselines in DirLinkBench are detailed below, where ``hidden" represents the number of hidden units, ``embedding" refers to the embedding dimension, ``undirected" indicates whether an undirected training graph is used, ``lr" stands for the learning rate, and ``wd" denotes weight decay.
\begin{itemize}
    \item MLP: hidden: 64, embedding: 64, layer: 2, lr: \{0.01, 0.005\}, wd:  \{0.0, 5e-4\}.
    \item GCN: hidden: 64, embedding: 64, layer: 2, undirected: \{True, False\}, lr: \{0.01, 0.005\}, wd:  \{0.0, 5e-4\}.
    \item GAT: hidden: 8, heads: 8, embedding: 64, layer: 2,  undirected: \{True, False\}, lr: \{0.01, 0.005\}, wd:  \{0.0, 5e-4\}
    \item APPNP: hidden: 64, embedding: 64, layer: 2, $K$: 10, $\alpha$: \{0.1,0.2\}, undirected: \{True, False\}, lr: \{0.01, 0.005\}, wd:  \{0.0, 5e-4\}.
    \item GPRGNN: hidden: 64, embedding: 64, layer: 2, Init: PPR, $K$: 10, $\alpha$: \{0.1,0.2\}, undirected: \{True, False\}, lr: \{0.01, 0.005\}, wd:  \{0.0, 5e-4\}.
    \item DGCN: hidden: 64, embedding: 64, lr: \{0.01, 0.005\}, wd:  \{0.0, 5e-4\}.
    \item DiGCN and DiGCNIB: hidden: 64, embedding: 64, $\alpha$: \{0.1,0.2\}, layer: 2, lr: \{0.01, 0.005\}, wd:  \{0.0, 5e-4\}.
    \item DirGNN:  hidden: 64, embedding: 64, layer: 2, $\alpha$: \{0.0, 0.5, 1.0\},  jk: \{``cat", ``max'"\}, normalize: \{True, False\}, lr: \{0.01, 0.005\}, wd:  \{0.0, 5e-4\}.
    \item MagNet: hidden: 64, embedding: 64, layer: 2, $K$: \{1, 2\}, $q$: \{0.05, 0.1, 0.15, 0.2, 0.25\}, lr: \{0.01, 0.005\}, wd:  \{0.0, 5e-4\}.
    \item DUPLEX: hidden: 64, embedding: 64, layer: \{2, 3\}, head: 1, loss weight:\{0.1, 0.3\}, loss decay: \{0.0, 1e-2, 1e-4\}, lr: \{0.01, 0.005\}, wd:  \{0.0, 5e-4\}.
    \item DHYPR: hidden: 64, embedding: 32, proximity: \{1, 2\}, $\lambda$: \{0.01, 0.05, 1, 5\}, lr: \{0.01, 0.001\}, wd:  \{0.0, 0.001\}.
    \item DiGAE: hidden: 64, embedding: 64, layer: \{1, 2\}, $(\alpha,\beta): \{0.0, 0.2, 0.4, 0.6, 0.8\}^2$, lr: \{0.01, 0.005\}, wd:  \{0.0, 5e-4\}.
\end{itemize}
Regarding the selection of these hyperparameters, we begin by setting each model to use two layers, with both hidden and embedding dimensions set to 64. This ensures that all methods have approximately the same number of learnable parameters, a common practice in many GNN benchmarks~\cite{ogb,gprgnn}. 
Some models have exceptions. For example, DUPLEX uses three layers across all datasets in its official implementation; given its status as a recent advanced method, we search over layers in \{2, 3\}. For DYHPR, due to its high computational complexity and the fact that it generates multiple embeddings for each node, we follow its original implementation and set the embedding dimension to 32. We retain the layer setting for DiGAE, which uses one or two layers in its released code.

For general hyperparameters such as learning rate and weight decay, we approximately keep the search space consistent across all models. For model-specific parameters (e.g., $q$ in MagNet, and $\alpha, \beta$ in DiGAE), we adhere to the search ranges reported in their respective papers. All hyperparameters are tuned via grid search to identify the optimal settings under our DirLinkBench benchmark.

\begin{table}[t]
\centering
\caption{The parameters of SDGAE on different datasets.}
\begin{tabular}{lcccccc}
\toprule
Datasets &hidden &embedding &MLP layer &$K$ &lr &wd  \\ \midrule
Cora-ML & 64 & 64 &1 &5 &0.01 &0.0 \\
CiteSeer& 64 & 64 &1 &5 &0.01 &0.0 \\
Photo & 64 & 64 &2 &5 &0.005 &0.0 \\
Computers & 64 & 64 &2 & 3 & 0.005 &0.0\\
WikiCS & 64 & 64 & 2&5 & 0.005 &5e-4 \\
Slashdot & 64 & 64 &2 &5 &0.01 & 5e-4 \\
Epinions & 64 & 64 &2 &5 & 0.005 &0.0 \\
\bottomrule
\end{tabular}
\label{app_para_sdage}
\end{table}

\subsection{Hyperparameter setting of SDGAE}\label{app_sdgae_setting}
For the SDGAE hyperparameter settings, we aligned our settings with those of other baselines in DirLinkBench to ensure fairness. For the MLP used in $\mX$ initialization, we set the number of layers to one or two, matching DiGAE's convolutional layer configurations. The number of hidden units and the embedding dimension were both set to 64. The learning rate (lr) was chosen as either 0.01 or 0.005, and weight decay (wd) was set to 0.0 or 5e-4, following the configurations of most GNN baselines in DirLinkBench. The polynomial order $K$ was searched in the set \{3, 4, 5\}.  We performed a grid search to optimize parameters on the validation set, and Table~\ref{app_para_sdage} presents the corresponding SDGAE parameters for different datasets. 


\end{document}